\documentclass[acmsmall]{acmart}
\usepackage{longtable}
\usepackage{lscape}

\usepackage{tikz}
\usetikzlibrary{matrix,positioning}

\AtBeginDocument{%
  \providecommand\BibTeX{{%
    \normalfont B\kern-0.5em{\scshape i\kern-0.25em b}\kern-0.8em\TeX}}}

\setcopyright{acmcopyright}
\acmJournal{CSUR}
\acmYear{2021} \acmVolume{1} \acmNumber{1} \acmArticle{1} \acmMonth{1} \acmPrice{15.00}\acmDOI{10.1145/3490235}

\acmJournal{JACM}
\acmVolume{37}
\acmNumber{4}
\acmArticle{111}
\acmMonth{8}

\begin{document}

\title{Gait Recognition Based on Deep Learning: A Survey}


\author{Claudio Filipi Gon\c{c}alves dos Santos}
\authornote{Authors contributed equally to this research.}
\email{cfsantos@ufscar.br}
\affiliation{%
  \institution{Federal Institute of S\~{a}o Carlos - UFSCar}
  \streetaddress{Rod. Washington Luiz, 235}
  \city{S\~{a}o Carlos}
  \state{S\~{a}o Paulo}
  \country{Brazil}}
\affiliation{%
  \institution{Eldorado Research Institute}
  \streetaddress{Av. Alan Turing, 275}
  \city{Campinas}
  \state{S\~{a}o Paulo}
  \country{Brazil}}

\author{Diego de Souza Oliveira}
\authornotemark[1]
\email{diego.s.oliveira@unesp.br}
\author{Leandro A. Passos}
\authornotemark[1]
\email{leandro.passos@unesp.br}
\orcid{0000-0003-3529-3109}
\author{Rafael Gon\c{c}alves Pires}
\email{rafapires@gmail.com}
\orcid{0000-0001-9597-055X}
\author{Daniel Felipe Silva Santos}
\email{danielfssantos1@gmail.com}
\author{Lucas Pascotti Valem}
\email{lucas.valem@unesp.br}
\orcid{0000-0002-3833-9072}
\author{Thierry P. Moreira}
\email{thierrypin@gmail.com}
\orcid{0000-0002-3410-6247}
\author{Marcos Cleison S. Santana}
\email{marcoscleison@gmail.com}
\orcid{0000-0003-2568-8019}
\author{Mateus Roder}
\email{mateus.roder@unesp.br}
\orcid{0000-0002-3112-5290}
\author{Jo\~{a}o Paulo Papa}
\email{joao.papa@unesp.br}
\orcid{0000-0002-6494-7514}
\affiliation{%
  \institution{S\~{a}o Paulo State University - UNESP}
  \streetaddress{Av. Eng. Lu\'{i}s Edmundo Carrijo Coube, 14-01}
  \city{Bauru}
  \state{S\~{a}o Paulo}
  \country{Brazil}
}

\author{Danilo Colombo}
\email{colombo.danilo@petrobras.com.br}
\affiliation{%
  \institution{Cenpes, Petroleo Brasileiro S.A. - Petrobras}
  \city{Rio de Janeiro}
  \state{Rio de Janeiro}
  \country{Brazil}
}

\renewcommand{\shortauthors}{Santos et al.}

\begin{abstract}
In general, biometry-based control systems may not rely on individual expected behavior or cooperation to operate appropriately. Instead, such systems should be aware of malicious procedures for unauthorized access attempts. Some works available in the literature suggest addressing the problem through gait recognition approaches. Such methods aim at identifying human beings through intrinsic perceptible features, despite dressed clothes or accessories. Although the issue denotes a relatively long-time challenge, most of the techniques developed to handle the problem present several drawbacks related to feature extraction and low classification rates, among other issues. However, deep learning-based approaches recently emerged as a robust set of tools to deal with virtually any image and computer-vision related problem, providing paramount results for gait recognition as well. Therefore, this work provides a surveyed compilation of recent works regarding biometric detection through gait recognition with a focus on deep learning approaches, emphasizing their benefits, and exposing their weaknesses. Besides, it also presents categorized and characterized descriptions of the datasets, approaches, and architectures employed to tackle associated constraints. 
\end{abstract}



\maketitle

\section{Introduction}
\label{s.intro}

Gait recognition emerged in the last decades as a branch of biometric identification that focuses on detecting individuals through personal measurements and relationships, e.g., trunk and limbs' size, as well as space-time information related to the intrinsic patterns in individuals movements~\cite{bolle2013guide}. Such an approach presented itself extremely useful in the contexts of surveillance systems or hazy environments monitoring, for instance, where unique elements usually employed for biometric identification such as the fingerprint and the face are hard or impossible to distinguish.

Besides, gait recognition approaches possess some advantages regarding other biometric identification models, since hacking such architectures poses a hard task~\cite{nixon2004advances}. The difficulty primary lies in the concept's intrinsic characteristics, i.e., identification based on the silhouette and its movement, whose reproduction is particularly complicated. The same is not valid for other techniques, where the individual can trick the system by hiding the face, for instance. Further, gait recognition models do not require high-resolution images and specialized equipment for proper identification, like iris and fingerprints, for instance. Furthermore, gait recognition methods are independent of individual cooperation, while other methods require the analyzed person collaborates with the identification system.  

Moreover, gait information is easily collected from distance, which stands for an enormous advantage regarding other techniques, especially when the identification is not assisted by the analyzed person, e.g., criminal investigations. Besides, since it does not require sophisticated equipment for data extraction, these methods are commonly cheaper than other approaches, majorly due to the popularization of surveillance systems and the advent of cellphones equipped with accelerometers, which transformed the burden of extracting data signals into a straightforward task. 

Despite the simplicity regarding the tools mentioned above, identifying people by walking and moving is far from a trivial task. Standard gait recognition methods, i.e., which comprise data pre-processing and features extracted in a handcrafted fashion for further recognition, often suffer from several constraints and challenges imposed by the complexity of the task, such as viewing angle and large intra-class variations, occlusions, shadows, and locating the body segments, among others. A new trend on machine learning, known as deep learning, emerged in the last years as a revolutionary tool to handle topics in image and sound processing, computer vision, and speech, overwhelmingly outperforming virtually any baseline established until then. The new paradigm exempts the necessity of manually extract representative features from experts and also provides paramount results regarding gait recognition, surpassing existing challenges and opening room for further research.

Due to the increasing number of biometric and gait recognition works developed in the last years, many authors provided studies summarizing each field's main achievements. Pisani et al.~\cite{pisani2019adaptive}, Khan et al.~\cite{khan2020biometric}, and Sundararajan et al.~\cite{sundararajan2019survey}, for instance, exposed the main advances in the last years regarding biometrics identification in general, while Gui et al.~\cite{gui2019survey} explored the novelties regarding brain biometrics methods. Regarding gait recognition, both Wan et al.~\cite{wan2018survey}, and Rida et al.~\cite{rida2018robust} provided a comprehensive survey comprising general topics on the context in 2018. Meanwhile, Singh et al.~\cite{singh2018vision} published a study on vision-based gait recognition, while Bouchrika~\cite{bouchrika2018survey} proposed a similar work considering smart visual surveillance. Further in 2019, Nambiar et al.~\cite{nambiar2019gait} summarized the works regarding gait-based person re-identification, while Marsico and Mecca~\cite{marsico2019survey} focused on gait recognition via wearable sensors. However, none of them focus on presenting the main advances regarding deep learning-based approaches for gait recognition. Moreover, even though there are reviews in this context~\cite{alharthi2019deep}, most of the surveyed works comprise Convolutional Neural Networks (CNN)~\cite{lecunLeNet:1998} and Long short-term memory (LSTM)~\cite{Hochreiter,LoboICCS:20}, leaving behind some relevant and novel architectures, such as Autoencoders~\cite{sda}, Capsule Networks (CapsNet)~\cite{sabour2017dynamic}, Deep Belief Networks~\cite{Hinton:06}, and Generative Adversarial Networks (GANs)~\cite{gan}. Therefore, the main objectives of this work are three-fold:

\begin{itemize}
    \item to systematically introduce the most recent and significant works comprising strategies for gait recognition through deep learning approaches; and
    \item to provide the reader with a substantial and illustrated theoretical background regarding gait recognition, exploring its roots on biometric recognition and exposing the most popular tools employed to gait feature extraction and the architectures used to tackle the associated constraints; and
    \item to present an illustrated, categorized, and characterized catalog of the public datasets available for the task of gait recognition.
\end{itemize}

Regarding the selection of the reviewed papers, the keywords employed in the search were ``gait recognition using \emph{deep learning technique}'', such that \emph{deep learning technique} was replaced by the architectures themselves, e.g., convolutional neural networks, LSTM, and so on. Another filter considered was the work's innovation, i.e., papers with similar techniques and architectures were discarded. Finally, the studies considered in this research were published five years ago at most.

The remainder of this work is presented as follows. Section~\ref{s.theoretical} presents a theoretical background regarding the concepts and available methods for biometric detection, bestows an overview regarding the deep learning techniques employed in this work, and revisits the main concepts regarding gait recognition. Further, Section~\ref{sec:deep} introduces the most recent approaches using deep learning for gait recognition. Besides, the section also provides a brief discussion regarding the surveyed works. Datasets employed for the task are presented in Section~\ref{sec:datasets}. Finally, Section~\ref{sec:conclusions} presents the conclusions and future works.

\section{Theoretical Background}
\label{s.theoretical}

This section presents a theoretical background introducing the problem of biometric identification, describing the deep learning approaches employed for gait recognition, and a detailed introduction regarding the problem of gait recognition.

\subsection{Biometric Identification}
\label{ss.biometrics}

The problem of people identification poses a challenging task for humankind since long before the existence of computers, when specialists were responsible for analyzing and comparing documents, signatures, and other features in a handcrafted fashion to present some restricted information or to allow some banking transaction, for instance. The importance of an accurate identification was intensified insofar as the society informatization progressed, leading to the necessity of robust solutions for individuals' recognition. In this context, a wide range of techniques emerged in the literature as alternatives for effectively identifying people through images or another biometric means. 

To be considered a biometric qualification criterion, a candidate feature must meet the following conditions~\cite{jain}:

\begin{itemize}
	\item \textbf{Universality:} each person should have the characteristic;
	\item \textbf{Distinctiveness:} any two persons should be sufficiently different in terms of the characteristic;
	\item \textbf{Permanence:} the characteristic should be sufficiently invariant over a period of time;
	\item \textbf{Collectability:} the characteristic can be measured quantitatively.
\end{itemize}

However, in a practical biometric system, other issues should also be considered, including:

\begin{itemize}
	\item \textbf{Performance:} which refers to the achievable recognition accuracy and speed, the resources required to achieve the desired recognition accuracy and speed, as well as the operational and environmental factors that can affect the accuracy and the speed;
	\item \textbf{Acceptability:} which indicates the extent to which people are willing to accept the use of a particular biometric identifier in their daily lives;
	\item \textbf{Circumvention:} which denotes the easiness the system is fooled through fraudulent methods.
\end{itemize}

Among such techniques, stand out the fingerprint, iris, face, and gait recognition, among others. The next section briefly revisits some of these biometric approaches.

\subsubsection{Fingerprint Recognition}
\label{sss.fingerprint}

Formally known as dactyloscopy, fingerprint recognition systems are widely used due to the singularity of the ridges and furrows on the surface of a finger, which provides an intrinsic characteristic for each individual~\cite{li2009encyclopedia}. Moreover, such features are steady and poorly degraded over time, making the creation of a digital fingerprint image database extremely reliable. 

The first model for fingerprint recognition was designed in the late 1960s, based on a system created in the XIX century by Francis Galton called Galton points~\cite{ainsworthamitchell1920detection}. Since then, many works addressed the problem through different perspectives, such as digital image processing~\cite{alsmirat2019impact, zneit2017methodology}, Generative Adversarial Networks~ \cite{yu2019attributing}, and filter representation~\cite{lee2017partial}, to cite a few.

Fingerprint recognition systems are considered the most reliable and accurate biometric identification system. Nevertheless, the field is still facing several challenges, such as unsatisfactory accuracy under non-ideal conditions and security issues such as spoofing attacks. In this context, Hemalatha et al.~\cite{hemalatha2020systematic} present a systematic review comprising the most recent techniques employed to deal with the drawbacks mentioned above and others.

\subsubsection{Iris Recognition}
\label{sss.iris}

The term iris denotes a colorful thin circular structure in the eye responsible for controlling the pupil diameter, as well as the amount of light that reaches the retina. Such a structure presents a great advantage for people's identification since it is consistent against alterations on environmental conditions and generally suffers low degradation over time. Moreover, iris recognition is among the most accurate, low-cost, and convenient identification methods, since it is performed by image and do not require contact with the person~\cite{wang2019cross}.

Most commercial models employing iris recognition are developed based on the identification of iris' lower and upper limits using an integral-differential operator~\cite{karn2020experimental}, even when using eyelid demarcation. Such an operator assumes the pupil is circular and acts as an orbicular border detector. Further researches introduced distinct mathematical operations to the process to make it more flexible and robust,  such as parabolic curve identifiers and normalizations, to cite a few.

Moreover, a different approach was proposed by Garagad and Iyer~\cite{garagad2014novel}, which considers a scale- and slope-invariant method that radially draws the iris region for feature extraction. The method segments the pupil's central area using concentric circles and discards irrelevant regions using discontinuity detection techniques. Besides, the method is straightforward if compared to the approach mentioned above since it does not use integral-differential operators, even though its effectiveness is confirmed by hit rates above $83\%$. 

Recent studies toward iris recognition aim to overcome some challenges in the field, such as the adverse noise caused by specular reflections, the absence of iris, gaze deviation, motion/defocus blur, iris rotation, and occlusions due to hair/eyelid/glasses/eyelash~\cite{wang2020towards}. Besides, He et al.~\cite{he2020long} discuss how new trends towards long-distance iris recognition have been tackled recently and the future trends to deal with the problem.

\subsubsection{Face Recognition}
\label{sss.face}

Face recognition is a prominent biometric system extensively used in identification and authentication systems in the most diverse areas, like banks, military services, and public security~\cite{zou2018}. Such methods became popular in the earlies 90's when Turk and Pentland proposed the Eigenfaces method~\cite{turk1991eigenfaces}. In the following decade emerged several approaches so-called holistic, which are derived from low-dimensional distribution representations, like linear sub-spaces~\cite{deng2014transform}, and sparse reproduction~\cite{deng2018face}, among others.

Nevertheless, the major drawback regarding holistic methods lies is the intolerance to face positioning change, thus leading to a search for characteristics based methods. In this context, emerged the Local Binary Patters (LBP) and the Gabor feature based classification, achieving relevant results through filters embodying invariant properties. Later on, subsequent works started to focus on local descriptors-based approaches for face recognition~\cite{lei2014learning}. In short, the objective of these methods is to train spatial filters for image feature extraction, such that the difference between images from the same person is minimized. On the other hand, the distance between features extracted from different people should be maximized. Even though such methods obtained consistent results, techniques considered ``shallow'', in general, suffer from inevitable limitations due to the non-linear complexity of the face variations.   

A few years ago, a counterpoint for such approaches arose after the AlexNet neural network had won the ImageNet~\cite{krizhevsky2012imagenet} competition. Since then, deep learning models gained attention in diverse areas of biometrics, including face recognition. A particular family of networks known as Convolutional Neural Networks was the first to achieve a level of recognition near humans, i.e., the DeepFace~\cite{taigman2014deepface}. The model composed of six convolutional layers achieved accuracies less than $0.2\%$ different from one human competitor ($97.35\%$ against $97.53\%$). Nowadays, the model is considered relatively simple when compared to popular architectures, like Inception~\cite{szegedy2016rethinking}.

Even though face recognition approaches evolved to unprecedented standards in the last years, the field is still facing several challenges in terms of expression, pose, illumination, aging, and face partially or entirely obstructed by masks or veils. Recent research presented by Tiong et al.~\cite{tiong2020multimodal} suggest using multimodal biometrics to tackle the issue. Besides, Jaraman et al.~\cite{jayaraman2020recent} describe the most relevant alternatives freshly employed for face recognition in the last years, which comprise local and global features, neural networks, and deep learning-based approaches.

\subsubsection{Multimodal Biometric Recognition}
\label{sss.multimodal}

One of the main advantages of the human brain towards individual recognition over computer-based approaches concerns the ability to simultaneously evaluate multiple modalities of descriptive information, such as face, gait, and hair and eyes color. To adapt to this trend, a new paradigm of biometrics recognition emerged, the so-called ``multimodal biometrics''. Such an approach aims at combining distinct biometric recognition methods and auxiliary information to improve the performance and reliability when considering a single technique. In this context, Sultana et al.~\cite{sultana2017social} proposed a person recognition approach based on a combination of visual cues, such as face and ear, with the person's social behavior. Experiments over semi-real datasets provided a significant advantage over standard biometric systems.

A recent work proposed by Li et al.~\cite{li2021joint} implements a finger-based multimodal biometrics system that considers a fusion strategy to extract correlated features from different modalities of finger patterns, i.e., finger veins and knuckle print, obtaining state-of-the-art results over finger recognition methods. A similar work proposed by Tiong et al.~\cite{tiong2020multimodal} implements a multi-feature fusion convolutional neural network for multimodal facial biometrics, obtaining competing results over several benchmarking datasets.

The major advances of multimodal biometrics recognition are described in a systematic review presented by Dargan et al.~\cite{dargan2020comprehensive}. The authors provide a literature review concerning unimodal and multimodal biometric systems, datasets, feature extraction techniques, classifiers, results, efficiency, and reliability. Besides, they also discuss how multimodal biometric approaches can be employed to overcome some common challenges faced by unimodal methods, such as face, fingerprint, and iris recognition.

\subsubsection{Other Identifiers}
\label{sss.otherIdentifiers}

For different reasons, the traits mentioned so far may not be interesting in certain environments, either due to the imprecision of the model or the lack of specialized equipment for the correct identification. Among some unusual approaches, Ragan et al.~\cite{ragan2016ear} showed that ear-based biometrics has intrinsic characteristics capable of holding unique information from individuals. Applications can be observed in smart-phone-based applications\footnote{https://www.popsci.com/article/technology-tested-app-authenticates-you-shape-your-ear}. Meanwhile, Ramli et al.~\cite{ramli2016development} employed the heartbeat rate pattern for biometric identification. The method obtained significant accuracy and also has been employed for unlocking electronic devices\footnote{https://www.extremetech.com/computing/165537-nymi-wristband-turns-your-heartbeat-into-an-electronic-key-that-unlocks-your-devices}. Similarly, Chen et al.~\cite{chen2020integrating} offered a Raspberry Pi-based system for gesture recognition on an Internet of Things environment. The same group of authors~\cite{dewi2019human} presented a study comparing several machine learning approaches for human activity recognition.

As described in Section~\ref{sss.iris}, the characteristics of the eye present powerful descriptive biometric features, such that iris recognition plays the leading role in this context. However, some distinct approaches have been successfully employed for the task. The Research developed by Ma et al.~\cite{ma2019emir}, for instance, showed the eyes' movement pattern while following objects is unique in each person and employed such a feature for portable remote authentication smart grids. A study proposed by Zehngut et al.~\cite{zehngut2015investigating} has shown that it is possible to divide the types of noses between six different groups and, combined with specific proportions of the width and height of the nose, enable the identification of individuals with accuracy as good as recognition by iris. Among the method's advantages, one can highlight its visibility in almost $100\%$ of cases, even when people use glasses or hats. The major disadvantage regards plastic surgery, changing features relevant to the correct identification of the individual.

The main drawback regarding image-based biometric systems, e.g., fingerprints or irises methods, lies in the susceptibility to malicious attempts to fool the system. In short, a simple image could fool such systems, granting access to someone else's information in a bank account, for instance. On the other hand, some more sophisticated methods can identify people by fancier characteristics, such as the veins configuration~\cite{alariki2018review}. The method is more robust against fraud than simple image-based techniques since it demands specialized equipment for proper identification.

\subsection{Deep Learning Approaches Considered for Gait Recognition}
\label{ss.dl_works}

Even though standard machine learning and gait recognition strategies provided relatively satisfactory results in the past years, such methods are usually constrained to hand-crafted features and limited capacity for learning intrinsic patterns in data. In this context, deep learning-based approaches emerged as an elegant solution for tackling image/video and sequential problems, among others, also revealing themselves as a powerful tool for gait recognition. Thus, this section presents the most popular deep learning architectures once employed for gait recognition, namely Convolutional Neural Networks~\cite{lecunLeNet:1998}, Recurrent Neural Networks~\cite{8963659}, Autoencoders~\cite{babaee2019person}, Capsule Networks~\cite{sabour2017dynamic}, Generative Adversarial Networks~\cite{gan}, and Deep Belief Networks~\cite{Hinton:06}.

\subsubsection{Convolutional Neural Networks}
\label{sss.cnn}

\begin{figure}[htp]
\begin{tikzpicture}[scale=1.0]

  \matrix [nodes=draw,column sep=-0.2mm, minimum size=6mm]
  {
    \node {0}; & \node{1}; & \node {1}; & \node{$1_{\times 1}$}; & \node{$0_{\times 0}$}; 
    & \node{$0_{\times 1}$}; & \node{0}; \\
    \node {0}; & \node{0}; & \node {1}; & \node{$1_{\times 0}$}; & \node{$1_{\times 1}$}; 
    & \node{$0_{\times 0}$}; & \node{0}; \\
    \node {0}; & \node{0}; & \node {1}; & \node{$1_{\times 1}$}; & \node{$1_{\times 0}$}; 
    & \node{$1_{\times 1}$}; & \node{0}; \\
    \node {0}; & \node{0}; & \node {1}; & \node{\, 1 \,}; & \node{\, 1 \, }; 
    & \node{\, 0 \,}; & \node{0}; \\
    \node {0}; & \node{0}; & \node {1}; & \node{\, 0 \, }; & \node{\, 0 \, }; 
    & \node{\, 0 \,}; & \node{0}; \\
    \node {0}; & \node{1}; & \node {0}; & \node{\, 0 \, }; & \node{\, 0 \, }; 
    & \node{\, 0 \,}; & \node{0}; \\
    \node {1}; & \node{1}; & \node {0}; & \node{\, 0 \,}; & \node{\, 0 \, }; 
    & \node{\, 0 \,}; & \node{0}; \\
  };

  \coordinate (A) at (-0.6,0.3);
  \coordinate (B) at (1.78,0.3);
  \coordinate (C) at (1.78,2.12);
  \coordinate (D) at (-0.6,2.12);
  \fill[red, opacity=0.3] (A)--(B)--(C)--(D)--cycle;
  \begin{scope}[shift={(3.3,0)}]
    \node[] at (0,0) {\Large $\ast$};
  \end{scope}[shift={(2.5,0)}]

  \begin{scope}[shift={(5,0)}]

    \matrix [nodes=draw,column sep=-0.2mm, minimum size=6mm]
    {
      \node{1};  & \node{0};   & \node{1};  \\
      \node{0};  & \node{1};   & \node{0};  \\
      \node{1}; & \node{0}; & \node{1}; \\
    };
    \coordinate (A1) at (-0.9,-0.9);
    \coordinate (B1) at (0.93,-0.9);
    \coordinate (C1) at (0.93,0.92);
    \coordinate (D1) at (-0.9,0.92);
    \fill[blue, opacity=0.2] (A1)--(B1)--(C1)--(D1)--cycle;
    \draw[blue, line width=2] (A1)--(B1)--(C1)--(D1)--cycle;
  \end{scope}

  \draw[dotted, line width=1, color=blue] (A)--(A1);
  \draw[dotted, line width=1, color=blue] (B)--(B1);
  \draw[dotted, line width=1, color=blue] (C)--(C1);
  \draw[dotted, line width=1, color=blue] (D)--(D1);

  \begin{scope}[shift={(6.6,0)}]
    \node[] at (0,0) {\Large $=$};
  \end{scope}[shift={(2.5,0)}]

  \begin{scope}[shift={(9,0)}]

    \matrix [nodes=draw,column sep=-0.2mm, minimum size=6mm]
    {
      \node{1};  & \node{4};   & \node{3}; & \node{4}; & \node{1};  \\
      \node{l};  & \node{2};   & \node{4}; & \node{3}; & \node{3};  \\
      \node{1}; & \node{2}; & \node{3}; & \node{4} ; & \node{1};  \\
      \node{1}; & \node{3}; & \node{3}; & \node{1} ; & \node{1};  \\
      \node{3}; & \node{3}; & \node{1}; & \node{1} ; & \node{0};  \\
    };
    \coordinate (A2) at (0.3,0.9);
    \coordinate (B2) at (0.91,0.9);
    \coordinate (C2) at (0.91,1.507);
    \coordinate (D2) at (0.3,1.507);
    \fill[green, opacity=0.2] (A2)--(B2)--(C2)--(D2)--cycle;
    \draw[green, line width=2] (A2)--(B2)--(C2)--(D2)--cycle;
  \end{scope}

  \draw[dotted, line width=1, color=green] (A1)--(A2);
  \draw[dotted, line width=1, color=green] (B1)--(B2);
  \draw[dotted, line width=1, color=green] (C1)--(C2);
  \draw[dotted, line width=1, color=green] (D1)--(D2);
\end{tikzpicture}
\caption{An example of $3\times3$ convolution.}
\label{fig:convolution}
\end{figure}
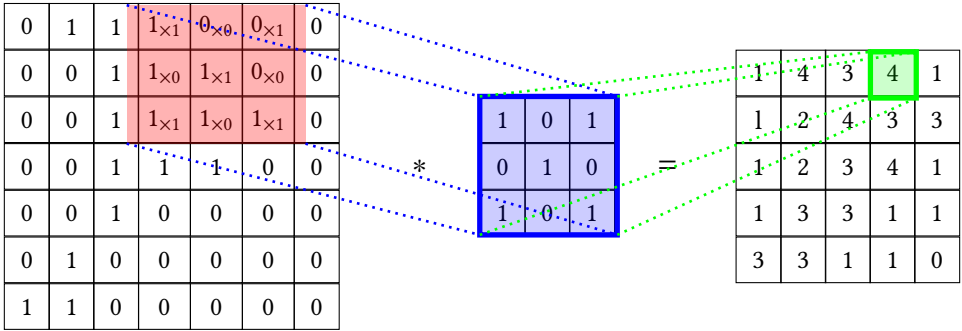

Convolutional Neural Networks~\cite{lecunLeNet:1998} obtained exceptional popularity since the beginning of the 2010s, becoming paramount to solve image processing-related problems, such as image classification~\cite{efficientnet} and segmentation~\cite{sota-segmentation}. As the name suggests, CNNs basic blocks are composed of convolutional neurons, usually composed of $3\times3$ or $5\times5$ kernels, which are responsible for performing convolutional operations over the input data.  Briefly speaking, this process's output generates a new set of matrices, which are then employed as the subsequent layers of the model. In signal processing, convolution is described as a  multiplication of two signals to generate a third one~\cite{oppenheim}. Figure~\ref{fig:convolution} depicts the idea. 

\begin{figure} [H]	
	\includegraphics[width=0.7\textwidth]{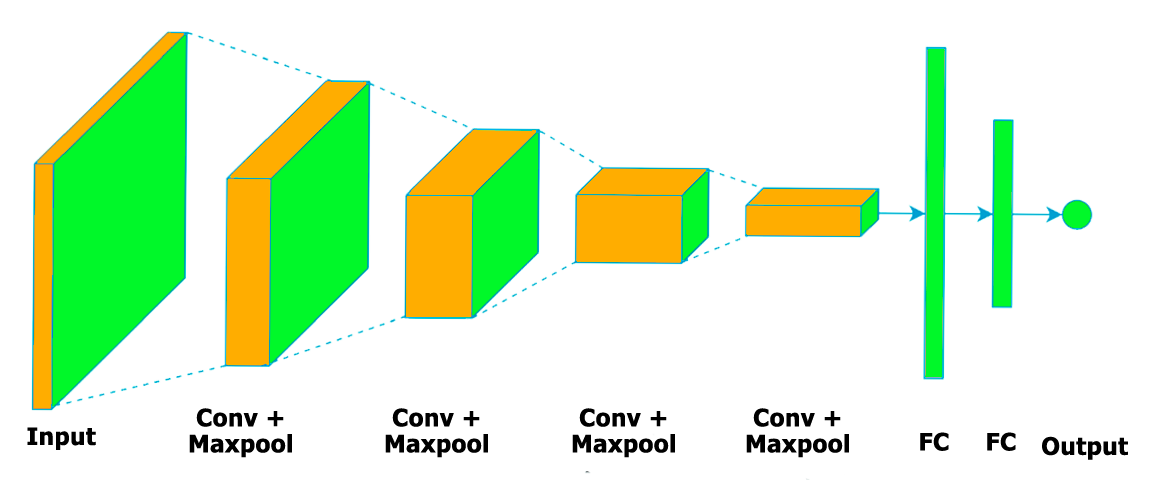}
	\caption{A standard CNN architecture's example.}
	\label{fig:cnn}
\end{figure}

Therefore, a CNN can be interpreted as a stack of convolutional kernels generating the successive layers' input. Notice some intermediate pooling layers may be included in the process, and a fully connected layer is coupled at the top of the architecture for classification purposes. The learning process is performed by the backpropagation algorithm working with a gradient descent calculation. Figure~\ref{fig:cnn}, inspired in LeNet-5, pictures a CNN architecture.

\subsubsection{Capsule Networks}
\label{sss.capsule}

Even though CNNs work very well for image features' understanding, they are prone to confuse spatial relationships between complex peculiarities. In other words,  a trained CNN is usually capable of recognizing dogs if it finds a dog's body, face, tail, and so on, even if the dog is assembled in a different sequence or if its parts are located in distinct sections of the image. On the other hand, Capsule Networks~\cite{sabour2017dynamic} consider a hierarchical approach to tackle this problem. In short, the model comprises a two-layer structure. The first is a convolutional encoder, which performs the recognition of low-level features. The second stands for a fully-connected linear decoder that employs the routing by agreement algorithm~\cite{sabour2017dynamic} to address such low-level features to the correct position in a hierarchical higher-level. Therefore, CapsNets are more robust to object orientation. Besides, they may perform better for identifying multiple or overlapping objects in a scene.

\subsubsection{Recurrent Neural Networks}
\label{sss.rnn}

Many works addressed the gait recognition problem as a sequence of images defining the individual's movement. A standard method to acknowledge such strategy using deep learning concerns Recurrent Neural Networks (RNNs)~\cite{8963659}, which compute each neuron's activation considering the information from the input data, as well as other neuron's output, in a recurrent fashion. Occasionally, the architecture is combined with a CNN to extract more information about the input images to perform the inference.

Since describing a person's gait usually requires a considerable amount of sequential features, a particular set of RNNs, namely gated RNN, are more suitable for the task due to their abilities to deal with long sequences. In this context, one can refer to two main architectures, i.e., the Long-Short Term Memory~\cite{Hochreiter} and the Gated Recurrent Unit (GRU)~\cite{gru-study}.

\begin{enumerate}
\item \textbf{Long-Short Term Memory:}

The Long-Short Term Memory was first implemented in 1997 by Hochreiter and Schmidhuber~\cite{Hochreiter}, where the main objective was to improve results on long sequences of data. In a nutshell, LSTMs work similarly to the traditional RNN, i.e., the output of a given neuron depends on recurrent information from previous neurons' outcomes. The main difference lies in the LSTM cell's architecture, which comprises more complex and fancy relationships. Such an architecture is composed of three main gates that control the flow of information, described as follows:

\begin{itemize}
\item The forget gate: this gate defines how much of the information should be kept. The previous and current state data is passed through a sigmoid function, which outputs values between $0$ and $1$. The closer to $1$, the more information is preserved.

\item The input gate: computes a new value to update the current hidden state. The input gate considers two central values: (i) a sigmoid function calculates the importance of the previously hidden state, and (ii) the original value is forwarded to a hyperbolic tangent (tanh) function, which is responsible for squishing this value between $-1$ and $1$. The multiplication of these two values defines the current hidden state.

\item The output gate: after estimating the forget and the input gates, the output gate defines the cell's output value. The process is performed as follows: (i) the values from the forget and the input gate are summed up and submitted to a tahn function, (ii) the cell's previous state is submitted to a sigmoid function, (iii) the output from both the sigmoid and the tahn functions are multiplied, yilding the cell's output.
\end{itemize}

\item \textbf{Gated Recurrent Unity:}

The Gated Recurrent Unit~\cite{gru} is a recurrent neural network originally idealized to improve results on neural machine translations. Like LSTMs,  GRUs possess internal gates that control the flow of information, and the main difference lies in the number of gates available in each model, i.e., GRU comprises only two, namely the forget and the output gates, instead of three. Studies show that, even though the GRU uses fewer gates than LSTM, it can reach similar results~\cite{gru-study}, with the advantage of a reduced computational burden and faster performance considering both tasks of training and inference.

\end{enumerate}

\subsubsection{Autoencoders}
\label{sss.autoencoder}

\begin{figure} [H]	
	\includegraphics[width=0.7 \textwidth]{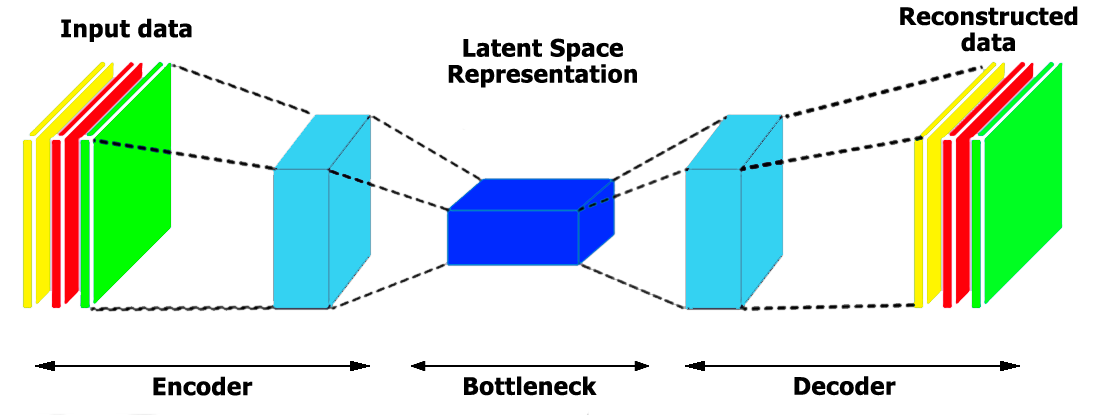}
	\caption{A standard Autoencoder architecture.}
	\label{fig:autoencoder}
\end{figure}

Autoencoders~\cite{sda} are generative neural networks usually employed for data reduction and image denoising. The model comprises two main steps, described as follows:

\begin{itemize}
\item Encoder: it is responsible for encoding the input information into an, usually, smaller feature space.
\item Decoder: it performs the unsupervised reconstruction of encoded data.
\end{itemize}
Figure~\ref{fig:autoencoder} depicts the architecture of the model.

\subsubsection{Deep Belief Networks}
\label{sss.dbn}

Deep Belief Networks~\cite{Hinton:06} are stochastic neural networks ideally designed for generative tasks, such that each layer denotes a greedy-fashioned trained Restricted Boltzmann Machine (RBM)~\cite{Hinton:06}. In short, RBM is a graphical model composed of a visible and a latent set of units, namely visible and hidden layers, respectively, which are connected by a weight matrix with no connection among neurons from the same layer. The model's learning approach consists of attaching a collection of input data into the visible layer and finding a representation of this data in the hidden units. Besides, the training procedure is performed by minimizing the system's energy, usually conducted using a Markov chain procedure through Gibbs sampling for optimization purposes.

Concerning classification tasks, the most common approach considers coupling a softmax layer at the top of the DBN architecture and, after greedily pre-training all the RBMs, performing a fine-tuning in the weights using Backpropagation, for instance, to adjust the weight matrices and fitting the labels for proper identification.

\subsubsection{Generative Adversarial Networks}
\label{sss.gan}

\begin{figure} [H]	
	\includegraphics[width=0.7 \textwidth]{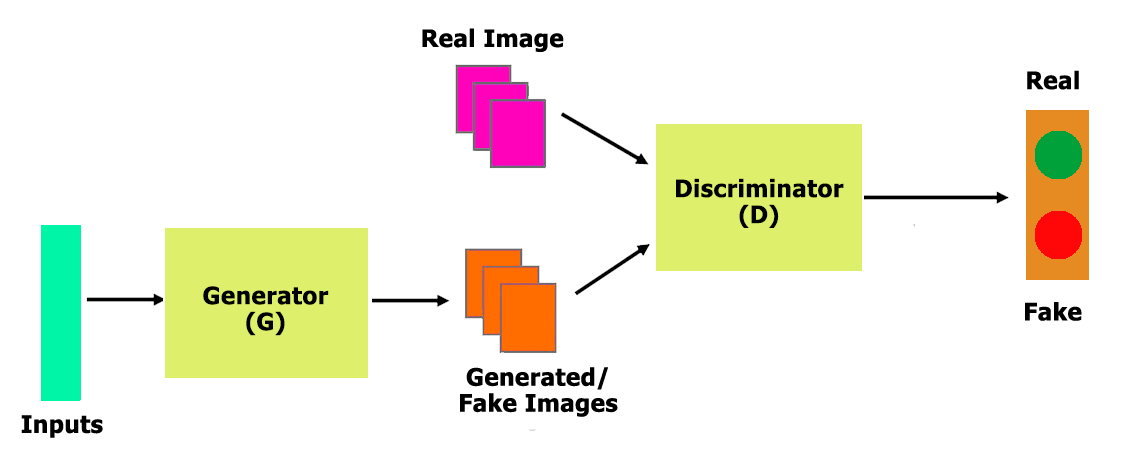}
	\caption{A standard GAN architecture.}
	\label{fig:gan}
\end{figure}

Generative Adversarial Networks~\cite{gan} became popular in the last years due to their outstanding ability to generate realistic synthetic images. The model comprises two distinct networks, i.e., a generator, which is responsible for learning the data's distribution and generating synthetic samples, and a discriminator, which tries to identify whether a given instance is original or synthetically created. The generator and the discriminator compete in an adversarial fashion such that the generator attempts to generate samples realistic enough to fool the discriminator. In contrast, the discriminator improves itself more and more to recognize such fake images. Figure~\ref{fig:gan} illustrates the model.

\subsubsection{Deep learning techniques' summary}

Table~\ref{t.architectures} presents a summary concerning the deep learning techniques considered in this work and how they are usually employed for gait recognition tasks. Notice that the same work can eventually use distinct architectures, thus appearing more than once in the table.

\begin{center}
\begin{longtable}{| p{.15\textwidth} | p{.33\textwidth} |p{.35\textwidth} |}
\caption{Deep learning-based gait recognition approaches organized byt type of neural network.} \\

\hline  \multicolumn{1}{|p{.15\textwidth}|}{\textbf{References}}&  \multicolumn{1}{p{.33\textwidth}|}{\textbf{Technique}}&  \multicolumn{1}{p{.35\textwidth}|}{\textbf{Task}}\\ \hline 
\endfirsthead

\multicolumn{3}{c}%
{{\bfseries \tablename\ \thetable{} -- continued from previous page}} \\
\hline \multicolumn{1}{|p{.15\textwidth}|}{\textbf{References}}&  \multicolumn{1}{p{.33\textwidth}|}{\textbf{Technique}}&  \multicolumn{1}{p{.35\textwidth}|}{\textbf{Task}} \\ \hline 
\endhead

\hline \multicolumn{3}{|r|}{{Continued on next page}} \\ \hline
\endfoot

\hline \hline
\endlastfoot

\cite{shiraga2016geinet,wu2017comprehensive,Li:2017Deepgait, sokolova2017gait, takemura2017input,sokolova2017posed,yu2017invariant,zou2018,wang2020human,xu2020cross}  &Convolutional Neural Networks (CNN) & Feature extraction from images or frames of a video.\\ \hline

\cite{zhang2019gait,babaee2019person}   & Auto Encoder & Works by compressing and decompressing features from the input.\\ \hline

\cite{xu2019gait,sepas2021gait,zhao2021associated} & Capsule   & Improve the semantic organization  of the outputs from a CNN.\\ \hline

\cite{fernandes2018artificial,xiong2020continuous} & Deep Belief Networks & Encode features and patterns into compressed representations. \\ \hline

\cite{he2018multi,jia2019attacking,hu2018robust}  & Generative Adversarial Networks  & A training method that relies on the differentiaton of an original input and a generate counterpart from a model, such as a CNN.\\ \hline

\cite{zou2018,zhang2019gait,potluri2019deep,wang2020human,tran2021multi}  & Recurrent Neural Networks & Comprise both GRUs and LSTMs, which are composed with several gates to controle the flow of information and are employed to deal with temporal information.

\label{t.architectures}
\end{longtable}
\end{center}

\subsection{Gait Recognition}
\label{ss.gait}

Several methods for people recognition through biological characteristics have been presented so far. Although the methods' reliability and safety are confirmed by their success in banks and public governance systems, two main hindrances must be stressed: (i) they depend on the passive provision of personal biometric information, i.e., the person must provide or register the required information for the recognition; and (ii) some of these systems rely on specialized equipment. 

An alternative to deal with such drawbacks may comprise gait recognition models, especially considering video-based approaches. Such methods do not suffer from the problems mentioned above, once the acquisition of the biometric information depends, most of the time, only on a camera with no specific features or non-evasive sensors and devices, and, disregarding legal problems, the collection of such information is performed passively. Therefore, the observed person is no longer required to cooperating in the identification process or providing any information. On the other hand, although video-based recognition is the most intuitive method, it does not stand for the only possibility. Gait identification can also be performed by distinct characteristics, e.g., the footprint~\cite{costilla2018analysis}, which considers the pressure and size of the area. The obvious disadvantage of such models is the cost of deploying the equipment necessary to acquire such features.

Moreover, techniques proposed for gait recognition can be divided into two main groups, i.e., template- and non-template-based methods. Template-based methods aim to obtain the movement of the trunk or legs, i.e., they usually focus on the dynamics of movement through space or spatio-temporal based methods~\cite{yeo2020accuracy}. Among such techniques, one can refer to Walking Path Image (WPI) information~\cite{Zhao}, Gait Information Image (GII)~\cite{arora}, and Gait Energy Image (GEI) features, which can be extracted through Canonical Correlation Analysis~\cite{luo2020multi}, Joint Sparsity Models~\cite{Yogarajah}, segmentation using Group Lasso Motion~\cite{Rida}, among others~\cite{ shiraga2016geinet, Li:2017Deepgait, Wu}. On the other hand, non-template based methods consider the shape and its attributes as the more relevant characteristics, i.e., the recognition of the individual is performed with measurements that reflect its shape~\cite{deng}.

\begin{figure}[htp]	
	\centering
	\includegraphics[width=.7\textwidth]{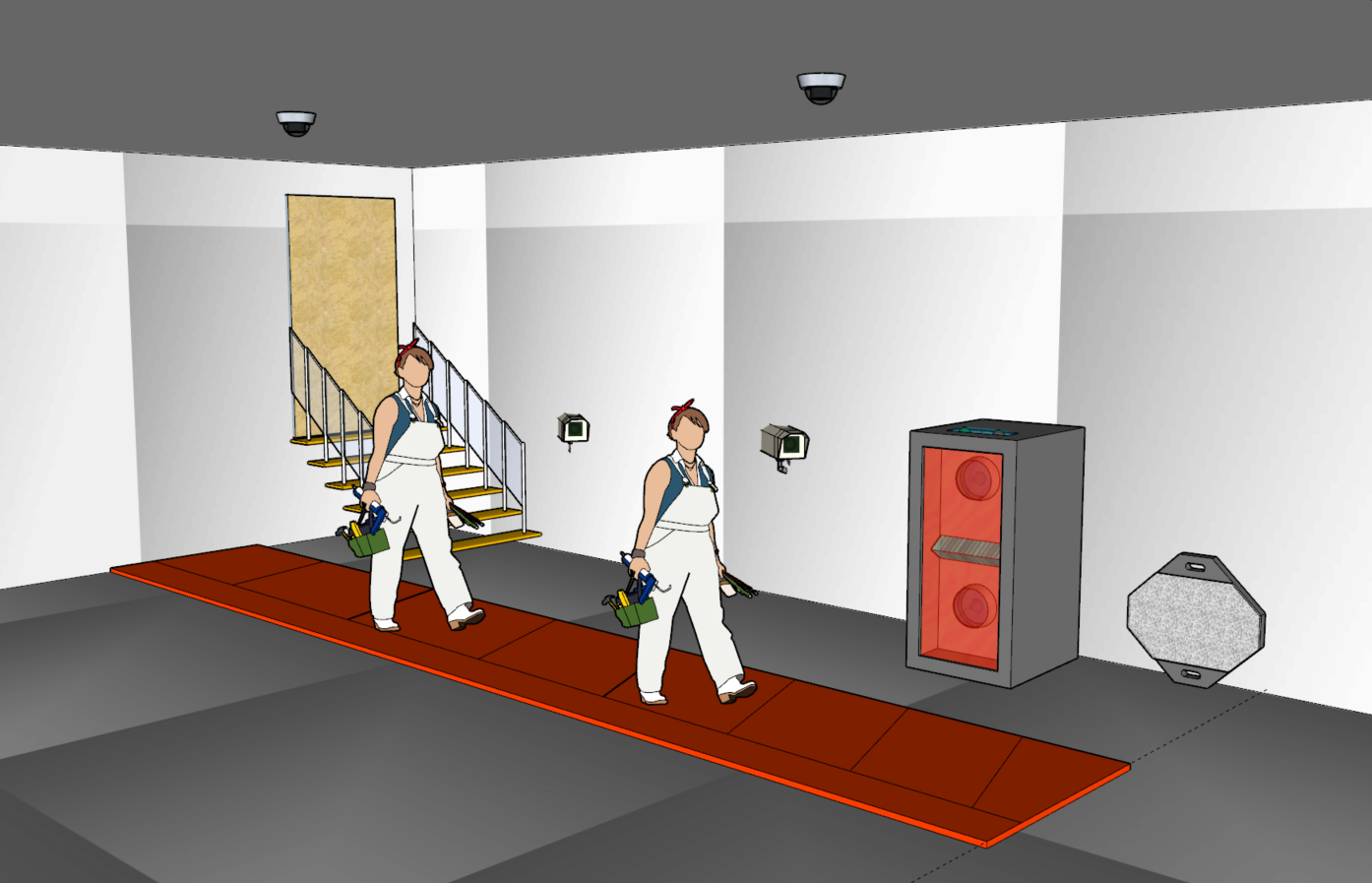}\hfill
	\caption{Example of gait data acquisition environment. The scene comprises the following devices: sidewall cameras, followed by a multi-spectrum photographic sensor. The roof accommodates velocity sensors and gyroscopes, which measure the space-time relationship during the walk. Finally, the floor is equipped with a carpet-like sensor for collecting the foot pressure on the ground, completing the gait cycle.}
	\label{fig:sensors}
\end{figure}

Regarding the process of gait information acquisition, it may comprise several sorts of sensors and devices, as illustrated in Figure~\ref{fig:sensors}. The image depicts a laboratory-like environment designed with several embedded devices for gait data acquisition, such as cameras, which are the most commonly employed device for the task since they are quite accessible and are capable of acquiring data over greater distances, and multispectral photographic sensors that are very efficient, but with less accessible due to high costs. The image also comprises gyroscope and velocity sensors on the roof, responsible for describing the spatio-temporal context acquisition. Finally, the walking rhythm is measured using a step-pressure carpet-like sensor, which is highly valuable to detecte gait abnormalities, such as the ones caused by Parkinson's disease and other neurodegenerative problems.

\begin{figure}[!htb]
    \centering
    \includegraphics[width=.8\textwidth]{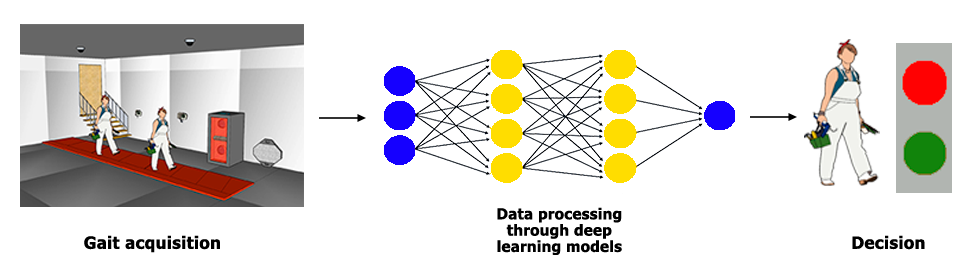}
    \includegraphics[width=.8\textwidth]{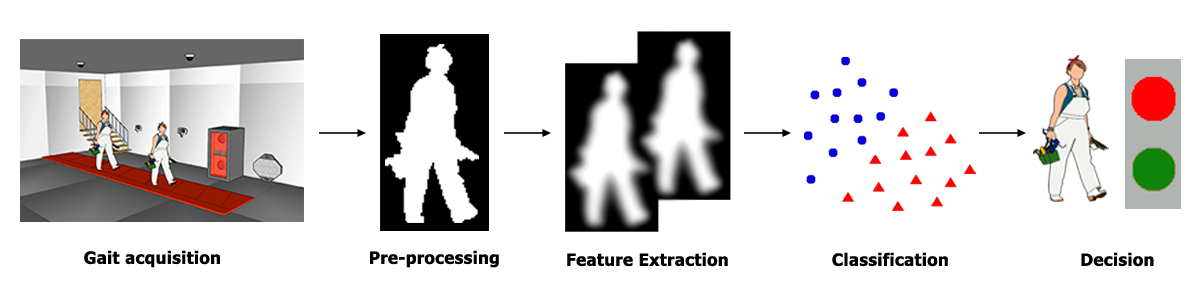}
    \caption{Schematic diagrams from deep learning (top) and standard (bottom) gait recognition pipelines. As the image suggests, deep learning approaches abstract the steps concerning data pre-processing, feature extraction, and classification in a single architecture.}
    \label{fig:diagram}
\end{figure}

Despite the success obtained by the methods mentioned above, deep learning-based approaches induced a paradigm shift in the field of gait recognition, obtaining paramount results in a variety of applications. Therefore, the next section introduces an in-depth presentation of several surveyed works in the context of deep learning-based approaches for gait recognition. Figure~\ref{fig:diagram} provides a schematic diagram comprising an overview of the main differences between deep learning and the standard methods for gait recognition.

\section{Gait Recognition Through Deep Learning-Based Approaches}
\label{sec:deep}

This section presents a systematic review of recent works using distinct deep learning architectures, i.e., Convolutional Neural Networks, Recurrent Neural Networks, Generative Adversarial Networks, Deep Belief Networks, and Autoencoder-based approaches for gait recognition. Besides, it also presents a discussion regarding the compared methods and a summary of the surveyed studies.

\subsection{Convolutional Neural Networks}
\label{ss.cnn}

Convolutional Neural Networks~\cite{lecunLeNet:1998} use a concept of a neuron based on the visual cortex of mammals, which was first validated for the task of digit classification and today is probably the most widely employed neural network for classification, reconstruction, and object detection, among others. In the context of gait recognition, Shiraga et al.~\cite {shiraga2016geinet} used an architecture similar to LENet~\cite{lecunLeNet:1998} to create a gleaming-based recognizer through Gait Energy Image (GEI). The model achieved an accuracy of $91.5\%$ in the OU-ISIR~\cite{Makihara_CVATN2012} Large Population dataset. A similar work proposed by Wang et al.~\cite{wang2019nonstandard} employ nonstandard periodic GEI approaches for gait recognition and data augmentation. 

Regarding cross-view gait-based human identification, Wu et al.~\cite{wu2017comprehensive} claim to propose the first work using CNNs in the context of grimace recognition. The authors employed several network architectures and demonstrated the power of CNNs in the context of biometric-based identification, obtaining gains above $10\%$ to previous results. Besides, they also compared three types of data arrangement, described as follows:

\begin{itemize}
	\item \textit{Local Bottom}: a combination is made among the input data, and then it is determined whether the inputs belong to the same or different individuals;
	\item \textit{Mid-Level Top}: the neural network extracts some characteristics of both inputs before combining them and then determining whether or not they come from the same person; and
	\item \textit{Global Top}: similar to the previous network. However, it has an extra level of convolutions and \textit{Perceptrons}, such that the combination of characteristics is made in the penultimate layer.
\end{itemize}

Further, Li et al.~\cite{Li:2017Deepgait} proposed the DeepGait, a model that combines deep convolutional features and Joint Bayesian for video sensor-based gait representation. The model outperformed hand-crafted features, such as GEI, Frequency-Domain Feature, and Gait Flow Image, obtaining state-of-the-art results over OU-ISR large population dataset~\cite{Makihara_CVATN2012}. Meanwhile, a study proposed by Sokolova and Konushin~\cite{sokolova2017gait} demonstrated the difficulty in identifying people by their behavior due to the intersection of information, despite satisfactory results over TUM-GAID database~\cite{hofmann2014tum}, obtaining $ 97.5\% $ accuracy and $ 99.89\% $ Rank-5. However, the performance was severely degraded concerning CASIA B dataset~\cite{casiab}, bringing an accuracy of $58.20\%$ over such a scenario, demonstrating how this type of biometry is sensitive to the data's context.

The same authors proposed a multiple-stage model for gait recognition using Optical Flow (OF)~\cite{sokolova2017posed}. In this model, the data is pre-processed in two main steps: motion map computation and frame-by-frame evaluation of the individual's pose. Further, two distinct neural networks, i.e., VGG-19~\cite{simonyan2014very} and Wide Residual Network (WRN)~\cite{zagoruyko2016wide}, are employed to validate the technique in the context of video-based gait recognition. To increase classification speed, they incorporated the Principal Component Analysis (PCA) for dimensionality reduction. Such information goes through the normalization \textit{L2}, and finally, the features produced are fed to a Nearest Neighbor (NN) classifier. Using this combination of methods, they achieved very accurate results over TUM-GAID~\cite{hofmann2014tum}, CASIA B~\cite{casiab}, and OU-ISIR~\cite{Makihara_CVATN2012} datasets.

Recently, an architecture called Siamese Networks, which accepts two separate entries as input and computes a similarity value between them, was successfully employed for similar tasks, such as identifying obstructed routes or misbehavior in hazardous environments~\cite{santanaIEEE-IS:19}. In this context, Takemura et al.~\cite{takemura2017input} implemented four different frameworks to create a biometric identification system. All networks rely on GEI to make the comparisons. Two of these networks were developed based on the Triplet Ranking Loss, which necessarily needs three inputs for execution: (i) the information of the compared person, (ii) data from the same person, and (iii) an entrance from any other person. The other two networks use contrastive loss in their execution, whose entries depend only on the person's information for comparison. The authors obtained $ 91.9 \% $ of accuracy over the OU-ISIR Multi-view Large Population (OU-MVLP)~\cite{takemura2018multi} database.

Still, Xu et al.~\cite{xu2020cross} proposed the Pairwise Spatial Transformer Network, a unified model composed of pairwise spatial transformers and a recognition network for cross-view gait recognition. The model computes non-rigid deformation fields to match input pairs into intermediate frames, which are compared against a deformation suffered from source view to target. Experiments conducted over the OU-MVLP~\cite{takemura2018multi}, OU-ISIR Large Population (OU-LP)~\cite{Iwama_IFS2012}, and CASIA B~\cite{casiab} datasets grant the model robustness.

\subsection{Capsule Networks}
\label{ss.capsule}

Another well-known type of deep architecture employed for gait recognition is the Capsule Neural Network~\cite{sabour2017dynamic}. The network has been developed for image classification by modeling the hierarchical relationships between objects, i.e., capsules, in a scene. In this context, Xu et al.~\cite{xu2019gait} explored such features to propose two architectures for gait recognition:

\begin{itemize}
	\item using the Local Bottom Feature (LBC) combination of local features, which employs two input images and computes the differences between then in a unique network flow; and
	\item employing the so-called Matching Mid-Level Feature (MMF) to combine both images' characteristics after passing through a specific part of the neural network. 
\end{itemize}

In summary, the first architecture merges the images before inputting them to the neural network, while the other combines the images after going through two layers of transformations. Experiments considering GEIs, Chrono-gait image (CGI), and resolution of the input image present an identification accuracy of $74.4\%$ over OU-ISIR~\cite{Makihara_CVATN2012} dataset.

In a similar work, Sepas et al.~\cite{sepas2021gait} used capsule networks to develop a model capable of learning more discriminative features by transferring multi-scale partial gait representations. The model employs Bi-directional Gated Recurrent Units (BGRU) to learn the co-occurrences and correlations among patterns and further uses a capsule network to extract deeper relationships among such features. Experiments conducted over CASIA B~\cite{casiab} and OU-MVLP~\cite{takemura2018multi} assesses the superiority of the model against state-of-the-art approaches, especially when considering challenging conditions.

Finally, Zhao et al.~\cite{zhao2021associated} introduced an automated learning system called Associated Spatio-Temporal Capsule Network (ASTCapsNet). The model was trained over multi-sensor datasets to show that multi-modality data is more conducive to gait recognition. The model's effectiveness is confirmed on the experiments conducted over several datasets, in which results are compared to state-of-the-art approaches.

\subsection{Recurrent Neural Networks}
\label{ss.rnn}

Concerning Recurrent Neural Networks~\cite{8963659}, Wang and Yan~\cite{wang2020human} used the Long short-term memory~\cite{Hochreiter}, a popular RNN architecture capable of learning long-term dependencies, for cross-view human gait recognition based on frame-by-frame GEIs. Experiments conducted over CASIA B~\cite{casiab} and OU-ISIR~\cite{Makihara_CVATN2012} Large Population datasets demonstrate the model's robustness over several baselines.

A similar work proposed by Potluri et al.~\cite{potluri2019deep} employed LSTMs to detect gait abnormalities through a wearable sensor system. The experiments were conducted over a set of data extracted from ten healthy individuals, such that seven behave normally, while the three remaining simulate specific gait abnormalities, i.e., sensory ataxic, Parkinson's gait, and hemiplegic gait. The experiments focus on employing advanced technologies for gait-based diagnosis and treatment assistant systems. Other models combined specific information, such as data obtained from accelerometer and gyroscope, for gait recognition through deep learning-based approaches, reaching recognition rates above $91\%$~\cite{zou2018deep}.

Recently, Tran et al.~\cite{tran2021multi} proposed an Inertial Measurement Units (IMUs)-based gait recognition approach. The authors employed LSTMs to exploit the temporal information on video sequences, thus extracting hidden patterns inside such sequences. Experiments conducted over whuGAIT~\cite{zou2018deep} and OU-ISIR~\cite{Makihara_CVATN2012} provided state-of-the-art performance for both verification and identification tasks.

\subsection{Autoencoders}
\label{ss.autoencoders}

Babaee at al.~\cite{babaee2019person} also proposed a multiple-stage model for the recognition of gestures. The work tackle a problem commonly observed in several GEI-based identification datasets, i.e., the person's cycle of moviment is not entirely formed. In other words, the data available for identification is not complete. To deal with the problem, the authors proposed an autoencoder-based approach called Incomplete to Complete GEI Network (ITCNet). The model was trained to reconstruct the missing images using examples from the training set with a complete cycle. The network consists of $9$ fully convolutional networks, each of responsible for $1/9$ of the components of the cycle. Further, PCA is employed to compute the main features and an RNN is used for gait recognition. The model presented an index of $86\% $ and $95\% $ for Rank-1 and Rank-5, respectively, over OU-ISR Large Population dataset~\cite{Makihara_CVATN2012} using only $20$ images per instance. 

Autoencoders can also be employed to extract distinct GEI features. In this context, Yu et al.~\cite{yu2017invariant} proposed a study using Stacked Progressive Auto-Encoders (SPAE)~\cite{kan2014stacked} with appropriate changes for the extraction of invariant characteristics of an individual's behavior. In short, the model extracts information from independent components of the scene, such as clothes and other objects the person may eventually be carrying, instead of the movement exclusively. Experiments conducted over CASIA B~\cite{casiab} and SZU RGB~\cite{yu2013large} datasets show that, despite the GEI information for personal identification, intermediate and last layers also provide relevant characteristics regarding such components. Finally, the work combines such information and perform a dimensionality reduction step through PCA. 
 
Zhang et al.~\cite{zhang2019gait} also employed autoencoders for gait recognition. The authors used the technique to tackle a common issue faced by biometrical recognition literature, i.e., dealing with clothes blocking body members' views, such as arms and legs. The main problem concerning such occlusions is the difficulty of reading the movements and recognizing other features such as ligaments and the point of contact with the ground. To minimize the problem, they used autoencoders to disconnect the images' pose and appearance characteristics for further linking those characteristics to another architecture of recurrent networks to analyze the way the person gaits. Besides, the authors also propose the Frontal-View Gait (FVG) dataset.The system proved to be particularly robust when filming the person walking towards the camera, whose angulation is more complicated due to the difficulty of obtaining specific characteristics, such as the distance walked by each step.

\subsection{Deep Belief Networks}
\label{ss.dbn}

Deep Belief Networks (DBNs)~\cite{Hinton:06} are stochastic neural networks constructed using Restricted Boltzmann Machines~\cite{Hinton:02} as building blocks. Such models became very popular due to their ability of performing several tasks, such as feature selection~\cite{Souza2021computer}, classification~\cite{RoderICAISC:20}, and image reconstruction~\cite{passosIJCNN:19,passosCAIP:2017}, among others. Regarding gait recognition, Fernandes et al.~\cite{fernandes2018artificial} employed DBNs to support gait assessment in the diagnosis of Parkinson's and movement disorder diseases. The authors employed wearable sensors to extract features from the subjects and performed a comparative classification analysis of parkinsonian gait, confirming that DBN-based approaches are suitable for the task.

Further, Xiong et al.~\cite{xiong2020continuous} demonstrated how to encode patterns using surface electromyography (sEMG) through Deep Belief Networks. The authors consider the knee and ankle joint angle during walking to estimate a combination of four-time domain features. The work showed a high potential for gait tracking problems.

\subsection{Generative Adversarial Networks}
\label{ss.gans}

A very interesting work by Hu et al.~\cite{hu2018robust} introduces Generative Adversarial Networks~\cite{gan,de2020assisting} to the context of Gait recognition. The authors propose the Discriminant Gait Generative Adversarial Network, i.e., DiGGAN, to extract view-invariant gait characteristics for cross-view gait recognition. Experiments conducted over OU-MVLP~\cite{takemura2018multi} and CASIA B~\cite{casiab} outperformed state-of-the-art results, demonstrating the model's capabilities.

Meanwhile, He et al.~\cite{he2018multi} proposed the Multi-task Generative Adversarial Networks (MGANs), a GAN-based network designed to learn view-specific gait features through the so-called Period Energy Image (PEI), a multi-channel gait template proposed to tackle the problems of view angles' variation faced by cross-view methods and loss of temporal information faced by  GEI-based approaches. The model employs a view-angle manifold to extract more significant features from video sequences, thus providing competitive results over OU-ISIR~\cite{Makihara_CVATN2012}, CASIA B~\cite{casiab}, and USF against various works available in the literature.

Further, Jia et al. ~\cite{jia2019attacking} studied methods to avoid attacks in gait recognition systems through GAN-generated syntectic image sequences. In such a scenario, the authors proposed a GAN-based approach capable of rendering fake videos from source walking sequence with realistic details. The method is compared against two state-of-the-art gait recognition systems, and results are analyzed under attacking scenarios considering both CASIA A~\cite{wang2003silhouette} and CASIA B~\cite{casiab} datasets. The effectiveness of the model is verified in both attacking detection ability and visual fidelity.



\subsection{Discussion}
\label{ss:discussion_dl}

Table~\ref{t.summary} provides a summary of the surveyed studies presented in this section. In general, Convolutional Neural Networks are the most popular choice when considering deep learning solutions, especially concerning image/video-based issues, including gait recognition. Such behavior is expected since CNNs obtained outstanding results in various applications and won most of the benchmark challenges in the last years. Nevertheless, the other architectures also provide a valuable contribution to the field, performing better for several specific tasks.

Capsule Networks, for instance, are capable of extracting partial gait representations in a hierarchical fashion, providing better results for cases where individuals or objects are presented in the scene with multiple orientations or overlapped. Similarly, Recurrent Neural Networks are paramount to deal with sequential data, such as videos, thus presenting themselves as an essential tool for this kind of gait recognition approach.

Although most deep learning-based approaches for gait recognition comprise image/video domain, other data sources, such as accelerometers, gyroscope,  sensor-based, and handcrafted features in general, also provided impressive results in a considerable number of works. Most of these works address unsupervised deep learning approaches in the process, such as Autoencoders and DBNs, which usually are more expressive over these data types, while CNNs are paramount when dealing with raw image/video files. These unsupervised methods can extract information regarding the data distribution and manipulate it, usually in a lower-dimensional space, providing more representative features for gait recognition. 

Finally, Generative Adversarial Networks describe a particular case when gait systems can learn a broader range of features, such as orientations, clothes, number of individuals in the scene, and so on, since they can generate synthetic data for training models. Besides, they are also useful for evaluating frauds in gait-based systems for security purposes by producing fake images for testing purposes.

Concerning the most used methods to represent gait image data, Gait Energy Image reflects the sequence of a simple energy image cycle using the weighted average method. Further, the sequences in a travel cycle are processed to align the binary silhouette \cite{grgei}. Therefore, GEI maintains the static and dynamic characteristics of human walking and significantly reduces image processing's computational cost. From an in-depth analysis of the method, one can observe a few points that characterize the model~\cite{bhanu}, described as follows:

\begin{itemize}
	\item GEIs are a little sensitive to silhouette noise in individual pictures.
	\item It focuses on specific representations of the human walk, which does not soften the context of vector images.
	\item It represents human motion in a single image while preserving temporal information.
\end{itemize}

Similarly, cross-view-based gait recognition is a popular approach used to deal with different visual angles. The input type requires multiple fully controlled cameras and cooperative environments, thus being restrained to real scenarios. Moreover, it visually normalizes gait characteristics before performing any combination, which allows the model to learn the relationships among visual movements in the scene~\cite{nixon, hongdong, makihara}. An example of the approach can be observed in the UK's Newcastle University cross-view gait recognition system, the so-called DiGGAN, which obtained state-of-the-art results over the largest multiview gait dataset in the world~\cite{takemura2018multi} (comprising more than $10,000$ people) using cross-view approaches~\cite{hu2018robust}.

Despite the success obtained by such methods, they still lack in some aspects, in which some less popular techniques aim to overcome. The Period Energy Image, for instance, comprises a multi-channel gait template proposed to tackle view angles' variation faced by cross-view and GEI loss of temporal information~\cite{he2018multi}. Another approach observed in some works comprises Optical-flow, which provides relevant information regarding the objects observed in the scene, their spatial arrangements, and the changes in sich arrangements~\cite{luo2016gait}. Finally, a wide range of hand-crafted features or sensor-based input data types is also employed for deep learning-based gait recognition approaches. Among such methods, one can find the Inertial Measurement Units, accelerometers, gyroscopes, sensor output, and so on.

Concerning gait recognition's actual state-of-the-art scenario, one can consider two main approaches, i.e., literature- and representation-based. Regarding the literature-based, 2D-CNNs are the most widely used type of deep neural network (DNN) for gait recognition using deep learning, with approximately 50\% of solutions just based on 2D-CNN architectures for classification. Solutions using 3D-CNN and GAN are the next popular categories, each corresponding to 10\% of the published content. In addition, deep autoencoder (DAE), RNN, CapsNet, DBN, and graph convolutional networks are less considered among DNNs, corresponding to $5\%$, $3\%$, $2\%$, $1\%$, and $1\%$, respectively. On the other hand, the hybrid methods that constitute $26\%$ of the solutions, in which the CNN-RNN combinations are the most widely adopted approach with about $15\%$ of presence, while the combination of DAE with GAN and RNN corresponds to $10\%$ of the methods, followed by the RNN-CapsNet methods that constitute 2\% of the solutions.

Regarding state-of-the-art solutions based on representation, the silhouettes are the most widely adopted for gait recognition, corresponding to more than $85\%$ of the solutions. Although it is a promising approach, skeletons have been considered less frequently in relation to silhouettes, corresponding to only $10\%$ of the available solutions. There were also some methods, that is, approximately $5\%$ of the available literature, that explore the representations of the skeleton and the silhouette, notably using unraveled representation learning or score of fusion strategies.

Considering deep learning-based gait recognition main challenges, one can refer to the complexity of gait data, which arises from the interaction between many factors, such as occlusion/obstruction, camera points of view, the appearance of individuals, order of sequence, movement of body parts, or light sources present in the data, among others~\cite{zanglim,lilim}. Such factors can interfere in a complicated way and can hinder the task of gait recognition. Currently, there is an increasing number of methods in other areas related to pattern recognition, such as face recognition, emotion, and pose estimation. These professionals focus on learning confusing contexts, extracting representations that separate the various explanatory factors in the data's high-dimensional space. However, most gait recognition methods available using deep learning have not yet explored such approaches. Therefore, they are not explicitly able to separate the underlying structure of gait data in the form of significant disjoint variables. Despite recent progress in using confusing context approaches in some gait recognition methods, there is still room for improvement.

\begin{center}
\begin{longtable}{| p{.05\textwidth} | p{.05\textwidth} | p{.18\textwidth} | p{.18\textwidth} |p{.15\textwidth} |p{.08\textwidth} |p{.14\textwidth} |}
\caption{Deep learning-based gait recognition approaches.}  \\

\hline \multicolumn{1}{|p{.05\textwidth} |}{\textbf{Ref.}} & \multicolumn{1}{p{.05\textwidth}|}{\textbf{Year}}& \multicolumn{1}{p{.18\textwidth}|}{\textbf{Model}}& \multicolumn{1}{p{.18\textwidth}|}{\textbf{Input Type}} & \multicolumn{1}{p{.15\textwidth}|}{\textbf{Dataset}}& \multicolumn{1}{p{.08\textwidth}|}{\textbf{Result}}& \multicolumn{1}{p{.14\textwidth}|}{\textbf{Measure}}\\ \hline 
\endfirsthead

\multicolumn{7}{c}%
{{\bfseries \tablename\ \thetable{} -- continued from previous page}} \\
\hline \multicolumn{1}{|p{.05\textwidth} |}{\textbf{Ref.}}  & \multicolumn{1}{p{.05\textwidth}|}{\textbf{Year}}& \multicolumn{1}{p{.18\textwidth}|}{\textbf{Model}}& \multicolumn{1}{p{.18\textwidth}|}{\textbf{Input Type}} & \multicolumn{1}{p{.15\textwidth}|}{\textbf{Dataset}} & \multicolumn{1}{p{.08\textwidth}|}{\textbf{Result}}& \multicolumn{1}{p{.14\textwidth}|}{\textbf{Measure}} \\ \hline 
\endhead

\hline \multicolumn{7}{|r|}{{Continued on next page}} \\ \hline
\endfoot

\hline \hline
\endlastfoot

\cite{shiraga2016geinet} & 2016& CNN & GEI & OU-ISIR & $94.6\%$&Identification rate\\ \hline

\cite{wu2017comprehensive} &2017& CNN & Cross-view  & CASIA B & $90.8\%$ &Accuracy\\ \hline
\cite{Li:2017Deepgait} & 2017 & CNN + Joint Bayesian & Sensors &OU-ISR & $97.6\%$&Identification rate\\ \hline
\cite{sokolova2017gait} & 2017 & CNN & Optical flow &TUM-GAID and CASIA B &$97.52\%$&Accuracy\\ \hline
\cite{takemura2017input}& 2017 & CNN + Siamese networks  & Cross-view &OU-ISIR  &$98.8\%$&Accuracy\\ \hline
\cite{sokolova2017posed} & 2017 & CNN + Nearest Neighbor & Optical Flow &TUM-GAID, CASIA B, and OU-ISIR &$99.8\%$&Identification rate\\ \hline
\cite{yu2017invariant} & 2017 & Autoencoders + PCA & GEI & CASIA B and SZU RGB &$97.58\%$&Identification rate\\ \hline

\cite{zou2018deep}& 2018 & CNN + LSTM  &Accelerometer and Gyroscope&   whuGAIT and OU-ISIR& $99.75\%$&Accuracy\\ \hline
\cite{he2018multi}& 2018 & GAN & PEI  & OU-ISIR, CASIA B and USF & $94.7\%$&Accuracy\\ \hline
\cite{fernandes2018artificial}& 2018 & DBN & Sensors & Data collected by the authors & $93\%$&Accuracy\\ \hline

\cite{xu2019gait} & 2019 &Capsule & LBC and MMF & OU-ISIR &$74.4\%$ &Accuracy \\ \hline
\cite{wang2019nonstandard} & 2019& CNN &GEI + Data Augmentation & CASIA B&$98\%$ &Accuracy \\ \hline
\cite{zhang2019gait} & 2019 & Autoencoders + LSTN & Croos- and Frontal-view&  CASIA B, USF, and FVG &$99.1\%$&Accuracy\\ \hline
\cite{babaee2019person} & 2019 & Autoencoders + PCA & GEI & OU-ISIR and CASIA B& $96.15\%$&Accuracy\\ \hline
\cite{potluri2019deep} & 2019 & LSTM & Sensors & Data collected by the authors &$0.02$&Prediction error\\ \hline
\cite{jia2019attacking}& 2019 & GAN & GEI & CASIA A and CASIA B  & $82\%$ &Recognition result\\ \hline

\cite{wang2020human} &2020& LSTM & GEI & CASIA B and OU-ISIR & $99.1\%$&Recognition rate \\ \hline
\cite{xu2020cross} & 2020 & CNN & Cross-view  & OU-MVLP, OU-LP, and CASIA B &$98.93\%$&Identification rate\\ \hline
\cite{hu2018robust} & 2020 & GAN & Cross-view  & OU-MVLP and CASIA B &$93.2\%$&Identification rate\\ \hline
\cite{sepas2021gait}& 2020 & Capsule &Multi-scale representations & CASIA-B and OU-MVLP & $84.5\%$&Identification rate\\ \hline
\cite{xiong2020continuous}& 2020 & DBN &Sensors & Data collected by the authors & $2.61$ &RMSE\\ \hline

\cite{tran2021multi}& 2021 & LSTM & IMU & whuGAIT and OU-ISIR& $94.15\%$ &Accuracy\\ \hline
\cite{zhao2021associated}& 2021 & Capsule & Sensors & Several  & $99.69\%$ &Accuracy

\label{t.summary}
\end{longtable}
\end{center}

\section{Datasets}
\label{sec:datasets}

Machine learning models' training and evaluation steps, regardless of the adopted paradigm, i.e., supervised, unsupervised, or any other, depend on a dataset comprising the task's subject. Besides, the employment of such datasets makes it possible to determine how effective a method is to solve a specific problem and compare it with other solutions.

Regarding gait recognition, the availability of specific datasets for the task is minimal, considering both public or private solutions. Such a lack of training data hampers the development of new artificial intelligence models capable of recognizing people by how they walk or move. Two problems stand out for obtaining and creating a dataset:

\begin{itemize}
    \item Gait biometry demands a reasonable amount of movement recordings for a subject, implying recording and generating multiple videos for each individual. Besides, such videos usually possess an inherent high dimensionality, entailing in the dataset's growth and thus requiring a high storage capacity.

    \item The extraction and public distribution of biometric data require permission from each participant. The eventual creation of a dataset without formal consent from each individual may bring about lawsuits.
\end{itemize}

The next section gathers the most-used datasets available for gait recognition related tasks.

\subsection{CMU MoBo Dataset}
\label{ss.cmuMobo}

The CMU MoBo~\cite{Gross-2001-8255} dataset possesses a relatively small amount of data, comprising several videos for gait recognition extracted from $20$ people. The dataset also provides a set of silhouette masks and bounding boxes, thus alleviating the segmentation process.

One of the main advantages of the dataset is that it is available for download without reservations or need to sign forms of agreement, requiring just connecting to the Calgary University File Transfer Protocol (FTP) server\footnote{ftp://ftp.cc.gatech.edu/pub/gvu/cpl/}.

\subsection{TUM GAID Dataset}
\label{ss.tumGaid}

The TUM Gait from Audio, Image, and Depth (GAID)~\cite{hofmann2014tum} database, as the name suggests, is compounded by RGB images, audio, and depth, recorded from $305$ people in three different variations at first. Further, $32$ people were re-recorded to have some variation, comprising $3,370$ records in total. According to the authors, it is the only dataset that allows recognition by combining video, depth, and audio. 

\begin{figure}[!htb]
    \centering
    \includegraphics[width=.5\textwidth]{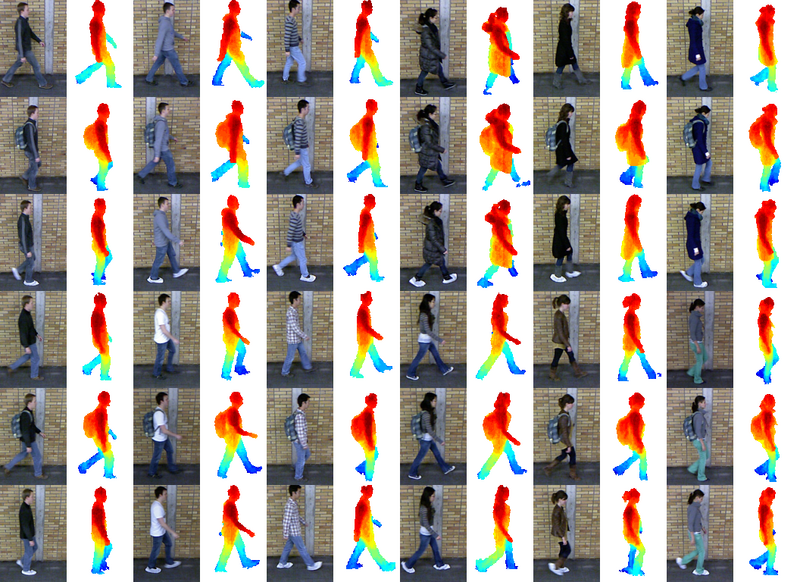}
    \caption{Examples of three male (top rows) and three female (bottom rows) participants in six variations: normal (columns $1$ and $2$), backpack (columns $3$ and $4$), coating shoes (columns $5$ and $6$), time (columns $7$ and $8$), time + backpack (columns $9$ and $10$), and time + coating shoes (columns $10$ and $11$). Extracted from~\url{https://www.ei.tum.de/mmk/verschiedenes/tum-gaid-database/}}.
    \label{fig:tumgaid}
\end{figure}

Variations include different shows, carrying conditions (with and without bag), and time. The first record was performed in 2012's winter (January) and comprised $176$ recordings, using a jacket and winter boots. The second performance was made in April 2012 by $161$ people using considerably different clothes, since it was warmer. From this amount, $32$ people were recorded both times, which grant cloth and time variation. A strong advantage of this dataset is a well-defined evaluation protocol. Figure~\ref{fig:tumgaid} depicts some examples of images and depth. The dataset is available after sign a document request\footnote{https://www.ei.tum.de/mmk/verschiedenes/tum-gaid-database/}.

\subsection{HID-UMD Dataset}
\label{ss.UMD}

The Human Identification at a Distance (HID)-UMD dataset comprises several videos from people walking captured in four different angles and their respective binary masks for foreground segmentation. Its primary purpose is to help researchers develop new gait and facial biometrics recognition methods. Moreover, the dataset is an aggregate composed of two datasets, described as follows:

\begin{itemize}
    \item \textbf{Dataset 1}~\cite{kale2002framework}: composed of walking sequences of $25$ individuas in $4$ distinct poses:
    \begin{enumerate}
        \item Frontal view/walking-toward;
        \item Frontal view/walking-away;
        \item Frontal-parallel view/toward left; and
        \item Frontal-parallel view/toward right.
    \end{enumerate}
    \item \textbf{Dataset 2}~\cite{cuntoor2003combining}: comprises videos from $55$ individuals walking through a T-shape pathway. The sequences were obtained by two cameras that line orthogonal to each other.

\end{itemize}

More details can be found at \url{http://www.umiacs.umd.edu/labs/pirl/hid/umd-eval.html}. Further, the dataset is available to download through FTP, requiring credentials that can be requested at \url{http://www.umiacs.umd.edu/labs/pirl/hid/data.html}.

\subsection{CASIA}
\label{ss.casia}

\begin{figure}[!htb]
\centering

	 \begin{tabular}{cc}
	  \includegraphics[width=0.3\textwidth]{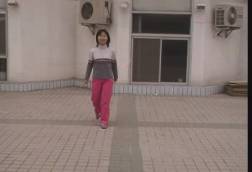} &
	  \includegraphics[width=0.3\textwidth]{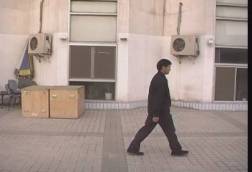}  \\
		(a) & (b)
	 \end{tabular}
	\centering
	 \begin{tabular}{c}
	  \includegraphics[width=0.3\textwidth]{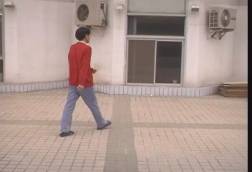}\\
		(c) 
	 \end{tabular}
    \caption{Example frames from CASIA A dataset from each angle. (a) corresponds to the parallel view, (b) corresponds to $90$ degrees, and (c) is an example from $45$ degrees. Images collectd from~\url{http://www.cbsr.ia.ac.cn/users/szheng/?page\_id=71.}}
    \label{fig:casiaA}
\end{figure}

\begin{figure}[!htb]
    \centering
    \includegraphics[width=.7\textwidth]{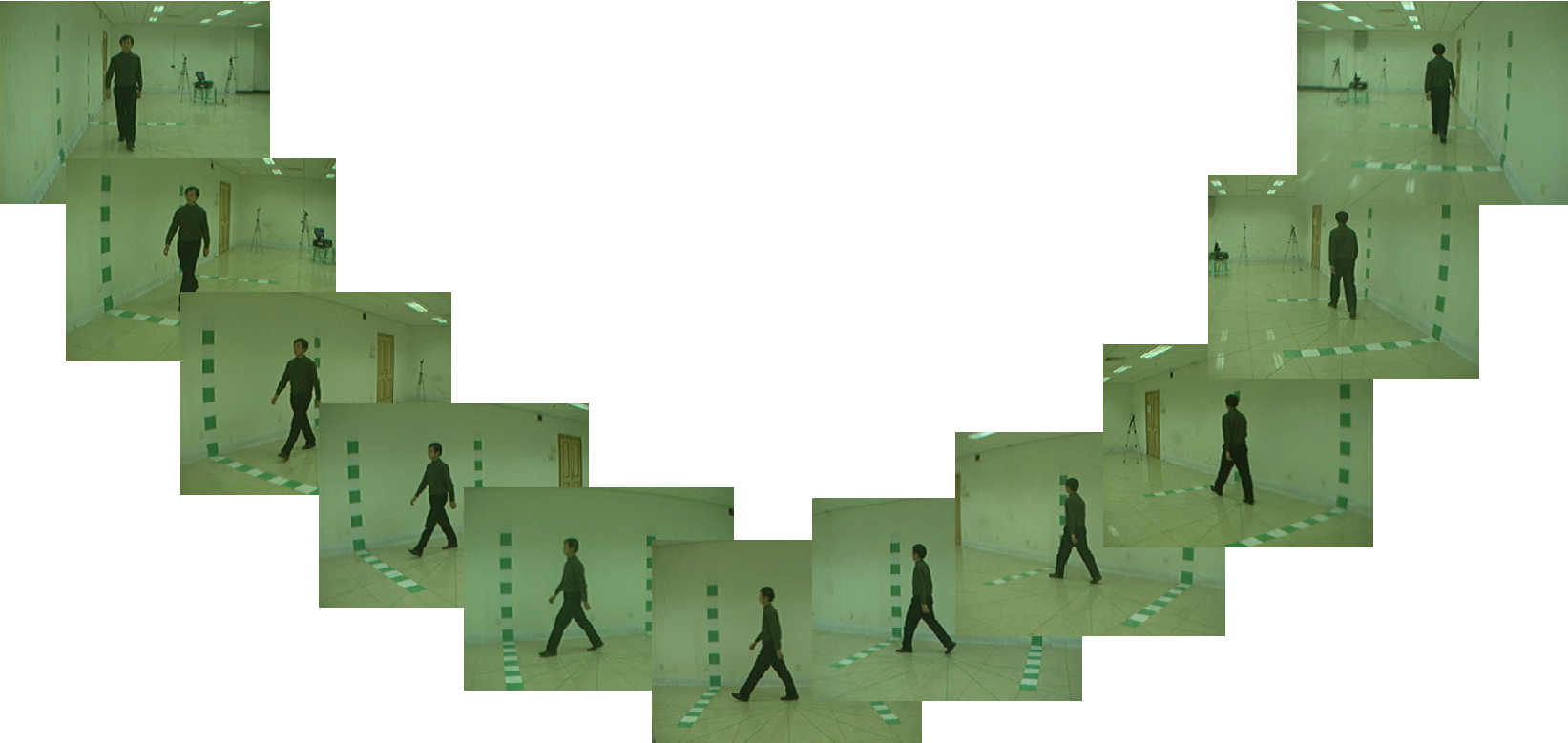}\hfill
    \caption{Example frames from CASIA B dataset. Each image corresponds to one of the $11$ comprised angles. Extracted from~\url{http://www.cbsr.ia.ac.cn/users/szheng/?page\_id=71.}}
    \label{fig:casiaB-var}
\end{figure}

\begin{figure}[htp]
    \centering
	 \begin{tabular}{cc}
	  \includegraphics[width=0.3\textwidth]{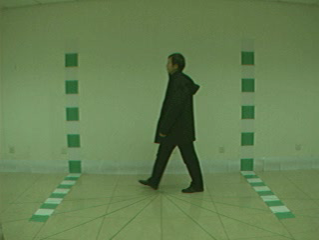} &
	  \includegraphics[width=0.3\textwidth]{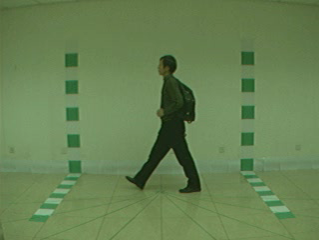}  \\
		(a) & (b)
	 \end{tabular}
	\centering
	 \begin{tabular}{c}
	  \includegraphics[width=0.3\textwidth]{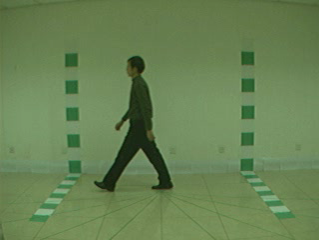}\\
		(c) 
	 \end{tabular}
    \caption{Example frames from CASIA B dataset. One can notice the change in clothes and personal objects, like backpacks. Images collectd from~\url{http://www.cbsr.ia.ac.cn/users/szheng/?page\_id=71.} }
\label{fig:casiaB-angles}

\end{figure}

\begin{figure}[htp]
  \centerline{
    \begin{tabular}{cc}
        \includegraphics[width=0.35\textwidth]{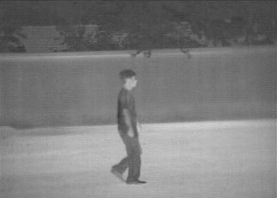}&
        \includegraphics[width=0.35\textwidth]{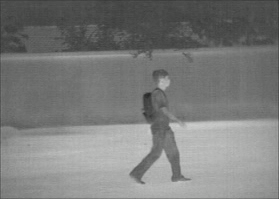}\\
		(a) & (b)
    \end{tabular}}	
    \caption{Example frames from CASIA C dataset. The images were obtained through an infrared camera at night and with variations in the manner of walking. Adapted from~\url{http://www.cbsr.ia.ac.cn/users/szheng/?page\_id=71.}}
    \label{fig:casiaC}
\end{figure}

\begin{figure}[htp]
    \centering
    \includegraphics[width=.45\textwidth]{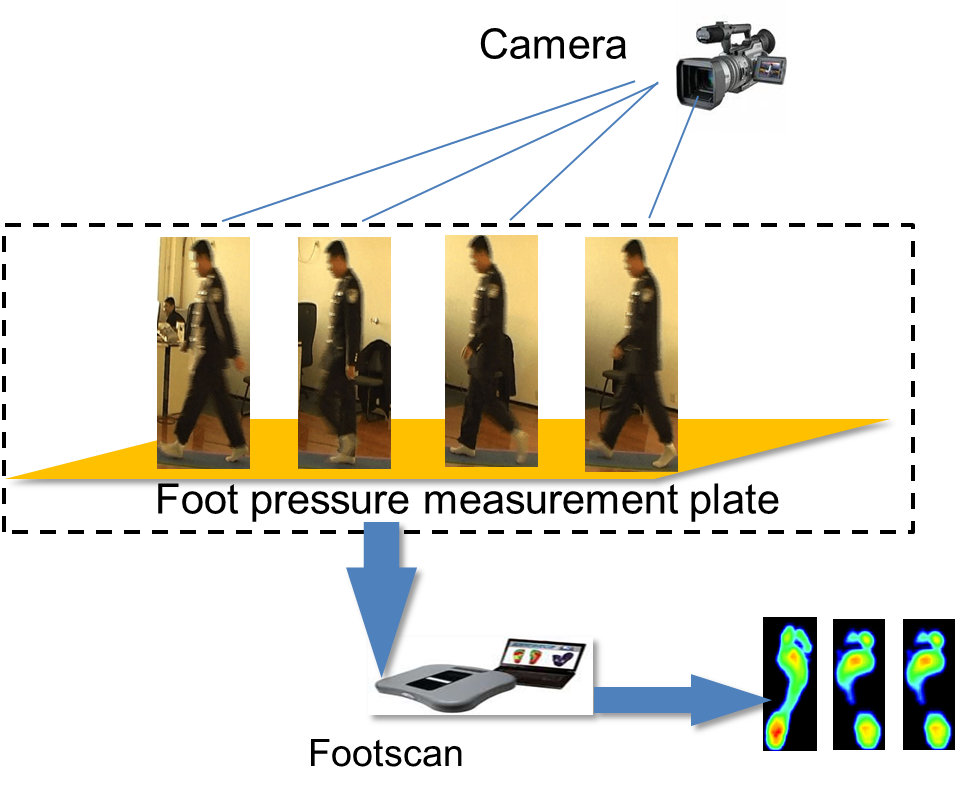}\hfill
    \caption{Example frames from CASIA D dataset. The images illustrate the image's acquisition process synchronously with the footsteps data. Image collectd from~\url{http://www.cbsr.ia.ac.cn/users/szheng/?page\_id=71.}}
    \label{fig:casiaD}
\end{figure}

The Institute of Automation, Chinese Academy of Sciences (CASIA) provides the CASIA Gait Database~\cite{zheng2011robust}, a collection of four datasets designed for gait recognition purposes, and described as follows:

\begin{itemize}
    \item \textbf{CASIA A:} Created in December 2001 and formerly known as the NLPR Gait Database~\cite{wang2003silhouette}, the CASIA A dataset includes 20 individuals, each of them comprising $12$ videos, i.e., $4$ videos for each of the $3$ directions, namely parallel, $45$ and $90$ degrees to the image plane. Besides, each image sequence possesses its duration, varying with the individual walking velocity. Figure~\ref{fig:casiaA} shows some image examples from different angles. The dataset's total size is approximately 2.2GB;

    \item \textbf{CASIA B:} Created in 2005, the CASIA B~\cite{casiab} comprises filming $124$ individuals from $11$ different angles. Each sequence was repeated three times, with variations such as clothing and walking speed. Moreover, the dataset also comprises a set of silhouettes provided for all sequences, provided for foreground segmentation. Figures~\ref{fig:casiaB-var} and~\ref{fig:casiaB-angles} show variations in angle and clothing.
    
    \item \textbf{CASIA C:} Acquired in 2005, the CASIA C~\cite{tan2006efficient} dataset contains $153$ subjects filmed by infrared cameras (thermal spectrum) in four different variations: normal walking, slow walking, fast walking, and normal walking carrying a backpack. Figure~\ref{fig:casiaC} show some examples. All images were captured at night;

    \item \textit{\textbf{CASIA D:}} The CASIA gait–footprint dataset~\cite{zheng2012cascade} contains both images and cumulative foot pressure information. The dataset comprises $3,496$ gait pose images and $2,658$ cumulative foot pressure images from $88$ individuals with a broad distribution of age, $20$ female and $68$ male, in an indoor environment. Figure~\ref{fig:casiaD} illustrates the data acquisition schematics. 
   
\end{itemize}

Silhouette for datasets A, B, and C are freely available for download\footnote{http://www.cbsr.ia.ac.cn/english/Gait\%20Databases.asp}. Regarding data acquisition, candidates should fill a form and wait for the dataset owner's approval.

\subsection{OU-ISIR Biometric Database}
\label{ss.ouisir}

The Institute of Scientific and Industrial Research (ISIR) of Osaka University (OU) has been creating, since 2007, the largest dataset for gait recognition in the world. The project is an aggregation of eight different groups:

\begin{itemize}
    \item \textit{\textbf{Treadmill Dataset:}} The group is composed of sequences from people walking on electronic treadmills surrounded by $25$ cameras filming at $60$ frames per second at a resolution of $640\times480$. It has $4$ subdivisions:
    \begin{enumerate}
         \item \textbf{Treadmill dataset A -- Speed variation}~\cite{tsuji2010silhouette}: this subset comprises $34$ subjects in a lateral vision with speed varying between $2$ and $10$km/h in $1$hm/h intervals;
         
         \item \textbf{Treadmill dataset B -- Clothes variation}~\cite{hossain2010clothing}: It is composed of $68$ people in lateral vision with $32$ clothing variations;
         
         \item \textbf{Treadmill dataset C -- View variation}~\cite{makihara2010gait}: A large-scale database comprising $168$ people with ages ranging from $4$ to $75$ years old. Moreover, the subset is composed of 25 views observed through a multi-view synchronous gait system;
         
         \item \textbf{Treadmill dataset D -- Gait fluctuation}~\cite{mori2010gait}: this set is composed of gait silhouette sequences with $185$ subjects, viewed from a lateral angle and with variations in velocity. The data were subdivided into two groups of $100$ subjects (with an overlap of 15 people) by high and low speed variations.
    \end{enumerate}

    \item \textbf{Large Population Dataset}~\cite{Iwama_IFS2012}: Collected since 2009 through outreach activity events, the Large Population Dataset is composed of $4,016$ subjects, each of them filmed twice from $4$ camera angles at $30$ FPS and a resolution of $640\times480$ pixels. 

    \item \textbf{Speed Transition Dataset}~\cite{mansur2014gait}: The Speed Transition Dataset comprises two subsets, described as follows:
    \begin{enumerate}
        \item \textbf{Dataset A:} It contains $179$ scenes from people walking at a constant velocity of $4$km/h on a treadmill or the ground. In this set, the background has been removed using a wallpaper;
     
        \item \textbf{Dataset B:} It comprises sequences from $25$ people walking on a treadmill with a velocity varying between $1$ and $5$ km/h. Each person is filmed twice. Acceleration and deacceleration are performed in three seconds, and middle sequences with one second were extracted from both.
    \end{enumerate}

    \item \textbf{Multi-view Large Population Dataset}~\cite{takemura2018multi}: This dataset is composed of $10,307$ samples such that $5,114$ regards men and the remaining $5,193$ stand for women, whose age ranges from $2$ to $87$ years, developed for motion recognition methods with cross-vision. The images were filmed in $14$ different angles, at a frame rate of $25$ frames per second and a resolution of $1280\times980$. The devices employed for capture were placed at a lateral distance and height of $8$ and $5$ meters, respectively.
 
    \item \textbf{Large Population Dataset with Bag}~\cite{uddin2018isir}: The dataset focuses on gait recognition concerning people carrying objects, aiming not only to rely on biometrics information but also on identifying the position of the transported part (if any) regarding the body. The Large Population Dataset with Bag comprises $62,528$ people aged from $2$ to $95$ years obtained through a camera at a distance of approximately $8$ meters and $5$ meters height. The sequences were filmed at $25$ frames per second with a resolution of $1280\times980$ pixels. Each person was filmed three times, such that the first, i.e., A1, is carrying or not an object, while the second and third do not bring anything. Finally, a total of four regions are marked in case of something being carried, i.e., lower side, upper side, front, and back. All videos are also presented with a respective binary mask for background removal.

    \item \textbf{Large Population Dataset with Age}~\cite{Xu_CVA2017}: The Large Population Dataset with Age was created to investigate gait recognition concerning people's age and gender. The dataset comprises $62,846$ individuals walking on a particular path with cameras capturing $640\times480$ pixels' resolution at a rate of $30$ frames per second. The sequences' people are between $2$ and $90$ years old, and all videos have binary masks obtained after the background removal.  

    \item \textbf{Inertial Sensor Dataset}~\cite{ngo2014largest}: Designated for research and evaluation of methods of individual identification by movements through motion sensors and accelerometers, the Inertial Sensor Dataset is the largest inertial sensor-based gait database, composed of images collected from $744$ subjects ($389$ males and $355$ females) whose ages range from $2$ to $78$ years.

    \item \textbf{Similar Actions Inertial Dataset}~\cite{ngo2015similar}: The Similar Actions Inertial Dataset comprises $460$ participants aged between $8$ and $78$ with gender virtually equal distributed, whose walking properties were obtained together with the data presented in~\cite{ngo2014largest}. Additionally, this dataset also presents six distinct characteristics of the floor: invalid, flat, stair climbing, stair climbing, ramp climbing, and ramp descent.

\end{itemize}

\subsection{University of South Florida Dataset }
\label{ss.usf}

The University of South Florida (USF) dataset~\cite{sarkar2005humanid} comprises $1,870$ sequences from $122$ subjects using two different shoe types. The dataset also considers individuals carrying or not a briefcase, diverse surface conditions such as grass and concrete, and distinct camera views, i.e., left or right viewpoints. The videos are captured at two different time instants filmed in outdoor environments

\subsection{Southampton Dataset }
\label{ss.SOTON}
		
The Southampton Human ID at a Distance (SOTON) database is a contribution of the University of Southampton composed of three major segments\footnote{More information available at~\url{http://www.eng.usf.edu/cvprg/Gait\_Data.html}.}: 

\begin{itemize}
	\item \textbf{SOTON Small database}~\cite{nixon2001experimental}: Comprises $12$ subjects walking around an inside track at varying speeds, wearing different shoes and clothes, and carrying or not bags;
	\item \textbf{SOTON Large database}~\cite{shutler2004large}: Containing $114$ subjects walking outside, inside on the laboratory track, and inside in a treadmill. Images were filmed from six different angles and provided in a collection with more than $5,000$ sequences.

	\item \textbf{SOTON Temporal}~\cite{matovski2010effect}: The data was captured using a Multi-Biometric Tunnel, which contains $12$ synchronized cameras to capture people's gait over time. The dataset is composed of dynamic environments, comprising distinct background, lighting, walking surface, and position of cameras. The dataset includes $25$ subjects ($17$ male and $8$ female) with ages ranging from $20$ to $55$ years old. Notice they are all filmed barefoot.
\end{itemize}

\subsection{AVA Multi-View Dataset for Gait Recognition (AVAMVG)}
\label{ss.AVAMVG}

\begin{figure}[!htb]
	
	\centering
	
	\includegraphics[width=.5\textwidth]{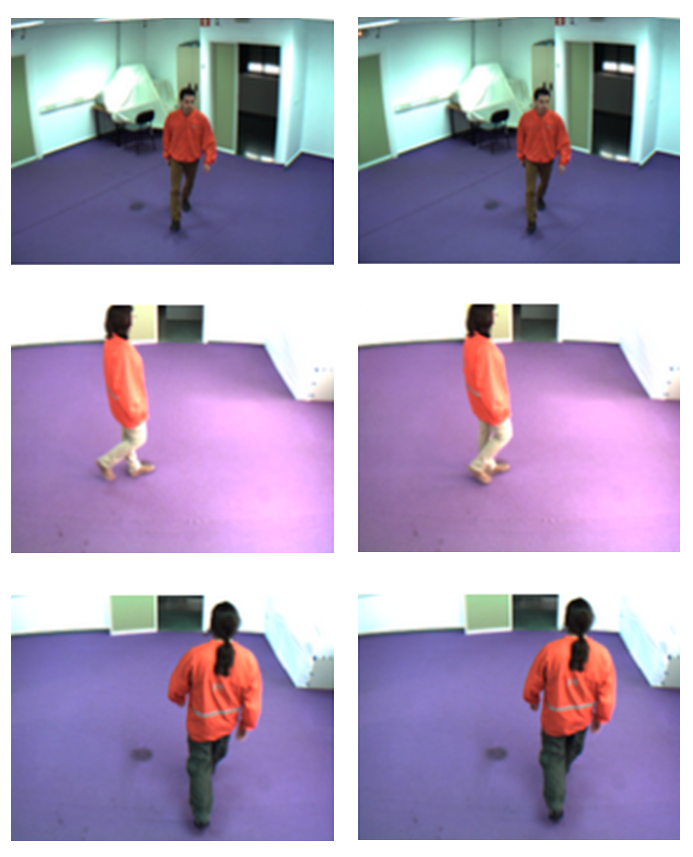}\hfill
	\caption{Example of the multiview dataset. The image presents people walking in different directions, from multiple points of view. Adapted from~\url{https://www.uco.es/investiga/grupos/ava/node/41.}}
	\label{fig:avam}
	
\end{figure}

The AVA Multi-View Dataset for Gait Recognition (AVAMVG)\footnote{More information at:~\url{http://www.uco.es/grupos/ava/node/41}}~\cite{AVAMVG} is a database specifically designed for 3D-based gait recognition algorithms, which comprises gait images from $20$ actors depicting different trajectories. The sequences were obtained using cameras specifically calibrated for the task, followed by a post-processing step using 3D image reconstruction algorithms. Besides, each sequence is also provided with a respective binary silhouette for segmentation. Finally, the database contains $200$ six-channel multi-view videos that can also be employed as $1,200$ single view videos, i.e., $6 \times 200$. Figure~\ref{fig:avam} depicts some samples.

\subsection{Kyushu University 4D Gait Database}
\label{ss.KY4D}

\begin{figure}[!htb]
	
	\centering
	
	\includegraphics[width=.55\textwidth]{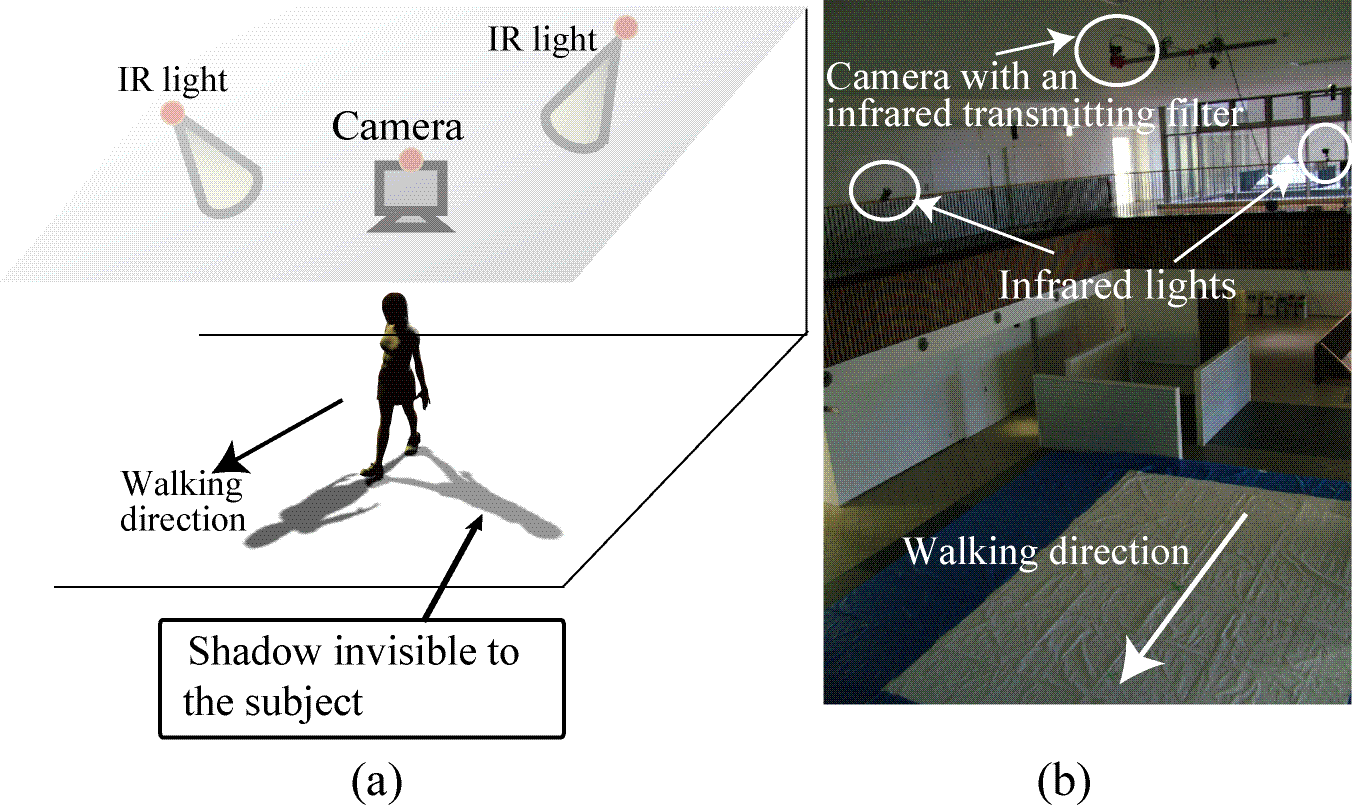}\hfill
	\caption{(a) Experimental setting, (b) actual scene. Adapted from~\url{http://robotics.ait.kyushu-u.ac.jp/~yumi/db/gait\_b.html.}}
	\label{fig:ky4dinfrared}
	
\end{figure}

\begin{figure}[!htb]
	
	\centering	
	\includegraphics[width=.55\textwidth]{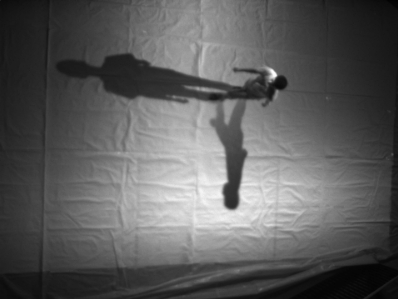}\hfill
	\caption{An example movie from IR shadow images. Image collectd from~\url{http://robotics.ait.kyushu-u.ac.jp/~yumi/db/gait\_b.html.}}
	\label{fig:ky4dinfrared1}
	
\end{figure}

The Kyushu University 4D Gait Database (KY4D)\footnote{Available at~\url{http://robotics.ait.kyushu-u.ac.jp/~yumi/db.html}}~\cite{yumi2014gait} is composed of sequential 3D models and image sequences of $42$ subjects walking along four straight and two curved trajectories. The videos were recorded by $16$ cameras, at a resolution of $1032\times776$ pixels, and divided into three subsets, described as follows:

\begin{itemize}
	\item \textbf{Dataset A (Straight)}: It is composed of sequential 3D models and image sequences of people walking along straight trajectories. Figure~\ref{fig:ky4d_a}(a) illustrates the trajectory, described as a red arrow. Figure~\ref{fig:ky4d_a}(b) depicts multiple 3D reconstructed models.
	
\begin{figure}[!htb]
	
	\centering
	\begin{tabular}{cc}
	    \includegraphics[width=.48\textwidth]{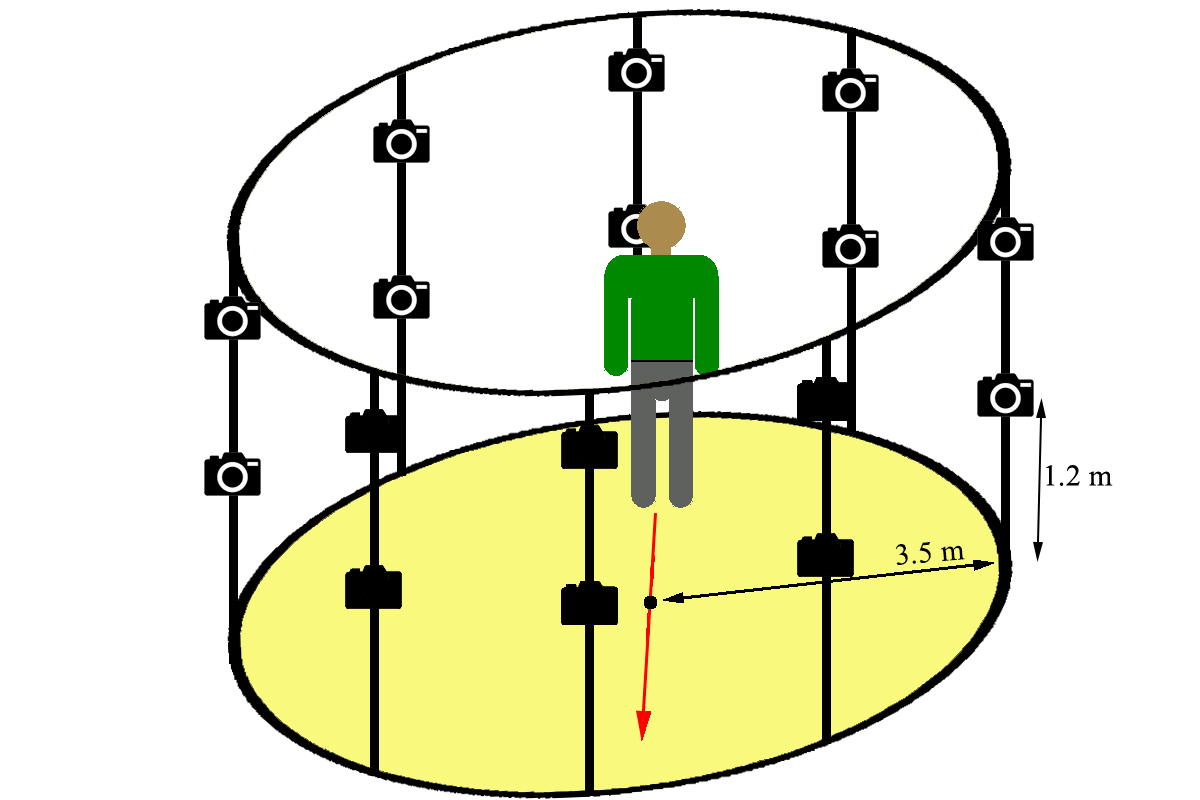}&
	    \includegraphics[width=.435\textwidth]{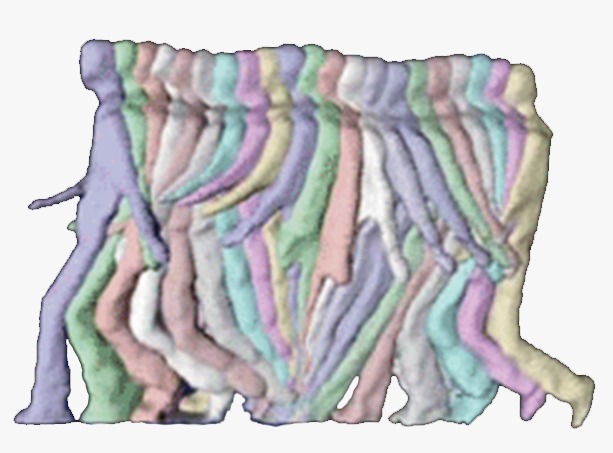}\\
	    (a) & (b)
    \end{tabular}
    \caption{The studio from KY4D Gait Database A: (a) describes the trajectory indicated by the arrow and (b) depicts sequential 3D models of a person walking straight. Images adapted from~\url{http://robotics.ait.kyushu-u.ac.jp/~yumi/db/gait\_b.html.}}

	\label{fig:ky4d_a}
	
\end{figure}

	\item \textbf{Database B (Curve):} It comprises image sequences from people walking along curved trajectories, as depicted in Figure~\ref{fig:ky4d_b}(a). The radius \textit{r} varies either $1.5$m or $3.0$m. Figure~\ref{fig:ky4d_b}(b) depicts multiple 3D reconstructed models considering the curve path.
	
\begin{figure}[!htb]
	
	\centering
	\begin{tabular}{cc}
	    \includegraphics[width=.48\textwidth]{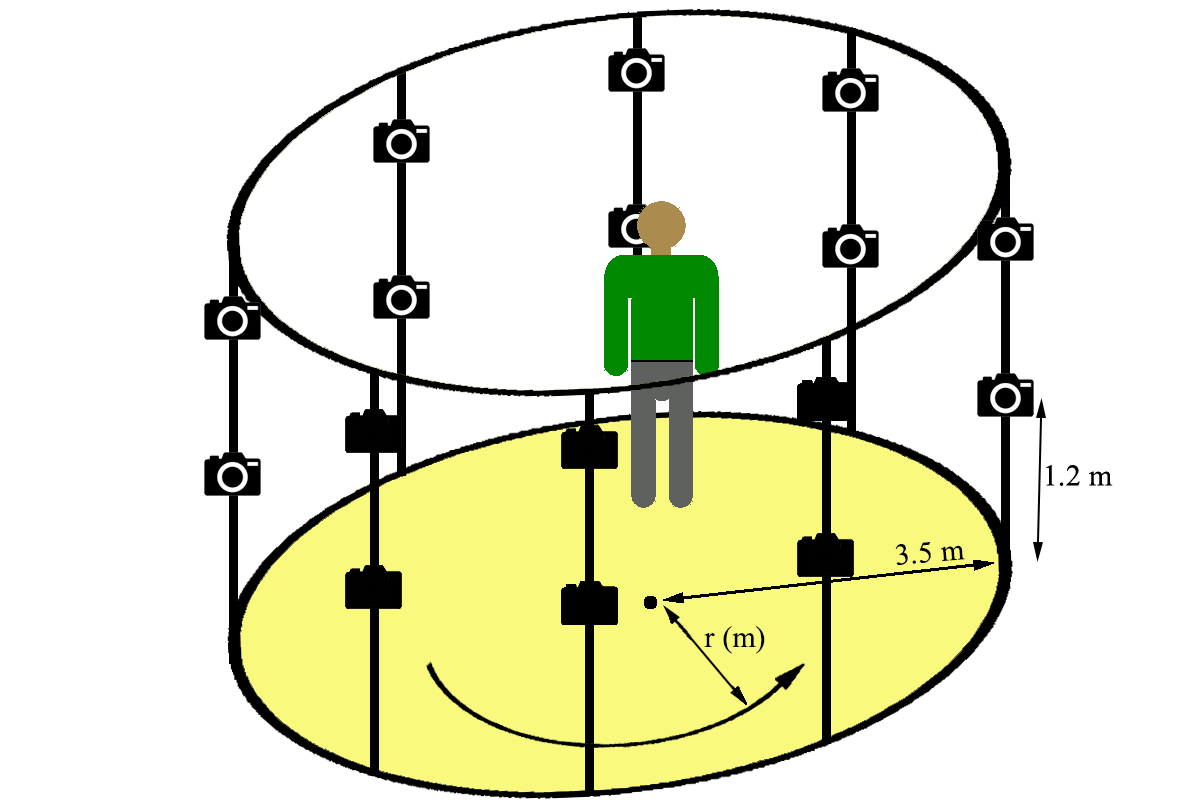}&
	    \includegraphics[width=.435\textwidth,height=4.45cm]{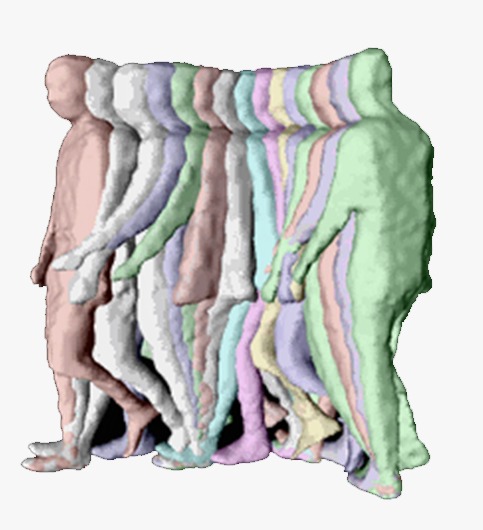}\\
	    (a) & (b)
	\end{tabular}
	\caption{The studio from KY4D Gait Database B: (a) describes the radius variation and (b) illustrates sequential 3D models of a person walking along a curved trajectory. Images adapted from~\url{http://robotics.ait.kyushu-u.ac.jp/~yumi/db/gait\_b.html.}} 
	\label{fig:ky4d_b}
	
\end{figure}
	 
	\item \textbf{KY Infrared (IR) Shadow Gait Database:} It is composed of time-series shadow images of $54$ subjects. As indicated by the walking direction arrow in Figure~\ref{fig:ky4dinfrared}(a), all people walk straight. Figure~\ref{fig:ky4dinfrared}(b) depicts two infrared lights and a camera employed to collect the shadow database~\cite{yumi2014IRgait}. The infrared lights were placed obliquely, and the camera was placed on the ceiling perpendicular to the ground. A sample result from such captures is observed in Figure~\ref{fig:ky4dinfrared1}. 
	
\end{itemize}

\subsection{WhuGAIT Datasets}
\label{ss.whugait}

The whuGAIT Datasets~\cite{zou2018} were released in 2018 by the Wuhan University and made available along with the source code and pre-trained models to replicate the paper results\footnote{Available at https://github.com/qinnzou/Gait-Recognition-Using-Smartphones}. Unlike other data collections, whuGAIT comprises $3$D accelerometers and $3$-axis gyroscope information collected from $118$ people, $20$ of them collected during three days, and $98$ collected in just one day. The dataset is divided into $6$ different sub-sets, according to the desired task:

\begin{itemize}
	\item \textbf{Dataset \#1:} composed of $33,104$ samples for training and $3,740$ for testing from $118$ individuals, divided into a two-step segmentation.
	\item \textbf{Dataset \#2:} similar to Dataset \#1, comprises a two-step segmentation dataset composed of $49,275$ samples for training and $4,936$ for testing, extracted from the $3$ days collected data of $20$ people.
	\item \textbf{Dataset \#3:} this subset is divided into time size windows, comprising $2.56$ seconds for each sample. The set consists of $26,283$ instances used for training and $2,991$ employed for testing purposes.
	\item \textbf{Dataset \#4:} similar to Dataset \#3, this subset is divided into time frames of $2.56$ seconds, but using the data from $20$ individuals collected during three days. The subset comprises $35,373$ for training and  $3,941$ for testing purposes.
	\item \textbf{Dataset \#5:} the subset is employed for authentication purposes. It is composed of $74,142$ instances from $118$ people, such that information extracted from $98$ individuals is utilized for training, while the remaining $20$ are used for validation. The authentication procedure is compounded by a pair of samples from one or two different subjects. The instances comprise two-step acceleration and gyroscopic data.
	\item \textbf{Dataset \#6:} this subset employs the same structure used in Dataset \#5, but instead of horizontal align, it uses vertical alignment.
\end{itemize}

\subsection{Datasets in Usage Contexts}
\label{ss.datasetcontext}

This section presented the datasets most commonly employed for gait recognition and used in the works considered in this survey. Thus, in the sequence, we also provide an overview regarding particular aspects, such as usage contexts, acquisition environments, and spectrum. Table \ref{t.dbcontext} shows such information for each dataset. The covariates were divided into twelve main features, i.e., viewpoint, pace, objects, shoe, clothing, time, surface, silhouette, gait fluctuation, treadmill walking, overground walking, and foot pressure.

\begin{table}[H]
	\caption{Datasets in Usage Contexts.} 
	\label{t.dbcontext}
			\begin{tabular}{|l|c|c|l|c|c|l|c|l|l|l|}
				\hline
				\multicolumn{1}{|c|}{Context/Dataset} & \ref{ss.cmuMobo} & \ref{ss.tumGaid}  & \multicolumn{1}{c|}{\ref{ss.UMD}} & \ref{ss.casia} & \ref{ss.ouisir} & \multicolumn{1}{c|}{\ref{ss.usf}} & \ref{ss.usf}   & \multicolumn{1}{c|}{\ref{ss.AVAMVG}} & \multicolumn{1}{c|}{\ref{ss.KY4D}} & \multicolumn{1}{c|}{\ref{ss.whugait}} \\ \hline
				Viewpoint  & x  & \multicolumn{1}{l|}{} & \multicolumn{1}{c|}{x}  & x & x & \multicolumn{1}{c|}{x} & x & \multicolumn{1}{c|}{x} & \multicolumn{1}{c|}{x} & \multicolumn{1}{c|}{x} \\ \hline
				Pace & x & \multicolumn{1}{l|}{} & & x & x & \multicolumn{1}{c|}{x} & x & & & \multicolumn{1}{c|}{x} \\ \hline
				Objects & x & x & & x & \multicolumn{1}{l|}{} & & x & &  & \\ \hline
				Shoe & x & x & \multicolumn{1}{c|}{x} & x & x & \multicolumn{1}{c|}{x} & x & \multicolumn{1}{c|}{x} & & \\ \hline
				Clothing & x & x & \multicolumn{1}{c|}{x} & x & x & & x                     & \multicolumn{1}{c|}{x} & & \\ \hline
				Time & \multicolumn{1}{l|}{} & \multicolumn{1}{l|}{} & & x & x  &  & x & & \multicolumn{1}{c|}{x} & \multicolumn{1}{c|}{x} \\ \hline
				Surface & x & x & \multicolumn{1}{c|}{x} & x & \multicolumn{1}{l|}{} & \multicolumn{1}{c|}{x} & x & & \multicolumn{1}{c|}{x} & \multicolumn{1}{c|}{x} \\ \hline
				Silhouette & \multicolumn{1}{l|}{} & x  & & x & x & & \multicolumn{1}{l|}{} & & \multicolumn{1}{c|}{x} & \\ \hline
				Gait Fluctuation & \multicolumn{1}{l|}{} & \multicolumn{1}{l|}{} & & \multicolumn{1}{l|}{} & x &  & \multicolumn{1}{l|}{} & & & \\ \hline
				Treadmill walking & x & \multicolumn{1}{l|}{} & & \multicolumn{1}{l|}{} & x & & x  & & & \\ \hline
				Overground walking & x & x & \multicolumn{1}{c|}{x} & x & \multicolumn{1}{l|}{} & \multicolumn{1}{c|}{x} & \multicolumn{1}{l|}{} & \multicolumn{1}{c|}{x} & \multicolumn{1}{c|}{x} & \multicolumn{1}{c|}{x} \\ \hline
				Foot pressure & \multicolumn{1}{l|}{} & \multicolumn{1}{l|}{} & & x & \multicolumn{1}{l|}{} & & \multicolumn{1}{l|}{} & & & \\ \hline
			\end{tabular}
		\end{table}

\begin{figure}[!htb]
	
	\centering	
	\includegraphics[width=0.8\textwidth]{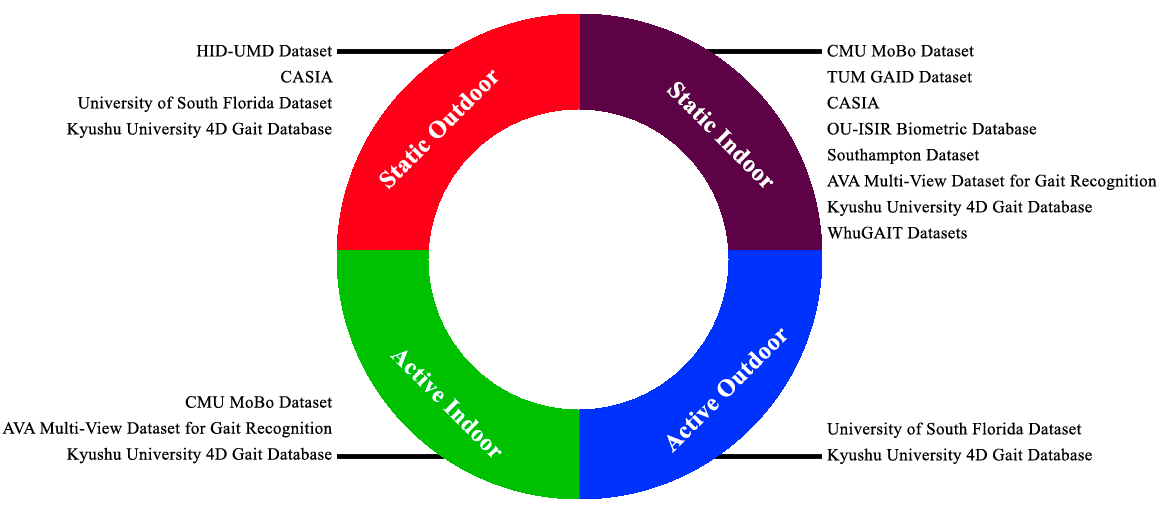}\hfill
	\caption{Representation of the dataset development environment.}
	\label{fig:environment}
	
\end{figure}

Further, Figure~\ref{fig:environment} presents the datasets' construction environments, which is divided into four scenarios, i.e., static indoor, static outdoor, active indoor, and active outdoor.

\begin{figure}[!htb]
	
	\centering	
	\includegraphics[width=.55\textwidth]{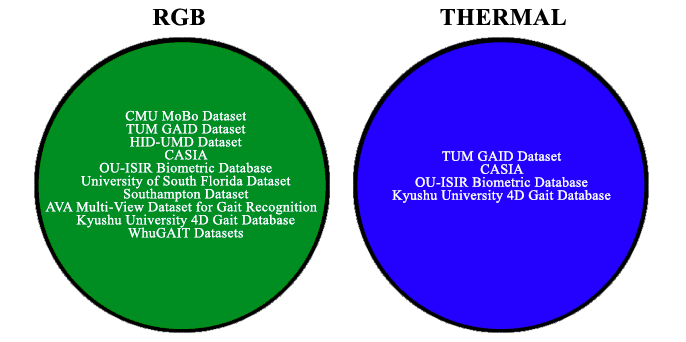}\hfill
	\caption{Data spectrum representation of the datasets.}
	\label{fig:spectrum}
	
\end{figure}

Finally, Figure~\ref{fig:spectrum} provides an illustrated schema categorizing the datasets' spectrum representation into two main classes, i.e., color-based and thermal information.

\section{Conclusions and Future Directions}
\label{sec:conclusions}

This survey provided an in-depth research towards the most significant works developed in the last years regarding gait recognition, highlighting deep learning techniques for such a task. Besides, it also provided a historical background concerning both methods for biometric identification, such as fingerprint, iris, and face, among others, as well as gait recognition, exposing the main concerns and challenges faced by the field.

Further, it provided a detailed description of the nine most employed datasets for the task, i.e., CMU MoBo~\cite{Gross-2001-8255}, TUM GAID~\cite{hofmann2014tum}, HID-UMD~\cite{kale2002framework,cuntoor2003combining}, CASIA~\cite{wang2003silhouette,casiab,tan2006efficient,zheng2012cascade}, OU-ISIR~\cite{tsuji2010silhouette,hossain2010clothing,makihara2010gait,mori2010gait,Iwama_IFS2012,mansur2014gait,takemura2018multi,uddin2018isir,Xu_CVA2017,ngo2014largest,ngo2014largest}, USF~\cite{sarkar2005humanid}, SOTON~\cite{nixon2001experimental,shutler2004large,matovski2010effect}, AVAMVG~\cite{AVAMVG}, KY4D~\cite{yumi2014gait}, and WhuGait~\cite{zou2018} as well as their extraction process and limitations. Regarding such datasets, one can clearly notice the efforts towards an detection independent from gender, angle in the scene, clothes and shoes, carry of bags and other itens, among other common issues.

The works surveyed in this paper discuss the main concerns and efforts made by the scientific community towards the development of techniques to guarantee access to monitored resources via gait identification and authorization. It also explores the advantages of a gait-based system against commonly employed biometric features, e.g., gait systems do not require the individuals' willingness to be identified. In this sense, the review attempted to point out the advantages of gait biometrics compared to other biometric techniques and highlight the most recent methodologies and state-of-the-art architectures.

There are a few critical points about video-based models and databases that are worth highlighting:
\begin{itemize}
	\item The presented datasets do not introduce more than one person per video, either for training or validation purposes.
	\item The environments in which they were recorded are fully controlled. Despite the variation of angles in some sets, it is possible to notice no variation of objects or even of colors in the background. Others videos were recorded with people walking on electric treadmills, denoting even greater control of the movements and the environment.
	\item Consequently, the presented models cited in the work are probably prone to show lack and misbehavior when applied to the real-world operation, such as detecting people of interest walking on public streets.
\end{itemize}

Regarding future trends, we expect an increase in the number of works related to:
\begin{enumerate}
 \item \textbf{Transformer networks}: the technique~\cite{transformers-original} targets data streaming such as video and fits the context of gait recognition perfectly. 
 \item \textbf{Gender and age recognition}: such an approach~\cite{sun2019deep} has gained popularity in the last years due to its importance in several applications.
\item \textbf{Hazardous environment monitoring}: gait features fit perfectly the purpose of monitoring hazardous environments since they usually are composed of limited illumination, and extracting more descriptive biometric identification characteristics denotes a difficult task~\cite{santanaIEEE-IS:19}.
\item \textbf{Multiple people in the scene}: most of the gait recognition works focus on a single individual in the scene in controlled environments. However, real-life problems usually require solutions robust to non-controlled environments with multiple people in the scene.
\end{enumerate}

Additionally, we observed an increasing demand for gait recognition on individuals with different clothes or carrying objects~\cite{yu2017invariant}, as well as hybrid approaches composed of both gait and complementary biometric characteristics, such as face and ear~\cite{mehraj2020feature}.

\begin{acks}
The authors are grateful to S\~ao Paulo Research Foundation (Fapesp) grants \#2013/07375-0, \#2014/12236-1, \#2019/07665-4, and \#2020/12101-0, to the Brazilian National Council for Research and Development (CNPq) \#307066/2017-7 and \#427968/2018-6, Eldorado Research Institute, as well as Petroleo Brasileiro S.A. (Petrobras) grant \#2017/00285-6.
\end{acks}

\bibliographystyle{ACM-Reference-Format}
\bibliography{references}


\begin{thebibliography}{141}


\ifx \showCODEN    \undefined \def \showCODEN     #1{\unskip}     \fi
\ifx \showDOI      \undefined \def \showDOI       #1{#1}\fi
\ifx \showISBNx    \undefined \def \showISBNx     #1{\unskip}     \fi
\ifx \showISBNxiii \undefined \def \showISBNxiii  #1{\unskip}     \fi
\ifx \showISSN     \undefined \def \showISSN      #1{\unskip}     \fi
\ifx \showLCCN     \undefined \def \showLCCN      #1{\unskip}     \fi
\ifx \shownote     \undefined \def \shownote      #1{#1}          \fi
\ifx \showarticletitle \undefined \def \showarticletitle #1{#1}   \fi
\ifx \showURL      \undefined \def \showURL       {\relax}        \fi
\providecommand\bibfield[2]{#2}
\providecommand\bibinfo[2]{#2}
\providecommand\natexlab[1]{#1}
\providecommand\showeprint[2][]{arXiv:#2}

\bibitem[\protect\citeauthoryear{Alariki, Faiz, Balagh, and Murray}{Alariki
  et~al\mbox{.}}{2018}]%
        {alariki2018review}
\bibfield{author}{\bibinfo{person}{Ala~Abdulhakim Alariki},
  \bibinfo{person}{Muqadas Faiz}, \bibinfo{person}{Sanaullah Balagh}, {and}
  \bibinfo{person}{Christine Murray}.} \bibinfo{year}{2018}\natexlab{}.
\newblock \showarticletitle{A Review of Finger-vein Biometric Recognition}. In
  \bibinfo{booktitle}{\emph{Proceedings of the World Congress on Engineering
  and Computer Science}}, Vol.~\bibinfo{volume}{1}.
\newblock


\bibitem[\protect\citeauthoryear{Alharthi, Yunas, and Ozanyan}{Alharthi
  et~al\mbox{.}}{2019}]%
        {alharthi2019deep}
\bibfield{author}{\bibinfo{person}{Abdullah~S Alharthi},
  \bibinfo{person}{Syed~U Yunas}, {and} \bibinfo{person}{Krikor~B Ozanyan}.}
  \bibinfo{year}{2019}\natexlab{}.
\newblock \showarticletitle{Deep learning for monitoring of human gait: A
  review}.
\newblock \bibinfo{journal}{\emph{IEEE Sensors Journal}} \bibinfo{volume}{19},
  \bibinfo{number}{21} (\bibinfo{year}{2019}), \bibinfo{pages}{9575--9591}.
\newblock


\bibitem[\protect\citeauthoryear{Alsmirat, Al-Alem, Al-Ayyoub, Jararweh, and
  Gupta}{Alsmirat et~al\mbox{.}}{2019}]%
        {alsmirat2019impact}
\bibfield{author}{\bibinfo{person}{Mohammad~A Alsmirat},
  \bibinfo{person}{Fatimah Al-Alem}, \bibinfo{person}{Mahmoud Al-Ayyoub},
  \bibinfo{person}{Yaser Jararweh}, {and} \bibinfo{person}{Brij Gupta}.}
  \bibinfo{year}{2019}\natexlab{}.
\newblock \showarticletitle{Impact of digital fingerprint image quality on the
  fingerprint recognition accuracy}.
\newblock \bibinfo{journal}{\emph{Multimedia Tools and Applications}}
  \bibinfo{volume}{78}, \bibinfo{number}{3} (\bibinfo{year}{2019}),
  \bibinfo{pages}{3649--3688}.
\newblock


\bibitem[\protect\citeauthoryear{Arora, Hanmandlu, and Srivastava}{Arora
  et~al\mbox{.}}{2015}]%
        {arora}
\bibfield{author}{\bibinfo{person}{P. Arora}, \bibinfo{person}{M. Hanmandlu},
  {and} \bibinfo{person}{S. Srivastava}.} \bibinfo{year}{2015}\natexlab{}.
\newblock \showarticletitle{Gait based authentication using gait information
  image features}.
\newblock \bibinfo{journal}{\emph{Pattern Recognition Letters}}
  \bibinfo{volume}{68} (\bibinfo{year}{2015}), \bibinfo{pages}{336–342}.
\newblock


\bibitem[\protect\citeauthoryear{Babaee, Li, and Rigoll}{Babaee
  et~al\mbox{.}}{2019}]%
        {babaee2019person}
\bibfield{author}{\bibinfo{person}{Maryam Babaee}, \bibinfo{person}{Linwei Li},
  {and} \bibinfo{person}{Gerhard Rigoll}.} \bibinfo{year}{2019}\natexlab{}.
\newblock \showarticletitle{Person identification from partial gait cycle using
  fully convolutional neural networks}.
\newblock \bibinfo{journal}{\emph{Neurocomputing}} (\bibinfo{year}{2019}).
\newblock


\bibitem[\protect\citeauthoryear{Bolle, Connell, Pankanti, Ratha, and
  Senior}{Bolle et~al\mbox{.}}{2013}]%
        {bolle2013guide}
\bibfield{author}{\bibinfo{person}{Ruud~M Bolle}, \bibinfo{person}{Jonathan~H
  Connell}, \bibinfo{person}{Sharath Pankanti}, \bibinfo{person}{Nalini~K
  Ratha}, {and} \bibinfo{person}{Andrew~W Senior}.}
  \bibinfo{year}{2013}\natexlab{}.
\newblock \bibinfo{booktitle}{\emph{Guide to biometrics}}.
\newblock \bibinfo{publisher}{Springer Science \& Business Media}.
\newblock


\bibitem[\protect\citeauthoryear{Bouchrika}{Bouchrika}{2018}]%
        {bouchrika2018survey}
\bibfield{author}{\bibinfo{person}{Imed Bouchrika}.}
  \bibinfo{year}{2018}\natexlab{}.
\newblock \showarticletitle{A survey of using biometrics for smart visual
  surveillance: Gait recognition}.
\newblock In \bibinfo{booktitle}{\emph{Surveillance in Action}}.
  \bibinfo{publisher}{Springer}, \bibinfo{pages}{3--23}.
\newblock


\bibitem[\protect\citeauthoryear{Chen, Dewi, Zhang, Liu, Huang,
  et~al\mbox{.}}{Chen et~al\mbox{.}}{2020}]%
        {chen2020integrating}
\bibfield{author}{\bibinfo{person}{Rung-Ching Chen}, \bibinfo{person}{Christine
  Dewi}, \bibinfo{person}{Wei-Wei Zhang}, \bibinfo{person}{Jia-Ming Liu},
  \bibinfo{person}{Su-Wen Huang}, {et~al\mbox{.}}}
  \bibinfo{year}{2020}\natexlab{}.
\newblock \showarticletitle{Integrating gesture control board and image
  recognition for gesture recognition based on deep learning}.
\newblock \bibinfo{journal}{\emph{International Journal of Applied Science and
  Engineering}} \bibinfo{volume}{17}, \bibinfo{number}{3}
  (\bibinfo{year}{2020}), \bibinfo{pages}{237--248}.
\newblock


\bibitem[\protect\citeauthoryear{Cho, Van~Merri{\"e}nboer, Bahdanau, and
  Bengio}{Cho et~al\mbox{.}}{2014}]%
        {gru}
\bibfield{author}{\bibinfo{person}{Kyunghyun Cho}, \bibinfo{person}{Bart
  Van~Merri{\"e}nboer}, \bibinfo{person}{Dzmitry Bahdanau}, {and}
  \bibinfo{person}{Yoshua Bengio}.} \bibinfo{year}{2014}\natexlab{}.
\newblock \showarticletitle{On the properties of neural machine translation:
  Encoder-decoder approaches}.
\newblock \bibinfo{journal}{\emph{arXiv preprint arXiv:1409.1259}}
  (\bibinfo{year}{2014}).
\newblock


\bibitem[\protect\citeauthoryear{Chung, Gulcehre, Cho, and Bengio}{Chung
  et~al\mbox{.}}{2014}]%
        {gru-study}
\bibfield{author}{\bibinfo{person}{Junyoung Chung}, \bibinfo{person}{Caglar
  Gulcehre}, \bibinfo{person}{KyungHyun Cho}, {and} \bibinfo{person}{Yoshua
  Bengio}.} \bibinfo{year}{2014}\natexlab{}.
\newblock \showarticletitle{Empirical evaluation of gated recurrent neural
  networks on sequence modeling}.
\newblock \bibinfo{journal}{\emph{arXiv preprint arXiv:1412.3555}}
  (\bibinfo{year}{2014}).
\newblock


\bibitem[\protect\citeauthoryear{Costilla-Reyes, Vera-Rodriguez, Scully, and
  Ozanyan}{Costilla-Reyes et~al\mbox{.}}{2018}]%
        {costilla2018analysis}
\bibfield{author}{\bibinfo{person}{Omar Costilla-Reyes}, \bibinfo{person}{Ruben
  Vera-Rodriguez}, \bibinfo{person}{Patricia Scully}, {and}
  \bibinfo{person}{Krikor~B Ozanyan}.} \bibinfo{year}{2018}\natexlab{}.
\newblock \showarticletitle{Analysis of spatio-temporal representations for
  robust footstep recognition with deep residual neural networks}.
\newblock \bibinfo{journal}{\emph{IEEE transactions on pattern analysis and
  machine intelligence}} \bibinfo{volume}{41}, \bibinfo{number}{2}
  (\bibinfo{year}{2018}), \bibinfo{pages}{285--296}.
\newblock


\bibitem[\protect\citeauthoryear{Cuntoor, Kale, and Chellappa}{Cuntoor
  et~al\mbox{.}}{2003}]%
        {cuntoor2003combining}
\bibfield{author}{\bibinfo{person}{Naresh Cuntoor}, \bibinfo{person}{Amit
  Kale}, {and} \bibinfo{person}{Rama Chellappa}.}
  \bibinfo{year}{2003}\natexlab{}.
\newblock \showarticletitle{Combining multiple evidences for gait recognition}.
  In \bibinfo{booktitle}{\emph{2003 IEEE International Conference on Acoustics,
  Speech, and Signal Processing, 2003. Proceedings.(ICASSP'03).}},
  Vol.~\bibinfo{volume}{3}. IEEE, \bibinfo{pages}{III--33}.
\newblock


\bibitem[\protect\citeauthoryear{Dargan and Kumar}{Dargan and Kumar}{2020}]%
        {dargan2020comprehensive}
\bibfield{author}{\bibinfo{person}{Shaveta Dargan} {and}
  \bibinfo{person}{Munish Kumar}.} \bibinfo{year}{2020}\natexlab{}.
\newblock \showarticletitle{A comprehensive survey on the biometric recognition
  systems based on physiological and behavioral modalities}.
\newblock \bibinfo{journal}{\emph{Expert Systems with Applications}}
  \bibinfo{volume}{143} (\bibinfo{year}{2020}), \bibinfo{pages}{113114}.
\newblock


\bibitem[\protect\citeauthoryear{{David L\'opez-Fern\'andez, Francisco J.
  Madrid-Cuevas, \'Angel Carmona-Poyato, Manuel J. Mar\'in-Jiménez and Rafael
  Mu\~noz-Salinas}}{{David L\'opez-Fern\'andez, Francisco J. Madrid-Cuevas,
  \'Angel Carmona-Poyato, Manuel J. Mar\'in-Jiménez and Rafael
  Mu\~noz-Salinas}}{2014}]%
        {AVAMVG}
\bibfield{author}{\bibinfo{person}{{David L\'opez-Fern\'andez, Francisco J.
  Madrid-Cuevas, \'Angel Carmona-Poyato, Manuel J. Mar\'in-Jiménez and Rafael
  Mu\~noz-Salinas}}.} \bibinfo{year}{2014}\natexlab{}.
\newblock \showarticletitle{{The AVA Multi-View Dataset for Gait Recognition}}.
\newblock In \bibinfo{booktitle}{\emph{Activity Monitoring by Multiple
  Distributed Sensing}}, \bibfield{editor}{\bibinfo{person}{Pier~Luigi Mazzeo},
  \bibinfo{person}{Paolo Spagnolo}, {and} \bibinfo{person}{Thomas~B. Moeslund}}
  (Eds.). \bibinfo{publisher}{Springer International Publishing},
  \bibinfo{pages}{26--39}.
\newblock
\showISBNx{978-3-319-13322-5}
\urldef\tempurl%
\url{https://doi.org/10.1007/978-3-319-13323-2_3}
\showDOI{\tempurl}


\bibitem[\protect\citeauthoryear{de~Souza, Silva, Passos, Roder, Santana,
  Pinheiro, and de~Albuquerque}{de~Souza et~al\mbox{.}}{2021}]%
        {Souza2021computer}
\bibfield{author}{\bibinfo{person}{Renato~WR de Souza},
  \bibinfo{person}{Daniel~S Silva}, \bibinfo{person}{Leandro~A Passos},
  \bibinfo{person}{Mateus Roder}, \bibinfo{person}{Marcos~C Santana},
  \bibinfo{person}{Pl{\'a}cido~R Pinheiro}, {and} \bibinfo{person}{Victor
  Hugo~C de Albuquerque}.} \bibinfo{year}{2021}\natexlab{}.
\newblock \showarticletitle{Computer-Assisted Parkinson's Disease Diagnosis
  Using Fuzzy Optimum-Path Forest and Restricted Boltzmann Machines}.
\newblock \bibinfo{journal}{\emph{Computers in Biology and Medicine}}
  (\bibinfo{year}{2021}), \bibinfo{pages}{104260}.
\newblock


\bibitem[\protect\citeauthoryear{Deng et~al\mbox{.}}{Deng
  et~al\mbox{.}}{2018a}]%
        {deng}
\bibfield{author}{\bibinfo{person}{Muqing Deng} {et~al\mbox{.}}}
  \bibinfo{year}{2018}\natexlab{a}.
\newblock \showarticletitle{Gait recognition under different clothing
  conditions via deterministic learning}.
\newblock \bibinfo{journal}{\emph{in IEEE/CAA Journal of Automatica Sinica}}
  (\bibinfo{year}{2018}).
\newblock


\bibitem[\protect\citeauthoryear{Deng, Hu, and Guo}{Deng
  et~al\mbox{.}}{2018b}]%
        {deng2018face}
\bibfield{author}{\bibinfo{person}{Weihong Deng}, \bibinfo{person}{Jiani Hu},
  {and} \bibinfo{person}{Jun Guo}.} \bibinfo{year}{2018}\natexlab{b}.
\newblock \showarticletitle{Face recognition via collaborative representation:
  Its discriminant nature and superposed representation}.
\newblock \bibinfo{journal}{\emph{IEEE transactions on pattern analysis and
  machine intelligence}} \bibinfo{volume}{40}, \bibinfo{number}{10}
  (\bibinfo{year}{2018}), \bibinfo{pages}{2513--2521}.
\newblock


\bibitem[\protect\citeauthoryear{Deng, Hu, Lu, and Guo}{Deng
  et~al\mbox{.}}{2014}]%
        {deng2014transform}
\bibfield{author}{\bibinfo{person}{Weihong Deng}, \bibinfo{person}{Jiani Hu},
  \bibinfo{person}{Jiwen Lu}, {and} \bibinfo{person}{Jun Guo}.}
  \bibinfo{year}{2014}\natexlab{}.
\newblock \showarticletitle{Transform-invariant PCA: A unified approach to
  fully automatic facealignment, representation, and recognition}.
\newblock \bibinfo{journal}{\emph{IEEE transactions on pattern analysis and
  machine intelligence}} \bibinfo{volume}{36}, \bibinfo{number}{6}
  (\bibinfo{year}{2014}), \bibinfo{pages}{1275--1284}.
\newblock


\bibitem[\protect\citeauthoryear{Dewi and Chen}{Dewi and Chen}{2019}]%
        {dewi2019human}
\bibfield{author}{\bibinfo{person}{Christine Dewi} {and}
  \bibinfo{person}{Rung-Ching Chen}.} \bibinfo{year}{2019}\natexlab{}.
\newblock \showarticletitle{Human activity recognition based on evolution of
  features selection and random Forest}. In \bibinfo{booktitle}{\emph{2019 IEEE
  international conference on systems, man and cybernetics (SMC)}}. IEEE,
  \bibinfo{pages}{2496--2501}.
\newblock


\bibitem[\protect\citeauthoryear{e~MS~Nixon}{e~MS~Nixon}{2011}]%
        {nixon}
\bibfield{author}{\bibinfo{person}{G.~Ariyanto e MS~Nixon}.}
  \bibinfo{year}{2011}\natexlab{}.
\newblock \showarticletitle{Model-based 3D gait biometrics}.
\newblock \bibinfo{journal}{\emph{in IJCB}} (\bibinfo{year}{2011}).
\newblock


\bibitem[\protect\citeauthoryear{Fernandes, Fonseca, Ferreira, Gago, Costa,
  Sousa, Ferreira, Gama, Erlhagen, and Bicho}{Fernandes et~al\mbox{.}}{2018}]%
        {fernandes2018artificial}
\bibfield{author}{\bibinfo{person}{Carlos Fernandes},
  \bibinfo{person}{Lu{\'\i}s Fonseca}, \bibinfo{person}{Flora Ferreira},
  \bibinfo{person}{Miguel Gago}, \bibinfo{person}{Lus Costa},
  \bibinfo{person}{Nuno Sousa}, \bibinfo{person}{Carlos Ferreira},
  \bibinfo{person}{Joao Gama}, \bibinfo{person}{Wolfram Erlhagen}, {and}
  \bibinfo{person}{Estela Bicho}.} \bibinfo{year}{2018}\natexlab{}.
\newblock \showarticletitle{Artificial neural networks classification of
  patients with parkinsonism based on gait}. In \bibinfo{booktitle}{\emph{2018
  IEEE International Conference on Bioinformatics and Biomedicine (BIBM)}}.
  IEEE, \bibinfo{pages}{2024--2030}.
\newblock


\bibitem[\protect\citeauthoryear{Garagad and Iyer}{Garagad and Iyer}{2014}]%
        {garagad2014novel}
\bibfield{author}{\bibinfo{person}{Vishwanath~G Garagad} {and}
  \bibinfo{person}{Nalini~C Iyer}.} \bibinfo{year}{2014}\natexlab{}.
\newblock \showarticletitle{A novel technique of iris identification for
  biometric systems}. In \bibinfo{booktitle}{\emph{2014 International
  Conference on Advances in Computing, Communications and Informatics
  (ICACCI)}}. IEEE, \bibinfo{pages}{973--978}.
\newblock


\bibitem[\protect\citeauthoryear{Goodfellow, Pouget-Abadie, Mirza, Xu,
  Warde-Farley, Ozair, Courville, and Bengio}{Goodfellow et~al\mbox{.}}{2014}]%
        {gan}
\bibfield{author}{\bibinfo{person}{Ian~J Goodfellow}, \bibinfo{person}{Jean
  Pouget-Abadie}, \bibinfo{person}{Mehdi Mirza}, \bibinfo{person}{Bing Xu},
  \bibinfo{person}{David Warde-Farley}, \bibinfo{person}{Sherjil Ozair},
  \bibinfo{person}{Aaron Courville}, {and} \bibinfo{person}{Yoshua Bengio}.}
  \bibinfo{year}{2014}\natexlab{}.
\newblock \showarticletitle{Generative adversarial networks}.
\newblock \bibinfo{journal}{\emph{arXiv preprint arXiv:1406.2661}}
  (\bibinfo{year}{2014}).
\newblock


\bibitem[\protect\citeauthoryear{Gross and Shi}{Gross and Shi}{2001}]%
        {Gross-2001-8255}
\bibfield{author}{\bibinfo{person}{Ralph Gross} {and} \bibinfo{person}{Jianbo
  Shi}.} \bibinfo{year}{2001}\natexlab{}.
\newblock \bibinfo{booktitle}{\emph{The CMU Motion of Body (MoBo) Database}}.
\newblock \bibinfo{type}{{T}echnical {R}eport} CMU-RI-TR-01-18.
  \bibinfo{institution}{Carnegie Mellon University}.
\newblock


\bibitem[\protect\citeauthoryear{Gui, Ruiz-Blondet, Laszlo, and Jin}{Gui
  et~al\mbox{.}}{2019}]%
        {gui2019survey}
\bibfield{author}{\bibinfo{person}{Qiong Gui}, \bibinfo{person}{Maria~V
  Ruiz-Blondet}, \bibinfo{person}{Sarah Laszlo}, {and}
  \bibinfo{person}{Zhanpeng Jin}.} \bibinfo{year}{2019}\natexlab{}.
\newblock \showarticletitle{A survey on brain biometrics}.
\newblock \bibinfo{journal}{\emph{ACM Computing Surveys (CSUR)}}
  \bibinfo{volume}{51}, \bibinfo{number}{6} (\bibinfo{year}{2019}),
  \bibinfo{pages}{1--38}.
\newblock


\bibitem[\protect\citeauthoryear{Han and Bhanu}{Han and Bhanu}{2016}]%
        {bhanu}
\bibfield{author}{\bibinfo{person}{Ju Han} {and} \bibinfo{person}{Bir Bhanu}.}
  \bibinfo{year}{2016}\natexlab{}.
\newblock \showarticletitle{Individual Recognition UsingGait Energy Image}.
\newblock \bibinfo{journal}{\emph{IEEE TRANSACTIONS ON PATTERN ANALYSIS AND
  MACHINE INTELLIGENCE}} (\bibinfo{year}{2016}).
\newblock


\bibitem[\protect\citeauthoryear{He, Yan, and Arora}{He et~al\mbox{.}}{2020}]%
        {he2020long}
\bibfield{author}{\bibinfo{person}{Jing~Selena He}, \bibinfo{person}{Mingyuan
  Yan}, {and} \bibinfo{person}{Sahil Arora}.} \bibinfo{year}{2020}\natexlab{}.
\newblock \showarticletitle{Long Range Iris Recognition a Reality or a Myth? A
  Survey}. In \bibinfo{booktitle}{\emph{Proceedings of the 2020 ACM Southeast
  Conference}}. \bibinfo{pages}{305--306}.
\newblock


\bibitem[\protect\citeauthoryear{He, Zhang, Shan, and Wang}{He
  et~al\mbox{.}}{2018}]%
        {he2018multi}
\bibfield{author}{\bibinfo{person}{Yiwei He}, \bibinfo{person}{Junping Zhang},
  \bibinfo{person}{Hongming Shan}, {and} \bibinfo{person}{Liang Wang}.}
  \bibinfo{year}{2018}\natexlab{}.
\newblock \showarticletitle{Multi-task GANs for view-specific feature learning
  in gait recognition}.
\newblock \bibinfo{journal}{\emph{IEEE Transactions on Information Forensics
  and Security}} \bibinfo{volume}{14}, \bibinfo{number}{1}
  (\bibinfo{year}{2018}), \bibinfo{pages}{102--113}.
\newblock


\bibitem[\protect\citeauthoryear{Hemalatha}{Hemalatha}{2020}]%
        {hemalatha2020systematic}
\bibfield{author}{\bibinfo{person}{S Hemalatha}.}
  \bibinfo{year}{2020}\natexlab{}.
\newblock \showarticletitle{A systematic review on Fingerprint based Biometric
  Authentication System}. In \bibinfo{booktitle}{\emph{2020 International
  Conference on Emerging Trends in Information Technology and Engineering
  (ic-ETITE)}}. IEEE, \bibinfo{pages}{1--4}.
\newblock


\bibitem[\protect\citeauthoryear{Hinton}{Hinton}{2002}]%
        {Hinton:02}
\bibfield{author}{\bibinfo{person}{G.~E. Hinton}.}
  \bibinfo{year}{2002}\natexlab{}.
\newblock \showarticletitle{Training Products of Experts by Minimizing
  Contrastive Divergence}.
\newblock \bibinfo{journal}{\emph{Neural Computation}} \bibinfo{volume}{14},
  \bibinfo{number}{8} (\bibinfo{year}{2002}), \bibinfo{pages}{1771--1800}.
\newblock


\bibitem[\protect\citeauthoryear{Hinton, Osindero, and Teh}{Hinton
  et~al\mbox{.}}{2006}]%
        {Hinton:06}
\bibfield{author}{\bibinfo{person}{G.~E. Hinton}, \bibinfo{person}{S.
  Osindero}, {and} \bibinfo{person}{Y.-W. Teh}.}
  \bibinfo{year}{2006}\natexlab{}.
\newblock \showarticletitle{A Fast Learning Algorithm for Deep Belief Nets}.
\newblock \bibinfo{journal}{\emph{Neural Computation}} \bibinfo{volume}{18},
  \bibinfo{number}{7} (\bibinfo{year}{2006}), \bibinfo{pages}{1527--1554}.
\newblock
\showISSN{0899-7667}


\bibitem[\protect\citeauthoryear{Hochreiter and Schmidhuber}{Hochreiter and
  Schmidhuber}{1997}]%
        {Hochreiter}
\bibfield{author}{\bibinfo{person}{Sepp Hochreiter} {and}
  \bibinfo{person}{J{\"u}rgen Schmidhuber}.} \bibinfo{year}{1997}\natexlab{}.
\newblock \showarticletitle{Long short-term memory}.
\newblock \bibinfo{journal}{\emph{Neural computation}} \bibinfo{volume}{9},
  \bibinfo{number}{8} (\bibinfo{year}{1997}), \bibinfo{pages}{1735--1780}.
\newblock


\bibitem[\protect\citeauthoryear{Hofmann, Geiger, Bachmann, Schuller, and
  Rigoll}{Hofmann et~al\mbox{.}}{2014}]%
        {hofmann2014tum}
\bibfield{author}{\bibinfo{person}{Martin Hofmann}, \bibinfo{person}{J{\"u}rgen
  Geiger}, \bibinfo{person}{Sebastian Bachmann}, \bibinfo{person}{Bj{\"o}rn
  Schuller}, {and} \bibinfo{person}{Gerhard Rigoll}.}
  \bibinfo{year}{2014}\natexlab{}.
\newblock \showarticletitle{The tum gait from audio, image and depth (gaid)
  database: Multimodal recognition of subjects and traits}.
\newblock \bibinfo{journal}{\emph{Journal of Visual Communication and Image
  Representation}} \bibinfo{volume}{25}, \bibinfo{number}{1}
  (\bibinfo{year}{2014}), \bibinfo{pages}{195--206}.
\newblock


\bibitem[\protect\citeauthoryear{Hossain, Makihara, Wang, and Yagi}{Hossain
  et~al\mbox{.}}{2010}]%
        {hossain2010clothing}
\bibfield{author}{\bibinfo{person}{Md~Altab Hossain}, \bibinfo{person}{Yasushi
  Makihara}, \bibinfo{person}{Junqiu Wang}, {and} \bibinfo{person}{Yasushi
  Yagi}.} \bibinfo{year}{2010}\natexlab{}.
\newblock \showarticletitle{Clothing-invariant gait identification using
  part-based clothing categorization and adaptive weight control}.
\newblock \bibinfo{journal}{\emph{Pattern Recognition}} \bibinfo{volume}{43},
  \bibinfo{number}{6} (\bibinfo{year}{2010}), \bibinfo{pages}{2281--2291}.
\newblock


\bibitem[\protect\citeauthoryear{Hu, Guan, Gao, Long, Lane, and Ploetz}{Hu
  et~al\mbox{.}}{2018}]%
        {hu2018robust}
\bibfield{author}{\bibinfo{person}{BingZhang Hu}, \bibinfo{person}{Yu Guan},
  \bibinfo{person}{Yan Gao}, \bibinfo{person}{Yang Long},
  \bibinfo{person}{Nicholas Lane}, {and} \bibinfo{person}{Thomas Ploetz}.}
  \bibinfo{year}{2018}\natexlab{}.
\newblock \showarticletitle{Robust Cross-View Gait Recognition with Evidence: A
  Discriminant Gait GAN (DiGGAN) Approach}.
\newblock \bibinfo{journal}{\emph{arXiv preprint arXiv:1811.10493}}
  (\bibinfo{year}{2018}).
\newblock


\bibitem[\protect\citeauthoryear{Iwama, Okumura, Makihara, and Yagi}{Iwama
  et~al\mbox{.}}{2012}]%
        {Iwama_IFS2012}
\bibfield{author}{\bibinfo{person}{H. Iwama}, \bibinfo{person}{M. Okumura},
  \bibinfo{person}{Y. Makihara}, {and} \bibinfo{person}{Y. Yagi}.}
  \bibinfo{year}{2012}\natexlab{}.
\newblock \showarticletitle{The OU-ISIR Gait Database Comprising the Large
  Population Dataset and Performance Evaluation of Gait Recognition}.
\newblock \bibinfo{journal}{\emph{IEEE Trans. on Information Forensics and
  Security}}  \bibinfo{volume}{7, Issue 5} (\bibinfo{date}{Oct.}
  \bibinfo{year}{2012}), \bibinfo{pages}{1511--1521}.
\newblock


\bibitem[\protect\citeauthoryear{Iwashita, Kurazume, and Stoica}{Iwashita
  et~al\mbox{.}}{2014a}]%
        {yumi2014IRgait}
\bibfield{author}{\bibinfo{person}{Y. Iwashita}, \bibinfo{person}{R. Kurazume},
  {and} \bibinfo{person}{A. Stoica}.} \bibinfo{year}{2014}\natexlab{a}.
\newblock \showarticletitle{Gait Identification Using Invisible Shadows:
  Robustness to Appearance Changes}. In \bibinfo{booktitle}{\emph{Int. Conf.
  Emerging Security Technologies (EST)}}. \bibinfo{address}{UK}.
\newblock


\bibitem[\protect\citeauthoryear{Iwashita, Ogawara, and Kurazume}{Iwashita
  et~al\mbox{.}}{2014b}]%
        {yumi2014gait}
\bibfield{author}{\bibinfo{person}{Y. Iwashita}, \bibinfo{person}{K. Ogawara},
  {and} \bibinfo{person}{R. Kurazume}.} \bibinfo{year}{2014}\natexlab{b}.
\newblock \showarticletitle{Identification of people walking along curved
  trajectories}. In \bibinfo{booktitle}{\emph{Pattern Recognition Letters}}.
\newblock


\bibitem[\protect\citeauthoryear{Jain et~al\mbox{.}}{Jain
  et~al\mbox{.}}{2004}]%
        {jain}
\bibfield{author}{\bibinfo{person}{Anil~K Jain} {et~al\mbox{.}}}
  \bibinfo{year}{2004}\natexlab{}.
\newblock \showarticletitle{An Introduction to Biometric Recognition}.
\newblock \bibinfo{journal}{\emph{in IEEE Transactions on Circuits and Systems
  for Video Technology, Special Issue on Image- and Video-Based Biometrics,
  Vol. 14, No. 1}} (\bibinfo{year}{2004}).
\newblock


\bibitem[\protect\citeauthoryear{Jayaraman, Gupta, Gupta, Arora, and
  Tiwari}{Jayaraman et~al\mbox{.}}{2020}]%
        {jayaraman2020recent}
\bibfield{author}{\bibinfo{person}{Umarani Jayaraman},
  \bibinfo{person}{Phalguni Gupta}, \bibinfo{person}{Sandesh Gupta},
  \bibinfo{person}{Geetika Arora}, {and} \bibinfo{person}{Kamlesh Tiwari}.}
  \bibinfo{year}{2020}\natexlab{}.
\newblock \showarticletitle{Recent development in face recognition}.
\newblock \bibinfo{journal}{\emph{Neurocomputing}}  \bibinfo{volume}{408}
  (\bibinfo{year}{2020}), \bibinfo{pages}{231--245}.
\newblock


\bibitem[\protect\citeauthoryear{Jia, Yang, Huang, and Wang}{Jia
  et~al\mbox{.}}{2019}]%
        {jia2019attacking}
\bibfield{author}{\bibinfo{person}{Meijuan Jia}, \bibinfo{person}{Hongyu Yang},
  \bibinfo{person}{Di Huang}, {and} \bibinfo{person}{Yunhong Wang}.}
  \bibinfo{year}{2019}\natexlab{}.
\newblock \showarticletitle{Attacking gait recognition systems via silhouette
  guided GANs}. In \bibinfo{booktitle}{\emph{Proceedings of the 27th ACM
  International Conference on Multimedia}}. \bibinfo{pages}{638--646}.
\newblock


\bibitem[\protect\citeauthoryear{Jing~Luo and Xiu}{Jing~Luo and Xiu}{2015}]%
        {grgei}
\bibfield{author}{\bibinfo{person}{Chunyuan Zi Ying Niu Huixin~Tian Jing~Luo,
  Jianliang~Zhang} {and} \bibinfo{person}{Chunbo Xiu}.}
  \bibinfo{year}{2015}\natexlab{}.
\newblock \showarticletitle{Gait Recognition Using GEI and AFDEI}.
\newblock \bibinfo{journal}{\emph{International Journal of Optics}}
  (\bibinfo{year}{2015}).
\newblock


\bibitem[\protect\citeauthoryear{{Jun}, {Lee}, {Lee}, {Lee}, and {Kim}}{{Jun}
  et~al\mbox{.}}{2020}]%
        {8963659}
\bibfield{author}{\bibinfo{person}{K. {Jun}}, \bibinfo{person}{D. {Lee}},
  \bibinfo{person}{K. {Lee}}, \bibinfo{person}{S. {Lee}}, {and}
  \bibinfo{person}{M.~S. {Kim}}.} \bibinfo{year}{2020}\natexlab{}.
\newblock \showarticletitle{Feature Extraction Using an RNN Autoencoder for
  Skeleton-Based Abnormal Gait Recognition}.
\newblock \bibinfo{journal}{\emph{IEEE Access}}  \bibinfo{volume}{8}
  (\bibinfo{year}{2020}), \bibinfo{pages}{19196--19207}.
\newblock


\bibitem[\protect\citeauthoryear{Kale, Cuntoor, and Chellappa}{Kale
  et~al\mbox{.}}{2002}]%
        {kale2002framework}
\bibfield{author}{\bibinfo{person}{Amit Kale}, \bibinfo{person}{Naresh
  Cuntoor}, {and} \bibinfo{person}{Rama Chellappa}.}
  \bibinfo{year}{2002}\natexlab{}.
\newblock \showarticletitle{A framework for activity-specific human
  identification}. In \bibinfo{booktitle}{\emph{2002 IEEE International
  Conference on Acoustics, Speech, and Signal Processing}},
  Vol.~\bibinfo{volume}{4}. IEEE, \bibinfo{pages}{IV--3660}.
\newblock


\bibitem[\protect\citeauthoryear{Kan, Shan, Chang, and Chen}{Kan
  et~al\mbox{.}}{2014}]%
        {kan2014stacked}
\bibfield{author}{\bibinfo{person}{Meina Kan}, \bibinfo{person}{Shiguang Shan},
  \bibinfo{person}{Hong Chang}, {and} \bibinfo{person}{Xilin Chen}.}
  \bibinfo{year}{2014}\natexlab{}.
\newblock \showarticletitle{Stacked progressive auto-encoders (spae) for face
  recognition across poses}. In \bibinfo{booktitle}{\emph{Proceedings of the
  IEEE Conference on Computer Vision and Pattern Recognition}}.
  \bibinfo{pages}{1883--1890}.
\newblock


\bibitem[\protect\citeauthoryear{Karn, He, Zhang, and Zhang}{Karn
  et~al\mbox{.}}{2020}]%
        {karn2020experimental}
\bibfield{author}{\bibinfo{person}{Pradeep Karn}, \bibinfo{person}{XiaoHai He},
  \bibinfo{person}{Jin Zhang}, {and} \bibinfo{person}{Yanteng Zhang}.}
  \bibinfo{year}{2020}\natexlab{}.
\newblock \showarticletitle{An experimental study of relative total variation
  and probabilistic collaborative representation for iris recognition}.
\newblock \bibinfo{journal}{\emph{Multimedia Tools and Applications}}
  \bibinfo{volume}{79}, \bibinfo{number}{43} (\bibinfo{year}{2020}),
  \bibinfo{pages}{31783--31801}.
\newblock


\bibitem[\protect\citeauthoryear{Khan, Parkinson, Grant, Liu, and Mcguire}{Khan
  et~al\mbox{.}}{2020}]%
        {khan2020biometric}
\bibfield{author}{\bibinfo{person}{Saad Khan}, \bibinfo{person}{Simon
  Parkinson}, \bibinfo{person}{Liam Grant}, \bibinfo{person}{Na Liu}, {and}
  \bibinfo{person}{Stephen Mcguire}.} \bibinfo{year}{2020}\natexlab{}.
\newblock \showarticletitle{Biometric systems utilising health data from
  wearable devices: applications and future challenges in computer security}.
\newblock \bibinfo{journal}{\emph{ACM Computing Surveys (CSUR)}}
  \bibinfo{volume}{53}, \bibinfo{number}{4} (\bibinfo{year}{2020}),
  \bibinfo{pages}{1--29}.
\newblock


\bibitem[\protect\citeauthoryear{Krizhevsky, Sutskever, and Hinton}{Krizhevsky
  et~al\mbox{.}}{2012}]%
        {krizhevsky2012imagenet}
\bibfield{author}{\bibinfo{person}{Alex Krizhevsky}, \bibinfo{person}{Ilya
  Sutskever}, {and} \bibinfo{person}{Geoffrey~E Hinton}.}
  \bibinfo{year}{2012}\natexlab{}.
\newblock \showarticletitle{Imagenet classification with deep convolutional
  neural networks}. In \bibinfo{booktitle}{\emph{Advances in neural information
  processing systems}}. \bibinfo{pages}{1097--1105}.
\newblock


\bibitem[\protect\citeauthoryear{LeCun, Bottou, Bengio, Haffner,
  et~al\mbox{.}}{LeCun et~al\mbox{.}}{1998}]%
        {lecunLeNet:1998}
\bibfield{author}{\bibinfo{person}{Yann LeCun}, \bibinfo{person}{L{\'e}on
  Bottou}, \bibinfo{person}{Yoshua Bengio}, \bibinfo{person}{Patrick Haffner},
  {et~al\mbox{.}}} \bibinfo{year}{1998}\natexlab{}.
\newblock \showarticletitle{Gradient-based learning applied to document
  recognition}.
\newblock \bibinfo{journal}{\emph{Proc. IEEE}} \bibinfo{volume}{86},
  \bibinfo{number}{11} (\bibinfo{year}{1998}), \bibinfo{pages}{2278--2324}.
\newblock


\bibitem[\protect\citeauthoryear{Lee, Cho, Choi, and Kim}{Lee
  et~al\mbox{.}}{2017}]%
        {lee2017partial}
\bibfield{author}{\bibinfo{person}{Wonjune Lee}, \bibinfo{person}{Sungchul
  Cho}, \bibinfo{person}{Heeseung Choi}, {and} \bibinfo{person}{Jaihie Kim}.}
  \bibinfo{year}{2017}\natexlab{}.
\newblock \showarticletitle{Partial fingerprint matching using minutiae and
  ridge shape features for small fingerprint scanners}.
\newblock \bibinfo{journal}{\emph{Expert Systems with Applications: An
  International Journal}} \bibinfo{volume}{87}, \bibinfo{number}{C}
  (\bibinfo{year}{2017}), \bibinfo{pages}{183--198}.
\newblock


\bibitem[\protect\citeauthoryear{Lei, Pietik{\"a}inen, and Li}{Lei
  et~al\mbox{.}}{2014}]%
        {lei2014learning}
\bibfield{author}{\bibinfo{person}{Zhen Lei}, \bibinfo{person}{Matti
  Pietik{\"a}inen}, {and} \bibinfo{person}{Stan~Z Li}.}
  \bibinfo{year}{2014}\natexlab{}.
\newblock \showarticletitle{Learning discriminant face descriptor}.
\newblock \bibinfo{journal}{\emph{IEEE Transactions on Pattern Analysis and
  Machine Intelligence}} \bibinfo{volume}{36}, \bibinfo{number}{2}
  (\bibinfo{year}{2014}), \bibinfo{pages}{289--302}.
\newblock


\bibitem[\protect\citeauthoryear{Li, Min, Sun, Lin, and Tang}{Li
  et~al\mbox{.}}{2017}]%
        {Li:2017Deepgait}
\bibfield{author}{\bibinfo{person}{C. Li}, \bibinfo{person}{X. Min},
  \bibinfo{person}{S. Sun}, \bibinfo{person}{W. Lin}, {and} \bibinfo{person}{Z.
  Tang}.} \bibinfo{year}{2017}\natexlab{}.
\newblock \showarticletitle{Deepgait: A learning deep convolutional
  representation for view-invariant gait recognition using joint bayesian}.
\newblock \bibinfo{journal}{\emph{Applied Sciences}} \bibinfo{volume}{7},
  \bibinfo{number}{3} (\bibinfo{year}{2017}), \bibinfo{pages}{210}.
\newblock


\bibitem[\protect\citeauthoryear{Li, Zhang, Fei, and Zhao}{Li
  et~al\mbox{.}}{2021}]%
        {li2021joint}
\bibfield{author}{\bibinfo{person}{Shuyi Li}, \bibinfo{person}{Bob Zhang},
  \bibinfo{person}{Lunke Fei}, {and} \bibinfo{person}{Shuping Zhao}.}
  \bibinfo{year}{2021}\natexlab{}.
\newblock \showarticletitle{Joint discriminative feature learning for
  multimodal finger recognition}.
\newblock \bibinfo{journal}{\emph{Pattern Recognition}}  \bibinfo{volume}{111}
  (\bibinfo{year}{2021}), \bibinfo{pages}{107704}.
\newblock


\bibitem[\protect\citeauthoryear{Li}{Li}{2009}]%
        {li2009encyclopedia}
\bibfield{author}{\bibinfo{person}{Stan~Z Li}.}
  \bibinfo{year}{2009}\natexlab{}.
\newblock \bibinfo{booktitle}{\emph{Encyclopedia of Biometrics: I-Z.}}
  Vol.~\bibinfo{volume}{2}.
\newblock \bibinfo{publisher}{Springer Science \& Business Media}.
\newblock


\bibitem[\protect\citeauthoryear{Lobo, Passos, and Papa}{Lobo
  et~al\mbox{.}}{2020}]%
        {LoboICCS:20}
\bibfield{author}{\bibinfo{person}{V.~C. Lobo}, \bibinfo{person}{L.~A. Passos},
  {and} \bibinfo{person}{J.~P. Papa}.} \bibinfo{year}{2020}\natexlab{}.
\newblock \showarticletitle{Evolving Long Short-Term Memory Networks}. In
  \bibinfo{booktitle}{\emph{International Conference on Computational Science
  (ICCS)}}. IEEE.
\newblock


\bibitem[\protect\citeauthoryear{Luo and Tjahjadi}{Luo and Tjahjadi}{2020}]%
        {luo2020multi}
\bibfield{author}{\bibinfo{person}{Jian Luo} {and} \bibinfo{person}{Tardi
  Tjahjadi}.} \bibinfo{year}{2020}\natexlab{}.
\newblock \showarticletitle{Multi-Set Canonical Correlation Analysis for 3D
  Abnormal Gait Behaviour Recognition Based on Virtual Sample Generation}.
\newblock \bibinfo{journal}{\emph{IEEE Access}}  \bibinfo{volume}{8}
  (\bibinfo{year}{2020}), \bibinfo{pages}{32485--32501}.
\newblock


\bibitem[\protect\citeauthoryear{Luo, Yang, and Liu}{Luo et~al\mbox{.}}{2016}]%
        {luo2016gait}
\bibfield{author}{\bibinfo{person}{Zhengping Luo}, \bibinfo{person}{Tianqi
  Yang}, {and} \bibinfo{person}{Yanjun Liu}.} \bibinfo{year}{2016}\natexlab{}.
\newblock \showarticletitle{Gait optical flow image decomposition for human
  recognition}. In \bibinfo{booktitle}{\emph{2016 IEEE Information Technology,
  Networking, Electronic and Automation Control Conference}}. IEEE,
  \bibinfo{pages}{581--586}.
\newblock


\bibitem[\protect\citeauthoryear{Ma, Yang, Liu, Liu, Ma, Ren, and Yao}{Ma
  et~al\mbox{.}}{2019}]%
        {ma2019emir}
\bibfield{author}{\bibinfo{person}{Zhuo Ma}, \bibinfo{person}{Yilong Yang},
  \bibinfo{person}{Ximeng Liu}, \bibinfo{person}{Yang Liu},
  \bibinfo{person}{Siqi Ma}, \bibinfo{person}{Kui Ren}, {and}
  \bibinfo{person}{Chang Yao}.} \bibinfo{year}{2019}\natexlab{}.
\newblock \showarticletitle{Emir-auth: Eye-movement and iris based portable
  remote authentication for smart grid}.
\newblock \bibinfo{journal}{\emph{IEEE Transactions on Industrial Informatics}}
  (\bibinfo{year}{2019}).
\newblock


\bibitem[\protect\citeauthoryear{Makihara, Mannami, Tsuji, Hossain, Sugiura,
  Mori, and Yagi}{Makihara et~al\mbox{.}}{2012}]%
        {Makihara_CVATN2012}
\bibfield{author}{\bibinfo{person}{Y. Makihara}, \bibinfo{person}{H. Mannami},
  \bibinfo{person}{A. Tsuji}, \bibinfo{person}{M.A. Hossain},
  \bibinfo{person}{K. Sugiura}, \bibinfo{person}{A. Mori}, {and}
  \bibinfo{person}{Y. Yagi}.} \bibinfo{year}{2012}\natexlab{}.
\newblock \showarticletitle{The OU-ISIR Gait Database Comprising the Treadmill
  Dataset}.
\newblock \bibinfo{journal}{\emph{IPSJ Trans. on Computer Vision and
  Applications}}  \bibinfo{volume}{4} (\bibinfo{date}{Apr.}
  \bibinfo{year}{2012}), \bibinfo{pages}{53--62}.
\newblock


\bibitem[\protect\citeauthoryear{Makihara, Mannami, and Yagi}{Makihara
  et~al\mbox{.}}{2010}]%
        {makihara2010gait}
\bibfield{author}{\bibinfo{person}{Yasushi Makihara},
  \bibinfo{person}{Hidetoshi Mannami}, {and} \bibinfo{person}{Yasushi Yagi}.}
  \bibinfo{year}{2010}\natexlab{}.
\newblock \showarticletitle{Gait analysis of gender and age using a large-scale
  multi-view gait database}. In \bibinfo{booktitle}{\emph{Asian Conference on
  Computer Vision}}. Springer, \bibinfo{pages}{440--451}.
\newblock


\bibitem[\protect\citeauthoryear{Mansur, Makihara, Aqmar, and Yagi}{Mansur
  et~al\mbox{.}}{2014}]%
        {mansur2014gait}
\bibfield{author}{\bibinfo{person}{Al Mansur}, \bibinfo{person}{Yasushi
  Makihara}, \bibinfo{person}{Rasyid Aqmar}, {and} \bibinfo{person}{Yasushi
  Yagi}.} \bibinfo{year}{2014}\natexlab{}.
\newblock \showarticletitle{Gait recognition under speed transition}. In
  \bibinfo{booktitle}{\emph{Proceedings of the IEEE Conference on Computer
  Vision and Pattern Recognition}}. \bibinfo{pages}{2521--2528}.
\newblock


\bibitem[\protect\citeauthoryear{Marsico and Mecca}{Marsico and Mecca}{2019}]%
        {marsico2019survey}
\bibfield{author}{\bibinfo{person}{Maria~De Marsico} {and}
  \bibinfo{person}{Alessio Mecca}.} \bibinfo{year}{2019}\natexlab{}.
\newblock \showarticletitle{A survey on gait recognition via wearable sensors}.
\newblock \bibinfo{journal}{\emph{ACM Computing Surveys (CSUR)}}
  \bibinfo{volume}{52}, \bibinfo{number}{4} (\bibinfo{year}{2019}),
  \bibinfo{pages}{1--39}.
\newblock


\bibitem[\protect\citeauthoryear{Matovski, Nixon, Mahmoodi, and
  Carter}{Matovski et~al\mbox{.}}{2010}]%
        {matovski2010effect}
\bibfield{author}{\bibinfo{person}{Darko~S Matovski}, \bibinfo{person}{Mark~S
  Nixon}, \bibinfo{person}{Sasan Mahmoodi}, {and} \bibinfo{person}{John~N
  Carter}.} \bibinfo{year}{2010}\natexlab{}.
\newblock \showarticletitle{The effect of time on the performance of gait
  biometrics}. In \bibinfo{booktitle}{\emph{2010 Fourth IEEE International
  Conference on Biometrics: Theory, Applications and Systems (BTAS)}}. IEEE,
  \bibinfo{pages}{1--6}.
\newblock


\bibitem[\protect\citeauthoryear{Mehraj and Mir}{Mehraj and Mir}{2020}]%
        {mehraj2020feature}
\bibfield{author}{\bibinfo{person}{Haider Mehraj} {and}
  \bibinfo{person}{Ajaz~Hussain Mir}.} \bibinfo{year}{2020}\natexlab{}.
\newblock \showarticletitle{Feature vector extraction and optimisation for
  multimodal biometrics employing face, ear and gait utilising artificial
  neural networks}.
\newblock \bibinfo{journal}{\emph{International Journal of Cloud Computing}}
  \bibinfo{volume}{9}, \bibinfo{number}{2-3} (\bibinfo{year}{2020}),
  \bibinfo{pages}{131--149}.
\newblock


\bibitem[\protect\citeauthoryear{Mitchell}{Mitchell}{1920}]%
        {ainsworthamitchell1920detection}
\bibfield{author}{\bibinfo{person}{C.~Ainsworth Mitchell}.}
  \bibinfo{year}{1920}\natexlab{}.
\newblock \showarticletitle{The detection of finger-prints on documents}.
\newblock \bibinfo{journal}{\emph{Analyst}} \bibinfo{volume}{45},
  \bibinfo{number}{529} (\bibinfo{year}{1920}), \bibinfo{pages}{122--129}.
\newblock


\bibitem[\protect\citeauthoryear{Mori, Makihara, and Yagi}{Mori
  et~al\mbox{.}}{2010}]%
        {mori2010gait}
\bibfield{author}{\bibinfo{person}{Atsushi Mori}, \bibinfo{person}{Yasushi
  Makihara}, {and} \bibinfo{person}{Yasushi Yagi}.}
  \bibinfo{year}{2010}\natexlab{}.
\newblock \showarticletitle{Gait recognition using period-based phase
  synchronization for low frame-rate videos}. In \bibinfo{booktitle}{\emph{2010
  20th International Conference on Pattern Recognition}}. IEEE,
  \bibinfo{pages}{2194--2197}.
\newblock


\bibitem[\protect\citeauthoryear{Nambiar, Bernardino, and Nascimento}{Nambiar
  et~al\mbox{.}}{2019}]%
        {nambiar2019gait}
\bibfield{author}{\bibinfo{person}{Athira Nambiar}, \bibinfo{person}{Alexandre
  Bernardino}, {and} \bibinfo{person}{Jacinto~C Nascimento}.}
  \bibinfo{year}{2019}\natexlab{}.
\newblock \showarticletitle{Gait-based person re-identification: A survey}.
\newblock \bibinfo{journal}{\emph{ACM Computing Surveys (CSUR)}}
  \bibinfo{volume}{52}, \bibinfo{number}{2} (\bibinfo{year}{2019}),
  \bibinfo{pages}{1--34}.
\newblock


\bibitem[\protect\citeauthoryear{Ngo, Makihara, Nagahara, Mukaigawa, and
  Yagi}{Ngo et~al\mbox{.}}{2014}]%
        {ngo2014largest}
\bibfield{author}{\bibinfo{person}{Thanh~Trung Ngo}, \bibinfo{person}{Yasushi
  Makihara}, \bibinfo{person}{Hajime Nagahara}, \bibinfo{person}{Yasuhiro
  Mukaigawa}, {and} \bibinfo{person}{Yasushi Yagi}.}
  \bibinfo{year}{2014}\natexlab{}.
\newblock \showarticletitle{The largest inertial sensor-based gait database and
  performance evaluation of gait-based personal authentication}.
\newblock \bibinfo{journal}{\emph{Pattern Recognition}} \bibinfo{volume}{47},
  \bibinfo{number}{1} (\bibinfo{year}{2014}), \bibinfo{pages}{228--237}.
\newblock


\bibitem[\protect\citeauthoryear{Ngo, Makihara, Nagahara, Mukaigawa, and
  Yagi}{Ngo et~al\mbox{.}}{2015}]%
        {ngo2015similar}
\bibfield{author}{\bibinfo{person}{Trung~Thanh Ngo}, \bibinfo{person}{Yasushi
  Makihara}, \bibinfo{person}{Hajime Nagahara}, \bibinfo{person}{Yasuhiro
  Mukaigawa}, {and} \bibinfo{person}{Yasushi Yagi}.}
  \bibinfo{year}{2015}\natexlab{}.
\newblock \showarticletitle{Similar gait action recognition using an inertial
  sensor}.
\newblock \bibinfo{journal}{\emph{Pattern Recognition}} \bibinfo{volume}{48},
  \bibinfo{number}{4} (\bibinfo{year}{2015}), \bibinfo{pages}{1289--1301}.
\newblock


\bibitem[\protect\citeauthoryear{Nixon, Carter, Shutler, and Grant}{Nixon
  et~al\mbox{.}}{2001}]%
        {nixon2001experimental}
\bibfield{author}{\bibinfo{person}{M Nixon}, \bibinfo{person}{J Carter},
  \bibinfo{person}{J Shutler}, {and} \bibinfo{person}{M Grant}.}
  \bibinfo{year}{2001}\natexlab{}.
\newblock \showarticletitle{Experimental plan for automatic gait recognition}.
\newblock \bibinfo{journal}{\emph{University of Southampton, Southampton, UK}}
  (\bibinfo{year}{2001}).
\newblock


\bibitem[\protect\citeauthoryear{Nixon and Carter}{Nixon and Carter}{2004}]%
        {nixon2004advances}
\bibfield{author}{\bibinfo{person}{Mark~S Nixon} {and} \bibinfo{person}{John~N
  Carter}.} \bibinfo{year}{2004}\natexlab{}.
\newblock \showarticletitle{Advances in automatic gait recognition}. In
  \bibinfo{booktitle}{\emph{null}}. IEEE, \bibinfo{pages}{139}.
\newblock


\bibitem[\protect\citeauthoryear{Oppenheim, Buck, and Schafer}{Oppenheim
  et~al\mbox{.}}{2001}]%
        {oppenheim}
\bibfield{author}{\bibinfo{person}{Alan~V Oppenheim}, \bibinfo{person}{John~R
  Buck}, {and} \bibinfo{person}{Ronald~W Schafer}.}
  \bibinfo{year}{2001}\natexlab{}.
\newblock \bibinfo{booktitle}{\emph{Discrete-time signal processing. Vol. 2}}.
\newblock \bibinfo{publisher}{Upper Saddle River, NJ: Prentice Hall}.
\newblock


\bibitem[\protect\citeauthoryear{Passos, Costa, and Papa}{Passos
  et~al\mbox{.}}{2017}]%
        {passosCAIP:2017}
\bibfield{author}{\bibinfo{person}{Leandro~A. Passos},
  \bibinfo{person}{Kelton~A.P. Costa}, {and} \bibinfo{person}{Jo{\~a}o~P.
  Papa}.} \bibinfo{year}{2017}\natexlab{}.
\newblock \showarticletitle{Deep Boltzmann Machines Using Adaptive
  Temperatures}. In \bibinfo{booktitle}{\emph{International Conference on
  Computer Analysis of Images and Patterns}}. Springer,
  \bibinfo{pages}{172--183}.
\newblock


\bibitem[\protect\citeauthoryear{Passos, Santana, Moreira, and Papa}{Passos
  et~al\mbox{.}}{2019}]%
        {passosIJCNN:19}
\bibfield{author}{\bibinfo{person}{L.~A. Passos}, \bibinfo{person}{M.~C.
  Santana}, \bibinfo{person}{T. Moreira}, {and} \bibinfo{person}{J.~P. Papa}.}
  \bibinfo{year}{2019}\natexlab{}.
\newblock \showarticletitle{$\kappa$-Entropy Based Restricted Boltzmann
  Machines}. In \bibinfo{booktitle}{\emph{The 2019 International Joint
  Conference on Neural Networks (IJCNN)}}. IEEE, \bibinfo{pages}{1--8}.
\newblock


\bibitem[\protect\citeauthoryear{Pisani, Mhenni, Giot, Cherrier, Poh,
  Ferreira~de Carvalho, Rosenberger, and Amara}{Pisani et~al\mbox{.}}{2019}]%
        {pisani2019adaptive}
\bibfield{author}{\bibinfo{person}{Paulo~Henrique Pisani},
  \bibinfo{person}{Abir Mhenni}, \bibinfo{person}{Romain Giot},
  \bibinfo{person}{Estelle Cherrier}, \bibinfo{person}{Norman Poh},
  \bibinfo{person}{Andr{\'e} Carlos Ponce de~Leon Ferreira~de Carvalho},
  \bibinfo{person}{Christophe Rosenberger}, {and} \bibinfo{person}{Najoua
  Essoukri~Ben Amara}.} \bibinfo{year}{2019}\natexlab{}.
\newblock \showarticletitle{Adaptive biometric systems: Review and
  perspectives}.
\newblock \bibinfo{journal}{\emph{ACM Computing Surveys (CSUR)}}
  \bibinfo{volume}{52}, \bibinfo{number}{5} (\bibinfo{year}{2019}),
  \bibinfo{pages}{1--38}.
\newblock


\bibitem[\protect\citeauthoryear{Potluri, Ravuri, Diedrich, and Schega}{Potluri
  et~al\mbox{.}}{2019}]%
        {potluri2019deep}
\bibfield{author}{\bibinfo{person}{Sasanka Potluri}, \bibinfo{person}{Srinivas
  Ravuri}, \bibinfo{person}{Christian Diedrich}, {and} \bibinfo{person}{Lutz
  Schega}.} \bibinfo{year}{2019}\natexlab{}.
\newblock \showarticletitle{Deep Learning based Gait Abnormality Detection
  using Wearable Sensor System}. In \bibinfo{booktitle}{\emph{2019 41st Annual
  International Conference of the IEEE Engineering in Medicine and Biology
  Society (EMBC)}}. IEEE, \bibinfo{pages}{3613--3619}.
\newblock


\bibitem[\protect\citeauthoryear{Ragan, Johnson, Milton, and Gill}{Ragan
  et~al\mbox{.}}{2016}]%
        {ragan2016ear}
\bibfield{author}{\bibinfo{person}{Elizabeth~J Ragan},
  \bibinfo{person}{Courtney Johnson}, \bibinfo{person}{Jacqueline~N Milton},
  {and} \bibinfo{person}{Christopher~J Gill}.} \bibinfo{year}{2016}\natexlab{}.
\newblock \showarticletitle{Ear biometrics for patient identification in global
  health: a cross-sectional study to test the feasibility of a simplified
  algorithm}.
\newblock \bibinfo{journal}{\emph{BMC research notes}} \bibinfo{volume}{9},
  \bibinfo{number}{1} (\bibinfo{year}{2016}), \bibinfo{pages}{484}.
\newblock


\bibitem[\protect\citeauthoryear{Ramli, Hooi, and Chee}{Ramli
  et~al\mbox{.}}{2016}]%
        {ramli2016development}
\bibfield{author}{\bibinfo{person}{DA Ramli}, \bibinfo{person}{MY Hooi}, {and}
  \bibinfo{person}{KJ Chee}.} \bibinfo{year}{2016}\natexlab{}.
\newblock \showarticletitle{Development of Heartbeat Detection Kit for
  Biometric Authentication System}.
\newblock \bibinfo{journal}{\emph{Procedia Computer Science}}
  \bibinfo{volume}{96} (\bibinfo{year}{2016}), \bibinfo{pages}{305--314}.
\newblock


\bibitem[\protect\citeauthoryear{Rida, Almaadeed, and Almaadeed}{Rida
  et~al\mbox{.}}{2018}]%
        {rida2018robust}
\bibfield{author}{\bibinfo{person}{Imad Rida}, \bibinfo{person}{Noor
  Almaadeed}, {and} \bibinfo{person}{Somaya Almaadeed}.}
  \bibinfo{year}{2018}\natexlab{}.
\newblock \showarticletitle{Robust gait recognition: a comprehensive survey}.
\newblock \bibinfo{journal}{\emph{IET Biometrics}} \bibinfo{volume}{8},
  \bibinfo{number}{1} (\bibinfo{year}{2018}), \bibinfo{pages}{14--28}.
\newblock


\bibitem[\protect\citeauthoryear{Rida, Jiang, , and Marcialis}{Rida
  et~al\mbox{.}}{2016}]%
        {Rida}
\bibfield{author}{\bibinfo{person}{I. Rida}, \bibinfo{person}{X. Jiang},
  \bibinfo{person}{}, {and} \bibinfo{person}{G.~L. Marcialis}.}
  \bibinfo{year}{2016}\natexlab{}.
\newblock \showarticletitle{Human body part selection by group lasso of motion
  for model-free gait recognition}.
\newblock \bibinfo{journal}{\emph{IEEE Signal Processing Letters}}
  \bibinfo{volume}{23}, \bibinfo{number}{1} (\bibinfo{year}{2016}),
  \bibinfo{pages}{154--158}.
\newblock


\bibitem[\protect\citeauthoryear{Roder, Passos, Ribeiro, Benato, Falc\~{a}o,
  and Papa}{Roder et~al\mbox{.}}{ress}]%
        {RoderICAISC:20}
\bibfield{author}{\bibinfo{person}{M. Roder}, \bibinfo{person}{L.~A. Passos},
  \bibinfo{person}{L.~C.~F. Ribeiro}, \bibinfo{person}{B.~C. Benato},
  \bibinfo{person}{A.~L. Falc\~{a}o}, {and} \bibinfo{person}{J.~P. Papa}.}
  \bibinfo{year}{In Press}\natexlab{}.
\newblock \showarticletitle{Intestinal Parasites Classification Using Deep
  Belief Networks}. In \bibinfo{booktitle}{\emph{The 19th International
  Conference on Artificial Intelligence and Soft Computing (ICAISC)}}. IEEE.
\newblock


\bibitem[\protect\citeauthoryear{Sabour, Frosst, and Hinton}{Sabour
  et~al\mbox{.}}{2017}]%
        {sabour2017dynamic}
\bibfield{author}{\bibinfo{person}{Sara Sabour}, \bibinfo{person}{Nicholas
  Frosst}, {and} \bibinfo{person}{Geoffrey~E Hinton}.}
  \bibinfo{year}{2017}\natexlab{}.
\newblock \showarticletitle{Dynamic routing between capsules}. In
  \bibinfo{booktitle}{\emph{Advances in neural information processing
  systems}}. \bibinfo{pages}{3856--3866}.
\newblock


\bibitem[\protect\citeauthoryear{Santana, Passos, Moreira, Colombo,
  de~Albuquerque, and Papa}{Santana et~al\mbox{.}}{2019}]%
        {santanaIEEE-IS:19}
\bibfield{author}{\bibinfo{person}{Marcos~C. Santana},
  \bibinfo{person}{Leandro~A. Passos}, \bibinfo{person}{Thierry~P. Moreira},
  \bibinfo{person}{Danilo Colombo}, \bibinfo{person}{Victor Hugo~C. de
  Albuquerque}, {and} \bibinfo{person}{Jo{\~a}o~P. Papa}.}
  \bibinfo{year}{2019}\natexlab{}.
\newblock \showarticletitle{A Novel Siamese-Based Approach for Scene Change
  Detection With Applications to Obstructed Routes in Hazardous Environments}.
\newblock \bibinfo{journal}{\emph{IEEE Intelligent Systems}}
  \bibinfo{volume}{35}, \bibinfo{number}{1} (\bibinfo{year}{2019}),
  \bibinfo{pages}{44--53}.
\newblock


\bibitem[\protect\citeauthoryear{Sarkar, Phillips, Liu, Vega, Grother, and
  Bowyer}{Sarkar et~al\mbox{.}}{2005}]%
        {sarkar2005humanid}
\bibfield{author}{\bibinfo{person}{Sudeep Sarkar}, \bibinfo{person}{P~Jonathon
  Phillips}, \bibinfo{person}{Zongyi Liu}, \bibinfo{person}{Isidro~Robledo
  Vega}, \bibinfo{person}{Patrick Grother}, {and} \bibinfo{person}{Kevin~W
  Bowyer}.} \bibinfo{year}{2005}\natexlab{}.
\newblock \showarticletitle{The humanid gait challenge problem: Data sets,
  performance, and analysis}.
\newblock \bibinfo{journal}{\emph{IEEE transactions on pattern analysis and
  machine intelligence}} \bibinfo{volume}{27}, \bibinfo{number}{2}
  (\bibinfo{year}{2005}), \bibinfo{pages}{162--177}.
\newblock


\bibitem[\protect\citeauthoryear{Sepas-Moghaddam, Ghorbani, Troje, and
  Etemad}{Sepas-Moghaddam et~al\mbox{.}}{2021}]%
        {sepas2021gait}
\bibfield{author}{\bibinfo{person}{Alireza Sepas-Moghaddam},
  \bibinfo{person}{Saeed Ghorbani}, \bibinfo{person}{Nikolaus~F Troje}, {and}
  \bibinfo{person}{Ali Etemad}.} \bibinfo{year}{2021}\natexlab{}.
\newblock \showarticletitle{Gait Recognition using Multi-Scale Partial
  Representation Transformation with Capsules}. In
  \bibinfo{booktitle}{\emph{2020 25th International Conference on Pattern
  Recognition (ICPR)}}. IEEE, \bibinfo{pages}{8045--8052}.
\newblock


\bibitem[\protect\citeauthoryear{Shiraga, Makihara, Muramatsu, Echigo, and
  Yagi}{Shiraga et~al\mbox{.}}{2016}]%
        {shiraga2016geinet}
\bibfield{author}{\bibinfo{person}{Kohei Shiraga}, \bibinfo{person}{Yasushi
  Makihara}, \bibinfo{person}{Daigo Muramatsu}, \bibinfo{person}{Tomio Echigo},
  {and} \bibinfo{person}{Yasushi Yagi}.} \bibinfo{year}{2016}\natexlab{}.
\newblock \showarticletitle{Geinet: View-invariant gait recognition using a
  convolutional neural network}. In \bibinfo{booktitle}{\emph{2016
  international conference on biometrics (ICB)}}. IEEE, \bibinfo{pages}{1--8}.
\newblock


\bibitem[\protect\citeauthoryear{Shutler, Grant, Nixon, and Carter}{Shutler
  et~al\mbox{.}}{2004}]%
        {shutler2004large}
\bibfield{author}{\bibinfo{person}{Jamie~D Shutler}, \bibinfo{person}{Michael~G
  Grant}, \bibinfo{person}{Mark~S Nixon}, {and} \bibinfo{person}{John~N
  Carter}.} \bibinfo{year}{2004}\natexlab{}.
\newblock \showarticletitle{On a large sequence-based human gait database}.
\newblock In \bibinfo{booktitle}{\emph{Applications and Science in Soft
  Computing}}. \bibinfo{publisher}{Springer}, \bibinfo{pages}{339--346}.
\newblock


\bibitem[\protect\citeauthoryear{Simonyan and Zisserman}{Simonyan and
  Zisserman}{2014}]%
        {simonyan2014very}
\bibfield{author}{\bibinfo{person}{Karen Simonyan} {and}
  \bibinfo{person}{Andrew Zisserman}.} \bibinfo{year}{2014}\natexlab{}.
\newblock \showarticletitle{Very deep convolutional networks for large-scale
  image recognition}.
\newblock \bibinfo{journal}{\emph{arXiv preprint arXiv:1409.1556}}
  (\bibinfo{year}{2014}).
\newblock


\bibitem[\protect\citeauthoryear{Singh, Jain, Arora, and Singh}{Singh
  et~al\mbox{.}}{2018}]%
        {singh2018vision}
\bibfield{author}{\bibinfo{person}{Jasvinder~Pal Singh},
  \bibinfo{person}{Sanjeev Jain}, \bibinfo{person}{Sakshi Arora}, {and}
  \bibinfo{person}{Uday~Pratap Singh}.} \bibinfo{year}{2018}\natexlab{}.
\newblock \showarticletitle{Vision-based gait recognition: A survey}.
\newblock \bibinfo{journal}{\emph{IEEE Access}}  \bibinfo{volume}{6}
  (\bibinfo{year}{2018}), \bibinfo{pages}{70497--70527}.
\newblock


\bibitem[\protect\citeauthoryear{Sokolova and Konushin}{Sokolova and
  Konushin}{2017a}]%
        {sokolova2017gait}
\bibfield{author}{\bibinfo{person}{A Sokolova} {and} \bibinfo{person}{A
  Konushin}.} \bibinfo{year}{2017}\natexlab{a}.
\newblock \showarticletitle{Gait Recognition Based on Convolutional Neural
  Networks}.
\newblock \bibinfo{journal}{\emph{International Archives of the Photogrammetry,
  Remote Sensing \& Spatial Information Sciences}}  \bibinfo{volume}{42}
  (\bibinfo{year}{2017}).
\newblock


\bibitem[\protect\citeauthoryear{Sokolova and Konushin}{Sokolova and
  Konushin}{2017b}]%
        {sokolova2017posed}
\bibfield{author}{\bibinfo{person}{Anna Sokolova} {and} \bibinfo{person}{Anton
  Konushin}.} \bibinfo{year}{2017}\natexlab{b}.
\newblock \showarticletitle{Pose-based Deep Gait Recognition}.
\newblock \bibinfo{journal}{\emph{CoRR}}  \bibinfo{volume}{abs/1710.06512}
  (\bibinfo{year}{2017}).
\newblock
\showeprint[arxiv]{1710.06512}
\urldef\tempurl%
\url{http://arxiv.org/abs/1710.06512}
\showURL{%
\tempurl}


\bibitem[\protect\citeauthoryear{Souza~Jr, Passos, Mendel, Ebigbo, Probst,
  Messmann, Palm, and Papa}{Souza~Jr et~al\mbox{.}}{2020}]%
        {de2020assisting}
\bibfield{author}{\bibinfo{person}{Luis~A. Souza~Jr},
  \bibinfo{person}{Leandro~A. Passos}, \bibinfo{person}{Robert Mendel},
  \bibinfo{person}{Alanna Ebigbo}, \bibinfo{person}{Andreas Probst},
  \bibinfo{person}{Helmut Messmann}, \bibinfo{person}{Christoph Palm}, {and}
  \bibinfo{person}{Jo{\~a}o~P. Papa}.} \bibinfo{year}{2020}\natexlab{}.
\newblock \showarticletitle{Assisting Barrett's esophagus identification using
  endoscopic data augmentation based on Generative Adversarial Networks}.
\newblock \bibinfo{journal}{\emph{Computers in Biology and Medicine}}
  \bibinfo{volume}{126} (\bibinfo{year}{2020}), \bibinfo{pages}{104029}.
\newblock


\bibitem[\protect\citeauthoryear{Sultana, Paul, and Gavrilova}{Sultana
  et~al\mbox{.}}{2017}]%
        {sultana2017social}
\bibfield{author}{\bibinfo{person}{Madeena Sultana},
  \bibinfo{person}{Padma~Polash Paul}, {and} \bibinfo{person}{Marina~L
  Gavrilova}.} \bibinfo{year}{2017}\natexlab{}.
\newblock \showarticletitle{Social behavioral information fusion in multimodal
  biometrics}.
\newblock \bibinfo{journal}{\emph{IEEE Transactions on Systems, Man, and
  Cybernetics: Systems}} \bibinfo{volume}{48}, \bibinfo{number}{12}
  (\bibinfo{year}{2017}), \bibinfo{pages}{2176--2187}.
\newblock


\bibitem[\protect\citeauthoryear{Sun, Lo, and Lo}{Sun et~al\mbox{.}}{2019}]%
        {sun2019deep}
\bibfield{author}{\bibinfo{person}{Yingnan Sun}, \bibinfo{person}{Frank P-W
  Lo}, {and} \bibinfo{person}{Benny Lo}.} \bibinfo{year}{2019}\natexlab{}.
\newblock \showarticletitle{A deep learning approach on gender and age
  recognition using a single inertial sensor}. In
  \bibinfo{booktitle}{\emph{2019 IEEE 16th International Conference on Wearable
  and Implantable Body Sensor Networks (BSN)}}. IEEE, \bibinfo{pages}{1--4}.
\newblock


\bibitem[\protect\citeauthoryear{Sundararajan, Sarwat, and Pons}{Sundararajan
  et~al\mbox{.}}{2019}]%
        {sundararajan2019survey}
\bibfield{author}{\bibinfo{person}{Aditya Sundararajan},
  \bibinfo{person}{Arif~I Sarwat}, {and} \bibinfo{person}{Alexander Pons}.}
  \bibinfo{year}{2019}\natexlab{}.
\newblock \showarticletitle{A survey on modality characteristics, performance
  evaluation metrics, and security for traditional and wearable biometric
  systems}.
\newblock \bibinfo{journal}{\emph{ACM Computing Surveys (CSUR)}}
  \bibinfo{volume}{52}, \bibinfo{number}{2} (\bibinfo{year}{2019}),
  \bibinfo{pages}{1--36}.
\newblock


\bibitem[\protect\citeauthoryear{Szegedy, Vanhoucke, Ioffe, Shlens, and
  Wojna}{Szegedy et~al\mbox{.}}{2016}]%
        {szegedy2016rethinking}
\bibfield{author}{\bibinfo{person}{Christian Szegedy}, \bibinfo{person}{Vincent
  Vanhoucke}, \bibinfo{person}{Sergey Ioffe}, \bibinfo{person}{Jon Shlens},
  {and} \bibinfo{person}{Zbigniew Wojna}.} \bibinfo{year}{2016}\natexlab{}.
\newblock \showarticletitle{Rethinking the inception architecture for computer
  vision}. In \bibinfo{booktitle}{\emph{Proceedings of the IEEE conference on
  computer vision and pattern recognition}}. \bibinfo{pages}{2818--2826}.
\newblock


\bibitem[\protect\citeauthoryear{Taigman, Yang, Ranzato, and Wolf}{Taigman
  et~al\mbox{.}}{2014}]%
        {taigman2014deepface}
\bibfield{author}{\bibinfo{person}{Yaniv Taigman}, \bibinfo{person}{Ming Yang},
  \bibinfo{person}{Marc'Aurelio Ranzato}, {and} \bibinfo{person}{Lior Wolf}.}
  \bibinfo{year}{2014}\natexlab{}.
\newblock \showarticletitle{Deepface: Closing the gap to human-level
  performance in face verification}. In \bibinfo{booktitle}{\emph{Proceedings
  of the IEEE conference on computer vision and pattern recognition}}.
  \bibinfo{pages}{1701--1708}.
\newblock


\bibitem[\protect\citeauthoryear{Takemura, Makihara, Muramatsu, Echigo, and
  Yagi}{Takemura et~al\mbox{.}}{2017}]%
        {takemura2017input}
\bibfield{author}{\bibinfo{person}{Noriko Takemura}, \bibinfo{person}{Yasushi
  Makihara}, \bibinfo{person}{Daigo Muramatsu}, \bibinfo{person}{Tomio Echigo},
  {and} \bibinfo{person}{Yasushi Yagi}.} \bibinfo{year}{2017}\natexlab{}.
\newblock \showarticletitle{On input/output architectures for convolutional
  neural network-based cross-view gait recognition}.
\newblock \bibinfo{journal}{\emph{IEEE Transactions on Circuits and Systems for
  Video Technology}} (\bibinfo{year}{2017}).
\newblock


\bibitem[\protect\citeauthoryear{Takemura, Makihara, Muramatsu, Echigo, and
  Yagi}{Takemura et~al\mbox{.}}{2018}]%
        {takemura2018multi}
\bibfield{author}{\bibinfo{person}{Noriko Takemura}, \bibinfo{person}{Yasushi
  Makihara}, \bibinfo{person}{Daigo Muramatsu}, \bibinfo{person}{Tomio Echigo},
  {and} \bibinfo{person}{Yasushi Yagi}.} \bibinfo{year}{2018}\natexlab{}.
\newblock \showarticletitle{Multi-view large population gait dataset and its
  performance evaluation for cross-view gait recognition}.
\newblock \bibinfo{journal}{\emph{IPSJ Transactions on Computer Vision and
  Applications}} \bibinfo{volume}{10}, \bibinfo{number}{1}
  (\bibinfo{year}{2018}), \bibinfo{pages}{4}.
\newblock


\bibitem[\protect\citeauthoryear{Tan, Huang, Yu, and Tan}{Tan
  et~al\mbox{.}}{2006}]%
        {tan2006efficient}
\bibfield{author}{\bibinfo{person}{Daoliang Tan}, \bibinfo{person}{Kaiqi
  Huang}, \bibinfo{person}{Shiqi Yu}, {and} \bibinfo{person}{Tieniu Tan}.}
  \bibinfo{year}{2006}\natexlab{}.
\newblock \showarticletitle{Efficient night gait recognition based on template
  matching}. In \bibinfo{booktitle}{\emph{18th International Conference on
  Pattern Recognition (ICPR'06)}}, Vol.~\bibinfo{volume}{3}. IEEE,
  \bibinfo{pages}{1000--1003}.
\newblock


\bibitem[\protect\citeauthoryear{Tan and Le}{Tan and Le}{2019}]%
        {efficientnet}
\bibfield{author}{\bibinfo{person}{Mingxing Tan} {and} \bibinfo{person}{Quoc
  Le}.} \bibinfo{year}{2019}\natexlab{}.
\newblock \showarticletitle{Efficientnet: Rethinking model scaling for
  convolutional neural networks}. In \bibinfo{booktitle}{\emph{International
  Conference on Machine Learning}}. PMLR, \bibinfo{pages}{6105--6114}.
\newblock


\bibitem[\protect\citeauthoryear{Tiong, Kim, and Ro}{Tiong
  et~al\mbox{.}}{2020}]%
        {tiong2020multimodal}
\bibfield{author}{\bibinfo{person}{Leslie Ching~Ow Tiong},
  \bibinfo{person}{Seong~Tae Kim}, {and} \bibinfo{person}{Yong~Man Ro}.}
  \bibinfo{year}{2020}\natexlab{}.
\newblock \showarticletitle{Multimodal facial biometrics recognition:
  Dual-stream convolutional neural networks with multi-feature fusion layers}.
\newblock \bibinfo{journal}{\emph{Image and Vision Computing}}
  \bibinfo{volume}{102} (\bibinfo{year}{2020}), \bibinfo{pages}{103977}.
\newblock


\bibitem[\protect\citeauthoryear{Tran, Hoang, Nguyen, Kim, and Choi}{Tran
  et~al\mbox{.}}{2021}]%
        {tran2021multi}
\bibfield{author}{\bibinfo{person}{Lam Tran}, \bibinfo{person}{Thang Hoang},
  \bibinfo{person}{Thuc Nguyen}, \bibinfo{person}{Hyunil Kim}, {and}
  \bibinfo{person}{Deokjai Choi}.} \bibinfo{year}{2021}\natexlab{}.
\newblock \showarticletitle{Multi-Model Long Short-Term Memory Network for Gait
  Recognition Using Window-Based Data Segment}.
\newblock \bibinfo{journal}{\emph{IEEE Access}}  \bibinfo{volume}{9}
  (\bibinfo{year}{2021}), \bibinfo{pages}{23826--23839}.
\newblock


\bibitem[\protect\citeauthoryear{Tsuji, Makihara, and Yagi}{Tsuji
  et~al\mbox{.}}{2010}]%
        {tsuji2010silhouette}
\bibfield{author}{\bibinfo{person}{Akira Tsuji}, \bibinfo{person}{Yasushi
  Makihara}, {and} \bibinfo{person}{Yasushi Yagi}.}
  \bibinfo{year}{2010}\natexlab{}.
\newblock \showarticletitle{Silhouette transformation based on walking speed
  for gait identification}. In \bibinfo{booktitle}{\emph{2010 IEEE Computer
  Society Conference on Computer Vision and Pattern Recognition}}. IEEE,
  \bibinfo{pages}{717--722}.
\newblock


\bibitem[\protect\citeauthoryear{Turk and Pentland}{Turk and Pentland}{1991}]%
        {turk1991eigenfaces}
\bibfield{author}{\bibinfo{person}{Matthew Turk} {and} \bibinfo{person}{Alex
  Pentland}.} \bibinfo{year}{1991}\natexlab{}.
\newblock \showarticletitle{Eigenfaces for recognition}.
\newblock \bibinfo{journal}{\emph{Journal of cognitive neuroscience}}
  \bibinfo{volume}{3}, \bibinfo{number}{1} (\bibinfo{year}{1991}),
  \bibinfo{pages}{71--86}.
\newblock


\bibitem[\protect\citeauthoryear{Uddin, Ngo, Makihara, Takemura, Li, Muramatsu,
  and Yagi}{Uddin et~al\mbox{.}}{2018}]%
        {uddin2018isir}
\bibfield{author}{\bibinfo{person}{Md~Zasim Uddin},
  \bibinfo{person}{Thanh~Trung Ngo}, \bibinfo{person}{Yasushi Makihara},
  \bibinfo{person}{Noriko Takemura}, \bibinfo{person}{Xiang Li},
  \bibinfo{person}{Daigo Muramatsu}, {and} \bibinfo{person}{Yasushi Yagi}.}
  \bibinfo{year}{2018}\natexlab{}.
\newblock \showarticletitle{The OU-ISIR Large Population Gait Database with
  real-life carried object and its performance evaluation}.
\newblock \bibinfo{journal}{\emph{IPSJ Transactions on Computer Vision and
  Applications}} \bibinfo{volume}{10}, \bibinfo{number}{1}
  (\bibinfo{year}{2018}), \bibinfo{pages}{1--11}.
\newblock


\bibitem[\protect\citeauthoryear{Vaswani, Shazeer, Parmar, Uszkoreit, Jones,
  Gomez, Kaiser, and Polosukhin}{Vaswani et~al\mbox{.}}{2017}]%
        {transformers-original}
\bibfield{author}{\bibinfo{person}{Ashish Vaswani}, \bibinfo{person}{Noam
  Shazeer}, \bibinfo{person}{Niki Parmar}, \bibinfo{person}{Jakob Uszkoreit},
  \bibinfo{person}{Llion Jones}, \bibinfo{person}{Aidan~N Gomez},
  \bibinfo{person}{Lukasz Kaiser}, {and} \bibinfo{person}{Illia Polosukhin}.}
  \bibinfo{year}{2017}\natexlab{}.
\newblock \showarticletitle{Attention is all you need}.
\newblock \bibinfo{journal}{\emph{arXiv preprint arXiv:1706.03762}}
  (\bibinfo{year}{2017}).
\newblock


\bibitem[\protect\citeauthoryear{Vincent, Larochelle, Lajoie, Bengio, Manzagol,
  and Bottou}{Vincent et~al\mbox{.}}{2010}]%
        {sda}
\bibfield{author}{\bibinfo{person}{Pascal Vincent}, \bibinfo{person}{Hugo
  Larochelle}, \bibinfo{person}{Isabelle Lajoie}, \bibinfo{person}{Yoshua
  Bengio}, \bibinfo{person}{Pierre-Antoine Manzagol}, {and}
  \bibinfo{person}{L{\'e}on Bottou}.} \bibinfo{year}{2010}\natexlab{}.
\newblock \showarticletitle{Stacked denoising autoencoders: Learning useful
  representations in a deep network with a local denoising criterion.}
\newblock \bibinfo{journal}{\emph{Journal of machine learning research}}
  \bibinfo{volume}{11}, \bibinfo{number}{12} (\bibinfo{year}{2010}).
\newblock


\bibitem[\protect\citeauthoryear{W.~Kusakunniran and Li}{W.~Kusakunniran and
  Li}{2013}]%
        {hongdong}
\bibfield{author}{\bibinfo{person}{Jian Zhang Yi~Ma W.~Kusakunniran, Qiang~Wu}
  {and} \bibinfo{person}{Hongdong Li}.} \bibinfo{year}{2013}\natexlab{}.
\newblock \showarticletitle{A New View-Invariant Feature for Cross-View Gait
  Recognition}.
\newblock \bibinfo{journal}{\emph{IEEE TIFS}} (\bibinfo{year}{2013}),
  \bibinfo{pages}{1642–1653}.
\newblock


\bibitem[\protect\citeauthoryear{Wan, Wang, and Phoha}{Wan
  et~al\mbox{.}}{2018}]%
        {wan2018survey}
\bibfield{author}{\bibinfo{person}{Changsheng Wan}, \bibinfo{person}{Li Wang},
  {and} \bibinfo{person}{Vir~V Phoha}.} \bibinfo{year}{2018}\natexlab{}.
\newblock \showarticletitle{A survey on gait recognition}.
\newblock \bibinfo{journal}{\emph{ACM Computing Surveys (CSUR)}}
  \bibinfo{volume}{51}, \bibinfo{number}{5} (\bibinfo{year}{2018}),
  \bibinfo{pages}{1--35}.
\newblock


\bibitem[\protect\citeauthoryear{Wang, Muhammad, Wang, He, and Sun}{Wang
  et~al\mbox{.}}{2020}]%
        {wang2020towards}
\bibfield{author}{\bibinfo{person}{Caiyong Wang}, \bibinfo{person}{Jawad
  Muhammad}, \bibinfo{person}{Yunlong Wang}, \bibinfo{person}{Zhaofeng He},
  {and} \bibinfo{person}{Zhenan Sun}.} \bibinfo{year}{2020}\natexlab{}.
\newblock \showarticletitle{Towards complete and accurate iris segmentation
  using deep multi-task attention network for non-cooperative iris
  recognition}.
\newblock \bibinfo{journal}{\emph{IEEE Transactions on information forensics
  and security}}  \bibinfo{volume}{15} (\bibinfo{year}{2020}),
  \bibinfo{pages}{2944--2959}.
\newblock


\bibitem[\protect\citeauthoryear{Wang and Kumar}{Wang and Kumar}{2019}]%
        {wang2019cross}
\bibfield{author}{\bibinfo{person}{Kuo Wang} {and} \bibinfo{person}{Ajay
  Kumar}.} \bibinfo{year}{2019}\natexlab{}.
\newblock \showarticletitle{Cross-spectral iris recognition using CNN and
  supervised discrete hashing}.
\newblock \bibinfo{journal}{\emph{Pattern Recognition}}  \bibinfo{volume}{86}
  (\bibinfo{year}{2019}), \bibinfo{pages}{85--98}.
\newblock


\bibitem[\protect\citeauthoryear{Wang, Liu, Lee, Ding, and Lin}{Wang
  et~al\mbox{.}}{2019}]%
        {wang2019nonstandard}
\bibfield{author}{\bibinfo{person}{Kejun Wang}, \bibinfo{person}{Liangliang
  Liu}, \bibinfo{person}{Yilong Lee}, \bibinfo{person}{Xinnan Ding}, {and}
  \bibinfo{person}{Junyu Lin}.} \bibinfo{year}{2019}\natexlab{}.
\newblock \showarticletitle{Nonstandard Periodic Gait Energy Image for Gait
  Recognition and Data Augmentation}. In \bibinfo{booktitle}{\emph{Chinese
  Conference on Pattern Recognition and Computer Vision (PRCV)}}. Springer,
  \bibinfo{pages}{197--208}.
\newblock


\bibitem[\protect\citeauthoryear{Wang, Tan, Ning, and Hu}{Wang
  et~al\mbox{.}}{2003}]%
        {wang2003silhouette}
\bibfield{author}{\bibinfo{person}{Liang Wang}, \bibinfo{person}{Tieniu Tan},
  \bibinfo{person}{Huazhong Ning}, {and} \bibinfo{person}{Weiming Hu}.}
  \bibinfo{year}{2003}\natexlab{}.
\newblock \showarticletitle{Silhouette analysis-based gait recognition for
  human identification}.
\newblock \bibinfo{journal}{\emph{IEEE transactions on pattern analysis and
  machine intelligence}} \bibinfo{volume}{25}, \bibinfo{number}{12}
  (\bibinfo{year}{2003}), \bibinfo{pages}{1505--1518}.
\newblock


\bibitem[\protect\citeauthoryear{Wang and Yan}{Wang and Yan}{2020}]%
        {wang2020human}
\bibfield{author}{\bibinfo{person}{Xiuhui Wang} {and} \bibinfo{person}{Wei~Qi
  Yan}.} \bibinfo{year}{2020}\natexlab{}.
\newblock \showarticletitle{Human gait recognition based on frame-by-frame gait
  energy images and convolutional long short-term memory}.
\newblock \bibinfo{journal}{\emph{International journal of neural systems}}
  \bibinfo{volume}{30}, \bibinfo{number}{01} (\bibinfo{year}{2020}),
  \bibinfo{pages}{1950027}.
\newblock


\bibitem[\protect\citeauthoryear{Wu, Huang, Wang, Wang, and Tan}{Wu
  et~al\mbox{.}}{2017a}]%
        {wu2017comprehensive}
\bibfield{author}{\bibinfo{person}{Zifeng Wu}, \bibinfo{person}{Yongzhen
  Huang}, \bibinfo{person}{Liang Wang}, \bibinfo{person}{Xiaogang Wang}, {and}
  \bibinfo{person}{Tieniu Tan}.} \bibinfo{year}{2017}\natexlab{a}.
\newblock \showarticletitle{A comprehensive study on cross-view gait based
  human identification with deep cnns}.
\newblock \bibinfo{journal}{\emph{IEEE transactions on pattern analysis and
  machine intelligence}} \bibinfo{volume}{39}, \bibinfo{number}{2}
  (\bibinfo{year}{2017}), \bibinfo{pages}{209--226}.
\newblock


\bibitem[\protect\citeauthoryear{Wu, Huang, Wang, Wang, and Tan}{Wu
  et~al\mbox{.}}{2017b}]%
        {Wu}
\bibfield{author}{\bibinfo{person}{Z. Wu}, \bibinfo{person}{Y. Huang},
  \bibinfo{person}{L. Wang}, \bibinfo{person}{X. Wang}, {and}
  \bibinfo{person}{T. Tan}.} \bibinfo{year}{2017}\natexlab{b}.
\newblock \showarticletitle{A comprehensive study on cross-view gait based
  human identification with deep cnns}.
\newblock \bibinfo{journal}{\emph{IEEE transactions on pattern analysis and
  machine intelligence}} \bibinfo{volume}{39}, \bibinfo{number}{2}
  (\bibinfo{year}{2017}), \bibinfo{pages}{209--226}.
\newblock


\bibitem[\protect\citeauthoryear{X.~Li and Ren}{X.~Li and Ren}{2020}]%
        {lilim}
\bibfield{author}{\bibinfo{person}{C.~Xu Y.~Yagi X.~Li, Y.~Makihara} {and}
  \bibinfo{person}{M. Ren}.} \bibinfo{year}{2020}\natexlab{}.
\newblock \showarticletitle{Gait recognitionvia semi-supervised disentangled
  representation learning to identityand covariate features}.
\newblock \bibinfo{journal}{\emph{in Computer Vision and Pattern Recognition}}
  (\bibinfo{year}{2020}).
\newblock


\bibitem[\protect\citeauthoryear{Xiong, Zhang, Zhao, and Zhao}{Xiong
  et~al\mbox{.}}{2020}]%
        {xiong2020continuous}
\bibfield{author}{\bibinfo{person}{Dezhen Xiong}, \bibinfo{person}{Daohui
  Zhang}, \bibinfo{person}{Xingang Zhao}, {and} \bibinfo{person}{Yiwen Zhao}.}
  \bibinfo{year}{2020}\natexlab{}.
\newblock \showarticletitle{Continuous human gait tracking using sEMG signals}.
  In \bibinfo{booktitle}{\emph{2020 42nd Annual International Conference of the
  IEEE Engineering in Medicine \& Biology Society (EMBC)}}. IEEE,
  \bibinfo{pages}{3094--3097}.
\newblock


\bibitem[\protect\citeauthoryear{Xu, Makihara, Li, Yagi, and Lu}{Xu
  et~al\mbox{.}}{2020}]%
        {xu2020cross}
\bibfield{author}{\bibinfo{person}{Chi Xu}, \bibinfo{person}{Yasushi Makihara},
  \bibinfo{person}{Xiang Li}, \bibinfo{person}{Yasushi Yagi}, {and}
  \bibinfo{person}{Jianfeng Lu}.} \bibinfo{year}{2020}\natexlab{}.
\newblock \showarticletitle{Cross-view gait recognition using pairwise spatial
  transformer networks}.
\newblock \bibinfo{journal}{\emph{IEEE Transactions on Circuits and Systems for
  Video Technology}} (\bibinfo{year}{2020}).
\newblock


\bibitem[\protect\citeauthoryear{Xu, Makihara, Ogi, Li, Yagi, and Lu}{Xu
  et~al\mbox{.}}{2017}]%
        {Xu_CVA2017}
\bibfield{author}{\bibinfo{person}{Chi Xu}, \bibinfo{person}{Yasushi Makihara},
  \bibinfo{person}{Gakuto Ogi}, \bibinfo{person}{Xiang Li},
  \bibinfo{person}{Yasushi Yagi}, {and} \bibinfo{person}{Jianfeng Lu}.}
  \bibinfo{year}{2017}\natexlab{}.
\newblock \showarticletitle{The OU-ISIR Gait Database Comprising the Large
  Population Dataset with Age and Performance Evaluation of Age Estimation}.
\newblock \bibinfo{journal}{\emph{IPSJ Trans. on Computer Vision and
  Applications}} \bibinfo{volume}{9}, \bibinfo{number}{24}
  (\bibinfo{year}{2017}), \bibinfo{pages}{1--14}.
\newblock


\bibitem[\protect\citeauthoryear{Xu, Lu, Zhang, Yeung, and Chen}{Xu
  et~al\mbox{.}}{2019}]%
        {xu2019gait}
\bibfield{author}{\bibinfo{person}{Zhaopeng Xu}, \bibinfo{person}{Wei Lu},
  \bibinfo{person}{Qin Zhang}, \bibinfo{person}{Yuileong Yeung}, {and}
  \bibinfo{person}{Xin Chen}.} \bibinfo{year}{2019}\natexlab{}.
\newblock \showarticletitle{Gait Recognition Based on Capsule Network}.
\newblock \bibinfo{journal}{\emph{Journal of Visual Communication and Image
  Representation}} (\bibinfo{year}{2019}).
\newblock


\bibitem[\protect\citeauthoryear{Yasushi~Makihara and Yagi}{Yasushi~Makihara
  and Yagi}{2006}]%
        {makihara}
\bibfield{author}{\bibinfo{person}{Yasuhiro Mukaigawa Tomio~Echigo
  Yasushi~Makihara, Ryusuke~Sagawa} {and} \bibinfo{person}{Yasushi Yagi}.}
  \bibinfo{year}{2006}\natexlab{}.
\newblock \showarticletitle{Gait Recognition Using a View Transformation Model
  in the Frequency Domain}.
\newblock \bibinfo{journal}{\emph{In ECCV: v.3953}} (\bibinfo{year}{2006}),
  \bibinfo{pages}{151–163}.
\newblock


\bibitem[\protect\citeauthoryear{Yeo and Park}{Yeo and Park}{2020}]%
        {yeo2020accuracy}
\bibfield{author}{\bibinfo{person}{Sang~Seok Yeo} {and}
  \bibinfo{person}{Ga~Young Park}.} \bibinfo{year}{2020}\natexlab{}.
\newblock \showarticletitle{Accuracy verification of spatio-temporal and
  kinematic parameters for gait using inertial measurement unit system}.
\newblock \bibinfo{journal}{\emph{Sensors}} \bibinfo{volume}{20},
  \bibinfo{number}{5} (\bibinfo{year}{2020}), \bibinfo{pages}{1343}.
\newblock


\bibitem[\protect\citeauthoryear{Yogarajah, Chaurasia, Condell, and
  Prasad}{Yogarajah et~al\mbox{.}}{2015}]%
        {Yogarajah}
\bibfield{author}{\bibinfo{person}{P. Yogarajah}, \bibinfo{person}{P.
  Chaurasia}, \bibinfo{person}{J. Condell}, {and} \bibinfo{person}{G. Prasad}.}
  \bibinfo{year}{2015}\natexlab{}.
\newblock \showarticletitle{Enhancing gait based person identification using
  joint sparsity model and -norm minimization}.
\newblock \bibinfo{journal}{\emph{Pattern Recognition}}  \bibinfo{volume}{38}
  (\bibinfo{year}{2015}), \bibinfo{pages}{3--22}.
\newblock


\bibitem[\protect\citeauthoryear{Yu, Davis, and Fritz}{Yu
  et~al\mbox{.}}{2019}]%
        {yu2019attributing}
\bibfield{author}{\bibinfo{person}{Ning Yu}, \bibinfo{person}{Larry~S Davis},
  {and} \bibinfo{person}{Mario Fritz}.} \bibinfo{year}{2019}\natexlab{}.
\newblock \showarticletitle{Attributing fake images to gans: Learning and
  analyzing gan fingerprints}. In \bibinfo{booktitle}{\emph{Proceedings of the
  IEEE International Conference on Computer Vision}}.
  \bibinfo{pages}{7556--7566}.
\newblock


\bibitem[\protect\citeauthoryear{Yu, Chen, Wang, Shen, and Huang}{Yu
  et~al\mbox{.}}{2017}]%
        {yu2017invariant}
\bibfield{author}{\bibinfo{person}{Shiqi Yu}, \bibinfo{person}{Haifeng Chen},
  \bibinfo{person}{Qing Wang}, \bibinfo{person}{Linlin Shen}, {and}
  \bibinfo{person}{Yongzhen Huang}.} \bibinfo{year}{2017}\natexlab{}.
\newblock \showarticletitle{Invariant feature extraction for gait recognition
  using only one uniform model}.
\newblock \bibinfo{journal}{\emph{Neurocomputing}}  \bibinfo{volume}{239}
  (\bibinfo{year}{2017}), \bibinfo{pages}{81--93}.
\newblock


\bibitem[\protect\citeauthoryear{Yu, Tan, and Tan}{Yu et~al\mbox{.}}{2006}]%
        {casiab}
\bibfield{author}{\bibinfo{person}{Shiqi Yu}, \bibinfo{person}{Daoliang Tan},
  {and} \bibinfo{person}{Tieniu Tan}.} \bibinfo{year}{2006}\natexlab{}.
\newblock \showarticletitle{A Framework for Evaluating the Effect of View
  Angle, Clothing and Carrying Condition on Gait Recognition}. In
  \bibinfo{booktitle}{\emph{Pattern Recognition}}, Vol.~\bibinfo{volume}{4}.
  \bibinfo{pages}{441--444}.
\newblock
\showISSN{1051-4651}
\urldef\tempurl%
\url{https://doi.org/10.1109/ICPR.2006.67}
\showDOI{\tempurl}


\bibitem[\protect\citeauthoryear{Yu, Wang, and Huang}{Yu et~al\mbox{.}}{2013}]%
        {yu2013large}
\bibfield{author}{\bibinfo{person}{Shiqi Yu}, \bibinfo{person}{Qing Wang},
  {and} \bibinfo{person}{Yongzhen Huang}.} \bibinfo{year}{2013}\natexlab{}.
\newblock \showarticletitle{A large RGB-D gait dataset and the baseline
  algorithm}. In \bibinfo{booktitle}{\emph{Chinese Conference on Biometric
  Recognition}}. Springer, \bibinfo{pages}{417--424}.
\newblock


\bibitem[\protect\citeauthoryear{Z.~Zhang and Liu}{Z.~Zhang and Liu}{2020}]%
        {zanglim}
\bibfield{author}{\bibinfo{person}{F.~Liu Z.~Zhang, L.~Tran} {and}
  \bibinfo{person}{X. Liu}.} \bibinfo{year}{2020}\natexlab{}.
\newblock \showarticletitle{On learning disentangled representations for gait
  recognition}.
\newblock \bibinfo{journal}{\emph{IEEE Transactions on Pattern Analysis and
  Machine Intelligence, vol. in press}} (\bibinfo{year}{2020}).
\newblock


\bibitem[\protect\citeauthoryear{Zagoruyko and Komodakis}{Zagoruyko and
  Komodakis}{2016}]%
        {zagoruyko2016wide}
\bibfield{author}{\bibinfo{person}{Sergey Zagoruyko} {and}
  \bibinfo{person}{Nikos Komodakis}.} \bibinfo{year}{2016}\natexlab{}.
\newblock \showarticletitle{Wide residual networks}.
\newblock \bibinfo{journal}{\emph{arXiv preprint arXiv:1605.07146}}
  (\bibinfo{year}{2016}).
\newblock


\bibitem[\protect\citeauthoryear{Zehngut, Juefei-Xu, Bardia, Pal, Bhagavatula,
  and Savvides}{Zehngut et~al\mbox{.}}{2015}]%
        {zehngut2015investigating}
\bibfield{author}{\bibinfo{person}{Niv Zehngut}, \bibinfo{person}{Felix
  Juefei-Xu}, \bibinfo{person}{Rishabh Bardia}, \bibinfo{person}{Dipan~K Pal},
  \bibinfo{person}{Chandrasekhar Bhagavatula}, {and} \bibinfo{person}{Marios
  Savvides}.} \bibinfo{year}{2015}\natexlab{}.
\newblock \showarticletitle{Investigating the feasibility of image-based nose
  biometrics}. In \bibinfo{booktitle}{\emph{Image Processing (ICIP), 2015 IEEE
  International Conference on}}. IEEE, \bibinfo{pages}{522--526}.
\newblock


\bibitem[\protect\citeauthoryear{Zhang, Tran, Yin, Atoum, Liu, Wan, and
  Wang}{Zhang et~al\mbox{.}}{2019}]%
        {zhang2019gait}
\bibfield{author}{\bibinfo{person}{Ziyuan Zhang}, \bibinfo{person}{Luan Tran},
  \bibinfo{person}{Xi Yin}, \bibinfo{person}{Yousef Atoum},
  \bibinfo{person}{Xiaoming Liu}, \bibinfo{person}{Jian Wan}, {and}
  \bibinfo{person}{Nanxin Wang}.} \bibinfo{year}{2019}\natexlab{}.
\newblock \showarticletitle{Gait Recognition via Disentangled Representation
  Learning}.
\newblock \bibinfo{journal}{\emph{arXiv preprint arXiv:1904.04925}}
  (\bibinfo{year}{2019}).
\newblock


\bibitem[\protect\citeauthoryear{Zhao, Dong, Li, Qi, and Zhou}{Zhao
  et~al\mbox{.}}{2021}]%
        {zhao2021associated}
\bibfield{author}{\bibinfo{person}{Aite Zhao}, \bibinfo{person}{Junyu Dong},
  \bibinfo{person}{Jianbo Li}, \bibinfo{person}{Lin Qi}, {and}
  \bibinfo{person}{Huiyu Zhou}.} \bibinfo{year}{2021}\natexlab{}.
\newblock \showarticletitle{Associated Spatio-Temporal Capsule Network for Gait
  Recognition}.
\newblock \bibinfo{journal}{\emph{arXiv preprint arXiv:2101.02458}}
  (\bibinfo{year}{2021}).
\newblock


\bibitem[\protect\citeauthoryear{Zhao, Jiang, Stathaki, and Zhang}{Zhao
  et~al\mbox{.}}{2016}]%
        {Zhao}
\bibfield{author}{\bibinfo{person}{X. Zhao}, \bibinfo{person}{Y. Jiang},
  \bibinfo{person}{T. Stathaki}, {and} \bibinfo{person}{H. Zhang}.}
  \bibinfo{year}{2016}\natexlab{}.
\newblock \showarticletitle{Gait recognition method for arbitrary straight
  walking paths using appearance conversion machine}.
\newblock \bibinfo{journal}{\emph{Neurocomputing}}  \bibinfo{volume}{173}
  (\bibinfo{year}{2016}), \bibinfo{pages}{530--540}.
\newblock


\bibitem[\protect\citeauthoryear{Zheng, Huang, Tan, and Tao}{Zheng
  et~al\mbox{.}}{2012}]%
        {zheng2012cascade}
\bibfield{author}{\bibinfo{person}{Shuai Zheng}, \bibinfo{person}{Kaiqi Huang},
  \bibinfo{person}{Tieniu Tan}, {and} \bibinfo{person}{Dacheng Tao}.}
  \bibinfo{year}{2012}\natexlab{}.
\newblock \showarticletitle{A cascade fusion scheme for gait and cumulative
  foot pressure image recognition}.
\newblock \bibinfo{journal}{\emph{Pattern Recognition}} \bibinfo{volume}{45},
  \bibinfo{number}{10} (\bibinfo{year}{2012}), \bibinfo{pages}{3603--3610}.
\newblock


\bibitem[\protect\citeauthoryear{Zheng, Zhang, Huang, He, and Tan}{Zheng
  et~al\mbox{.}}{2011}]%
        {zheng2011robust}
\bibfield{author}{\bibinfo{person}{Shuai Zheng}, \bibinfo{person}{Junge Zhang},
  \bibinfo{person}{Kaiqi Huang}, \bibinfo{person}{Ran He}, {and}
  \bibinfo{person}{Tieniu Tan}.} \bibinfo{year}{2011}\natexlab{}.
\newblock \showarticletitle{Robust view transformation model for gait
  recognition}. In \bibinfo{booktitle}{\emph{2011 18th IEEE International
  Conference on Image Processing}}. IEEE, \bibinfo{pages}{2073--2076}.
\newblock


\bibitem[\protect\citeauthoryear{Zneit, AlQadi, and Zalata}{Zneit
  et~al\mbox{.}}{2017}]%
        {zneit2017methodology}
\bibfield{author}{\bibinfo{person}{R~Abu Zneit}, \bibinfo{person}{Ziad AlQadi},
  {and} \bibinfo{person}{M~Abu Zalata}.} \bibinfo{year}{2017}\natexlab{}.
\newblock \showarticletitle{A Methodology to Create a Fingerprint for RGB Color
  Image}.
\newblock \bibinfo{journal}{\emph{International Journal of Computer Science and
  Mobile Computing}} \bibinfo{volume}{16}, \bibinfo{number}{1}
  (\bibinfo{year}{2017}), \bibinfo{pages}{205--212}.
\newblock


\bibitem[\protect\citeauthoryear{Zoph, Ghiasi, Lin, Cui, Liu, Cubuk, and
  Le}{Zoph et~al\mbox{.}}{2020}]%
        {sota-segmentation}
\bibfield{author}{\bibinfo{person}{Barret Zoph}, \bibinfo{person}{Golnaz
  Ghiasi}, \bibinfo{person}{Tsung-Yi Lin}, \bibinfo{person}{Yin Cui},
  \bibinfo{person}{Hanxiao Liu}, \bibinfo{person}{Ekin~D Cubuk}, {and}
  \bibinfo{person}{Quoc~V Le}.} \bibinfo{year}{2020}\natexlab{}.
\newblock \showarticletitle{Rethinking pre-training and self-training}.
\newblock \bibinfo{journal}{\emph{arXiv preprint arXiv:2006.06882}}
  (\bibinfo{year}{2020}).
\newblock


\bibitem[\protect\citeauthoryear{Zou, Wang, Zhao, Wang, Shen, and Li}{Zou
  et~al\mbox{.}}{2018a}]%
        {zou2018}
\bibfield{author}{\bibinfo{person}{Qin Zou}, \bibinfo{person}{Yanling Wang},
  \bibinfo{person}{Yi Zhao}, \bibinfo{person}{Qian Wang}, \bibinfo{person}{Chao
  Shen}, {and} \bibinfo{person}{Qingquan Li}.}
  \bibinfo{year}{2018}\natexlab{a}.
\newblock \showarticletitle{Deep Learning Based Gait Recognition Using
  Smartphones in the Wild}.
\newblock \bibinfo{journal}{\emph{CoRR}}  \bibinfo{volume}{abs/1811.00338}
  (\bibinfo{year}{2018}).
\newblock
\showeprint[arxiv]{1811.00338}
\urldef\tempurl%
\url{http://arxiv.org/abs/1811.00338}
\showURL{%
\tempurl}


\bibitem[\protect\citeauthoryear{Zou, Wang, Zhao, Wang, Shen, and Li}{Zou
  et~al\mbox{.}}{2018b}]%
        {zou2018deep}
\bibfield{author}{\bibinfo{person}{Qin Zou}, \bibinfo{person}{Yanling Wang},
  \bibinfo{person}{Yi Zhao}, \bibinfo{person}{Qian Wang}, \bibinfo{person}{Chao
  Shen}, {and} \bibinfo{person}{Qingquan Li}.}
  \bibinfo{year}{2018}\natexlab{b}.
\newblock \showarticletitle{Deep Learning Based Gait Recognition Using
  Smartphones in the Wild}.
\newblock \bibinfo{journal}{\emph{arXiv preprint arXiv:1811.00338}}
  (\bibinfo{year}{2018}).
\newblock


\end{thebibliography}

\end{document}